%% file: main.tex
\documentclass[runningheads]{llncs}

\usepackage{eccv}

\definecolor{cvprblue}{rgb}{0.21,0.49,0.74}
\usepackage{amsmath}
\usepackage{pifont}
\usepackage{xcolor}
\usepackage{placeins}
\usepackage{array}
\usepackage{booktabs}    %
\usepackage{tabularx}    %
\usepackage{multirow}    %
\usepackage{siunitx}     %
\usepackage[table]{xcolor} %
\definecolor{rowhighlight}{gray}{0.9} %

\usepackage{eccvabbrv}
\usepackage{hyperref}
\usepackage{graphicx}
\usepackage{wrapfig}
\usepackage{booktabs}

\usepackage[accsupp]{axessibility}  %

\newcommand{\cmark}{\ding{51}}
\newcommand{\xmark}{\ding{55}}

\usepackage{hyperref}
\usepackage{orcidlink}

\newcommand{\ours}{ReconSplat\xspace}

\input{preamble}

\begin{document}

\title{\ours: Generalizable 3D Scene Reconstruction Beyond Observed Views} 

\titlerunning{\ours}

\author{Giuseppe Stracquadanio\inst{1}\orcidlink{0009-0005-5086-4856} \and
Kevin Raj\inst{1,2}\orcidlink{0009-0007-3271-5990} \and
Julia Grabinski\inst{1}\orcidlink{0000-0002-8371-1734} \and\\
Stefan Roth\inst{1,2,3}\orcidlink{0000-0001-9002-9832}}

\authorrunning{G.\ Stracquadanio et al.}

\institute{\textsuperscript{1}TU Darmstadt,  \textsuperscript{2}Zuse School ELIZA, 
\textsuperscript{3}hessian.AI \\
\email{\{name.surname\}@visinf.tu-darmstadt.de}}
\maketitle
\input{figures/fig_teaser_eccv}

\begin{abstract}
We introduce \ours, a feed-forward model for 3D scene reconstruction that aims to address the longstanding trade-off between plausible view generation for unobserved regions and geometric consistency, providing both geometrically aligned novel views and sharp depth estimates.
Our approach builds on 3D Gaussian splatting (3DGS) as an intermediate differentiable scene representation and integrates it with a multi-view latent diffusion model (MV-LDM) trained to act simultaneously as a \textit{refiner} and an \textit{inpainter} for appearance and scene geometry. 
We enforce geometric consistency by guiding the diffusion process with variational 3D latent features for appearance and geometry, encoded by the feed-forward 3DGS representation and rasterized to 2D latent space. \ours produces both photorealistic novel views and accurate depth maps on real-world benchmarks, RealEstate10K and DL3DV-10K, outperforming existing methods in challenging extrapolation setups.
Notably, \ours allows the extrapolation of unseen and challenging viewpoints jointly with coherent and precise scene geometry\footnote{Code and additional visual results are available on our \href{https://visinf.github.io/reconsplat}{project page}.}.

\keywords{3D Reconstruction \and Novel View Synthesis \and Generative Modeling}
\end{abstract}

\input{sec/intro}    
\input{sec/related_work}
\input{sec/method}

\input{sec/experiments}

\input{sec/conclusion}

\paragraph{\textit{Acknowledgements.}}
This project has received funding from the European Research Council (ERC) under the European Union’s Horizon 2020 research and innovation programme (grant agreement No.\ 866008), and from the DFG under Germany’s Excellence Strategy (EXC-3057/1 “Reasonable Artificial Intelligence”, Project No.\ 533677015). KR is supported by the Konrad Zuse School of Excellence in Learning and Intelligent Systems (\href{https://eliza.school/}{ELIZA}) through the DAAD programme Konrad Zuse Schools of Excellence in Artificial Intelligence, sponsored by the German Federal Ministry of Education and Research. We gratefully acknowledge support from the hessian.AI Service Center (funded by the Federal Ministry of Research, Technology and Space, BMFTR, grant No.\ 16IS22091) and the hessian.AI Innovation Lab (funded by the Hessian Ministry for Digital Strategy and Innovation, grant No.\ S-DIW04/0013/003). We thank Oliver Hahn and Christoph Reich for insightful discussions and valuable assistance with manuscript preparation. 
\bibliographystyle{splncs04}
\bibliography{main}

\clearpage
\appendix
\input{sec/appendix}

\end{document}

%% file: preamble.tex
\newcommand*{\inparagraph}[1]{\medskip\noindent\textbf{#1}\hspace{0.4em}}

\newcommand{\tablesize}{\fontsize{6pt}{7.2pt}\selectfont} %

\DeclareMathOperator*{\diag}{diag}

\usepackage[table]{xcolor}
\usepackage{svg}
\definecolor{rowhighlight}{gray}{0.9}
\usepackage[super]{nth}
\definecolor{cvprblue}{rgb}{0.21,0.49,0.74}
\usepackage{amsmath}
\usepackage{pifont}
\usepackage{xcolor}
\usepackage{placeins}
\usepackage{array}
\usepackage{booktabs}    %
\usepackage{tabularx}    %
\usepackage{multirow}    %
\usepackage{siunitx}     %
\usepackage[table]{xcolor} %
\definecolor{rowhighlight}{gray}{0.9} %

\usepackage{eccvabbrv}
\usepackage{hyperref}
\usepackage{graphicx}
\usepackage{wrapfig}
\usepackage{booktabs}

%% file: figures/fig_teaser_eccv.tex
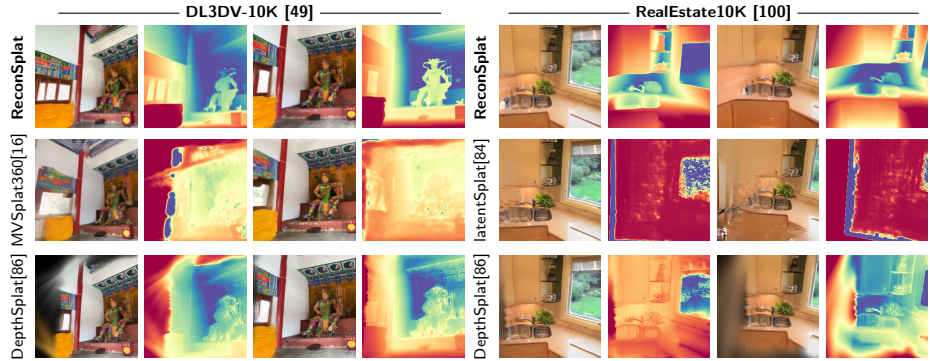
\begin{figure}[t] %
\centering
\input{figures/teaser/teaser_eccv}
\captionof{figure}{
\textbf{Qualitative comparison of \ours{}, our feed-forward 3D scene reconstruction model, to previous work.} 
Previous methods need to trade-off plausible generation in unobserved regions and geometric consistency. While \cite{chen2024mvsplat360,wewer2024latentsplat} produce plausible appearance for unobserved views, they suffer from texture inconsistencies and fail to capture the true scene geometry, resulting in noisy and fragmented depth maps. \cite{xu2025depthsplat} produces clean depth estimates, but fails to synthesize appearance and geometry in unobserved regions.  In contrast, our \ours{} achieves both consistent appearance and accurate depth across views by learning a diffusion prior guided by multi-view-consistent variational latents for RGB and depth. 
}
\label{fig:teaser}
\vspace{-1em}
\end{figure}

%% file: figures/teaser/teaser_eccv.tex
\fontsize{6pt}{7.2pt}
\sffamily
\renewcommand{\arraystretch}{1.0}

\newcommand{\imgwidth}{0.11}

\begin{tabular}{
    >{\centering\arraybackslash}m{0.02\textwidth}
    >{\centering\arraybackslash}m{\imgwidth\textwidth}
    >{\centering\arraybackslash}m{\imgwidth\textwidth}
    >{\centering\arraybackslash}m{\imgwidth\textwidth}
    >{\centering\arraybackslash}m{\imgwidth\textwidth}
    >{\centering\arraybackslash}m{0.022\textwidth}
    >{\centering\arraybackslash}m{\imgwidth\textwidth}
    >{\centering\arraybackslash}m{\imgwidth\textwidth}
    >{\centering\arraybackslash}m{\imgwidth\textwidth}
    >{\centering\arraybackslash}m{\imgwidth\textwidth}
}

& \multicolumn{4}{c}{\rule[0.5ex]{1.6cm}{0.4pt}\textbf{  DL3DV-10K~\cite{ling2024dl3dv} }\rule[0.5ex]{1.6cm}{0.4pt}}
& 
& \multicolumn{4}{c}{\rule[0.5ex]{1.6cm}{0.4pt}\textbf{ RealEstate10K~\cite{realestate10k} }\rule[0.5ex]{1.6cm}{0.4pt}} \\ [1pt]
\rotatebox[origin=lB]{90}{\hspace{-0.15em}\textbf{ReconSplat}}
& \includegraphics[width=\linewidth]{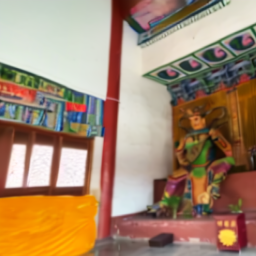} 
& \includegraphics[width=\linewidth]{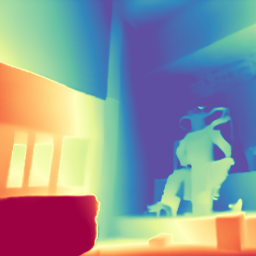}
& \includegraphics[width=\linewidth]{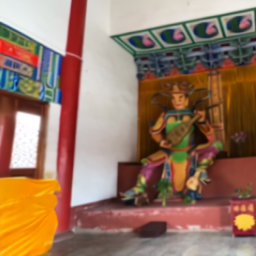} 
& \includegraphics[width=\linewidth]{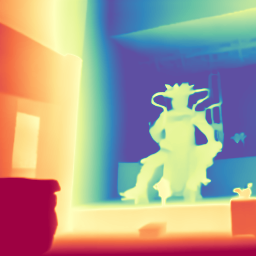} 
& \rotatebox[origin=lB]{90}{\hspace{-0.15em}\textbf{ReconSplat}}
& \includegraphics[width=\linewidth]{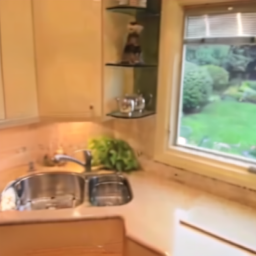} 
& \includegraphics[width=\linewidth]{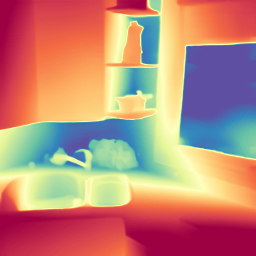} 
& \includegraphics[width=\linewidth]{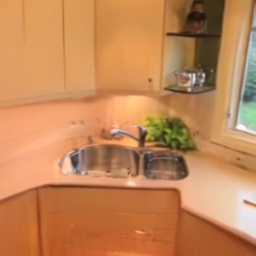} 
& \includegraphics[width=\linewidth]{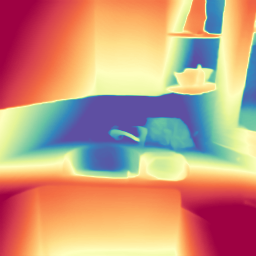} \\

\rotatebox[origin=lB]{90}{MVSplat360\hspace{-0.01em}\cite{chen2024mvsplat360}}
& \includegraphics[width=\linewidth]{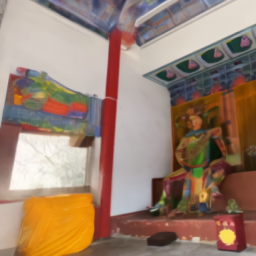} 
& \includegraphics[width=\linewidth]{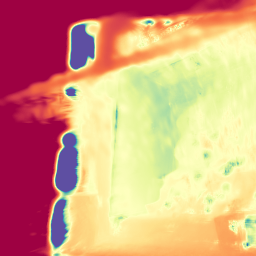} 
& \includegraphics[width=\linewidth]{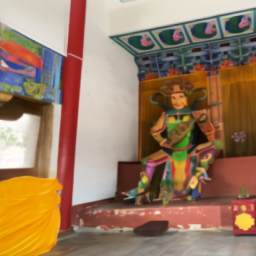} 
& \includegraphics[width=\linewidth]{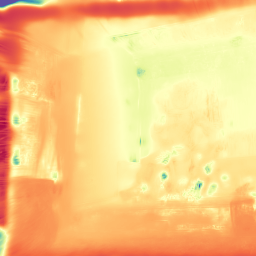} 
& \rotatebox[origin=lB]{90}{latentSplat\hspace{-0.01em}\cite{wewer2024latentsplat}}
& \includegraphics[width=\linewidth]{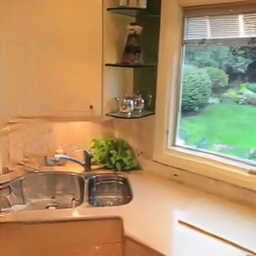} 
& \includegraphics[width=\linewidth]{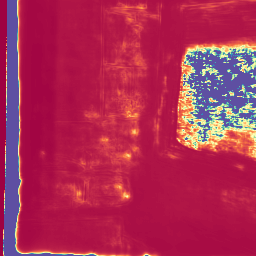} 
& \includegraphics[width=\linewidth]{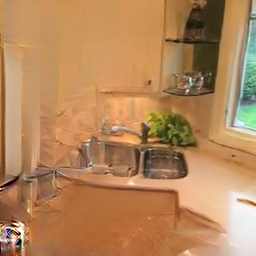} 
& \includegraphics[width=\linewidth]{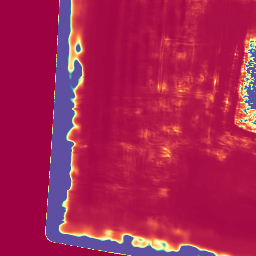} \\

\rotatebox[origin=lB]{90}{DepthSplat\hspace{-0.01em}\cite{xu2025depthsplat}}
& \includegraphics[width=\linewidth]{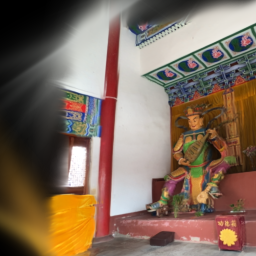} 
& \includegraphics[width=\linewidth]{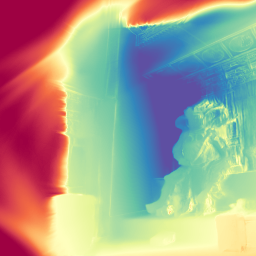} 
& \includegraphics[width=\linewidth]{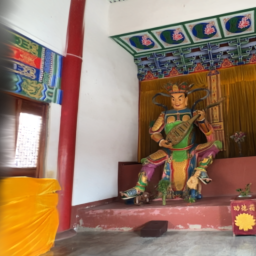}
& \includegraphics[width=\linewidth]{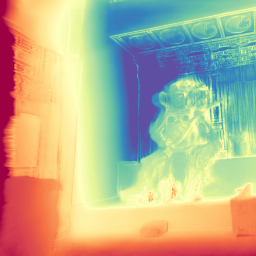}
& \rotatebox[origin=lB]{90}{DepthSplat\hspace{-0.01em}\cite{xu2025depthsplat}}
& \includegraphics[width=\linewidth]{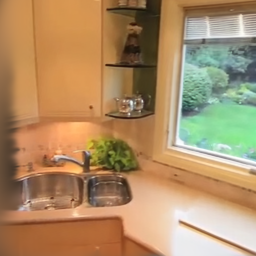} 
& \includegraphics[width=\linewidth]{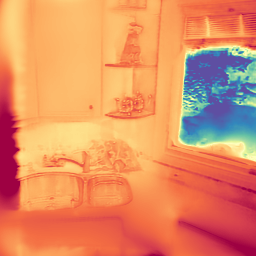} 
& \includegraphics[width=\linewidth]{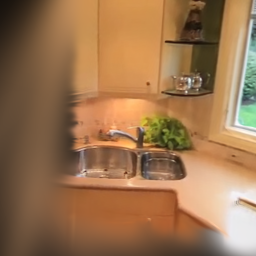} 
& \includegraphics[width=\linewidth]{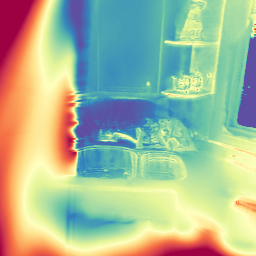} \\ [-8pt]

\end{tabular}
\normalsize \normalfont

%% file: sec/intro.tex
\section{Introduction}
\label{sec:intro}

Creating high-fidelity digital representations of real-world environments requires addressing two critical challenges: visual realism and geometric accuracy. Various differentiable rendering methods \cite{lombardi2019neural,mildenhall2021nerf,kerbl20233d,sitzmann2019scene,kato2020differentiable} laid the groundwork for 3D reconstruction and novel-view synthesis (NVS). However, most of these methods rely on per-scene optimization and dense multi-view supervision, which limit their scalability and generalization. Methods such as \cite{yu2021pixelnerf,charatan2024pixelsplat,chen2024mvsplat,chibane2021stereo} instead study feed-forward reconstruction from sparse input views, enabling generalization to unseen scenes without per-scene optimization. However, sparse-view reconstruction remains inherently ill-posed, since limited observations provide weak constraints on  unobserved geometry and appearance. 

To address this challenge, previous works \cite{wewer2024latentsplat,chen2024mvsplat360,szymanowicz2025bolt3d,liang2025wonderland,bahmani2026lyra} combine generative models \cite{rombach2022high,blattmann2023stable,gan} with a feed-forward 3D Gaussian splatting (3DGS) representation \cite{kerbl20233d}. They follow the broader success of diffusion models \cite{rombach2022high,blattmann2023stable} across multiple subfields of computer vision \cite{tian2024diffuse,chen2023diffusiondet,ruiz2023dreambooth}, including NVS \cite{gao2024cat3d, chen2024mvsplat360, yang2025prometheus,xu2025depthsplat,wewer2024latentsplat}.
However, current methods \cite{xu2025depthsplat,wewer2024latentsplat,chen2024mvsplat360} still
face a fundamental trade-off between plausible generation for unobserved regions and geometric consistency. As demonstrated in \cref{fig:teaser} \emph{(bottom)}, approaches that explicitly rely on geometric reasoning, \eg DepthSplat \cite{xu2025depthsplat}, tend to remain tightly coupled to the observed views, limiting their ability to generalize beyond the small-baseline interpolation scenario. Thus, they fail in rendering camera frustums that contain sparsely reconstructed or incomplete regions of the scene. On the other hand, generative methods like MVSplat360 \cite{chen2024mvsplat360} and latentSplat \cite{wewer2024latentsplat} produce visually plausible results for novel views, but fail to recover a structurally coherent 3D geometry of the scene, resulting in highly inaccurate depth estimates as shown in \cref{fig:teaser} \emph{(middle)}.

Our method, \emph{\ours}, achieves both \emph{(top)}: plausible novel view synthesis for unobserved viewpoints conditioned on sparse observations and 
coherent multi-view scene completion. 
We bridge this gap by jointly learning appearance and geometry priors within a feed-forward 3D reconstruction framework. 
\ours refactors a multi-view latent diffusion model (MV-LDM) to jointly sample \emph{appearance} and \emph{geometry} latent pairs that are mutually consistent and encode a shared 3D scene structure. 
To fully exploit this, we condition the diffusion process on rasterized 2D latent features derived from a variational 3DGS representation, providing structured yet coarse visual and geometric cues of the scene. Our feed-forward 3D reconstruction method \emph{(i)} yields high-fidelity reconstructions of complex scenes, \emph{(ii)} enables extrapolation to novel viewpoints besides interpolation, and \emph{(iii)} predicts multi-view-consistent depth as a by-product of our proposed joint appearance-geometry generative modeling.

We conduct experiments on the large-scale RealEstate10K (RE10K) \cite{realestate10k} and the challenging DL3DV-10K \cite{ling2024dl3dv} datasets, including complex unbounded scenes. Our \ours attains a new state-of-the-art on RE10K in sparse-view, wide-baseline extrapolation settings and demonstrates strong performance on DL3DV-10K. Moreover, it produces samples that remain consistent across viewpoints and yield dense, geometrically coherent 3D reconstructions. To the best of our knowledge, \ours is the first feed-forward model to jointly synthesize geometrically coherent novel views and depth in complex wide-baseline interpolation and extrapolation settings, as seen in \cref{fig:teaser} \emph{(top)}.

%% file: sec/related_work.tex
\section{Related Work}

\subsubsection{Sparse-view Novel View Synthesis.}
Early differentiable neural rendering methods~\cite{mildenhall2021nerf, barron2022mip, barron2023zip, sitzmann2019scene}, as well as explicit scene representations such as 3DGS~\cite{kerbl20233d}, achieve high-fidelity NVS through dense multi-view capture and per-scene optimization. While effective, both the reliance on abundant view coverage and the computational cost of per-scene optimization make these methods impractical for real-world sparse-view scenarios. To address these limitations, several works incorporate explicit geometric supervision into the optimization process. Some methods use monocular depth estimates or surface normals as auxiliary signals to constrain scene geometry~\cite{deng2022depth, yu2022monosdf, chung2024depth}, while others enforce multi-view consistency through epipolar constraints~\cite{chibane2021stereo} or local surface priors~\cite{raj2025spurfies}. Another line of work directly penalizes the rendered appearance and geometry of unobserved viewpoints during optimization to discourage degenerate solutions~\cite{niemeyer2022regnerf, hamdi2023sparf}. For better cross-scene generalization, \cite{yu2021pixelnerf,wang2021ibrnet} condition radiance fields on image features and \cite{wang2023sparsenerf, roessle2022dense} incorporate depth and semantic priors. However, all of these methods still require per-scene or test-time optimization, which is computationally expensive and impractical for real-world use. Addressing both shortcomings, our \ours learns generalizable priors within a feed-forward framework, enabling scalable and efficient scene completion without per-scene overhead.

\inparagraph{Feed-forward Scene Reconstruction and Synthesis.}
Feed-forward methods \cite{hong2024lrm,wang2025vggt,chen2024mvsplat,wang2024dust3r,charatan2024pixelsplat,li2024instantd,wang2025amb3r,jevtic2025scenedino} have significantly advanced 3D reconstruction by learning strong data-driven priors from large multi-view datasets \cite{realestate10k, ling2024dl3dv, yeshwanth2023scannet++, reizenstein2021common, huang2018deepmvs}, enabling fast and generalizable scene reconstruction without per-scene optimization. Initial works \cite{wang2024dust3r, leroy2024grounding, wang2025vggt} directly regress pointmaps from image pairs or small view sets for feed-forward scene reconstruction. Later methods \cite{yang2025fast3r, chen2025long3r} scale to long-range sequences by processing multiple frames in parallel, while \cite{wang20253d, wang2025continuous} incorporate spatial memory for scalability. %
A complementary line of research addresses scene reconstruction by directly predicting a 3DGS representation in a feed-forward manner \cite{charatan2024pixelsplat,chen2024mvsplat, szymanowicz2024splatter, zheng2024gps}. Building on an epipolar transformer \cite{he2020epipolar}, pixelSplat~\cite{charatan2024pixelsplat} proposes to directly regress pixel-aligned 3D Gaussians from pairs of images. %
Follow-up works \cite{chen2024mvsplat, szymanowicz2024splatter} extend this paradigm to handle more flexible input configurations. %
More recent methods \cite{zhang2024gs, ziwen2025long, kang2025ilrm, kang2025multi, cai2025baking, jiang2025anysplat, wang2024freesplat} scale this paradigm using end-to-end transformer-based architectures, effectively learning strong 3D priors from large multi-view datasets to directly regress Gaussian splats. While these methods are fast and generalizable, they are limited to interpolation between observed views and lack the generative components necessary for reasonable scene extrapolation. Our \ours directly addresses this limitation by incorporating multi-view latent diffusion priors to synthesize unobserved regions in a plausible and multi-view-consistent manner. 

\inparagraph{Appearance and Depth Priors for Novel-View Synthesis.}
Generative models have demonstrated remarkable progress in synthesizing high-fidelity images \cite{ruiz2023dreambooth, rombach2022high, saharia2022photorealistic, ramesh2022hierarchical}, videos \cite{blattmann2023align, ho2022imagen, blattmann2023stable}, and 3D assets \cite{hu2024mvd, poole2022dreamfusion,guizilini2025zero} from text or image inputs. To adapt these models as priors for NVS, recent works incorporate explicit camera-pose conditioning \cite{wang2024motionctrl, bahmani2025ac3d, bahmani2024vd3d, voleti2024sv3d}, enabling 3D-aware image and video generation. Early works \cite{chan2023generative, wu2024reconfusion} condition 2D diffusion models on 3D-derived features to improve view consistency, while \cite{gao2024cat3d, viewcrafter2025, schwarz2025generative, liang2025wonderland} leverage multi-view or video diffusion priors to plausibly synthesize missing content. Yet, these methods require post-hoc optimization or distillation~\cite{barron2023zip} to enforce 3D consistency. More recently, \cite{szymanowicz2025bolt3d, lin2025diffsplat} repurpose diffusion frameworks to directly generate 3DGS primitives and \cite{wewer2024latentsplat,chen2024mvsplat360} combine 3DGS representations with downstream generative models, decoding via GANs~\cite{gan}, or video diffusion~\cite{blattmann2023stable} to synthesize occluded and unobserved regions. While these methods produce multi-view consistent renderings, their underlying representations remain weakly grounded in 3D geometry, resulting in inaccurate and inconsistent depth estimates (see \cref{fig:teaser}). %

To improve geometric grounding, a complementary line of work leverages depth priors \cite{deng2022depth, zhang2025transplat, szymanowicz2025flash3d, xu2025depthsplat, yang2025prometheus}. Specifically, \cite{deng2022depth, zhang2025transplat} rely on monocular depth estimates, %
while \cite{szymanowicz2025flash3d, yang2025prometheus} lift single-view depth predictions directly into 3D Gaussian representations. More recently, DepthSplat~\cite{xu2025depthsplat} couples pre-trained monocular depth features with a feed-forward 3DGS framework, achieving strong results in narrow-baseline NVS. However, these methods are primarily interpolative, with performance degrading significantly in wide-baseline settings. Furthermore, they lack the generative models necessary to synthesize plausible geometry for unobserved regions, resulting in inconsistent depth estimates (see \cref{fig:teaser}).
In contrast, our \ours jointly learns appearance and geometry priors within a unified multi-view latent diffusion framework conditioned on a feed-forward 3DGS representation, enabling plausible and 3D-consistent scene completion under sparse, wide-baseline inputs.

%% file: sec/method.tex
\section{Method: ReconSplat}
\label{sec:method}
We propose \ours, a feed-forward generalizable 3D scene reconstruction approach. %
Given a set of $N$ context (\ie, input) RGB images $\mathcal{I}^c = \{\mathbf{I}_i^c\}_{i=1}^{N}$, $\mathbf{I}^c_i \in \mathbb{R}^{H\times W\times 3}$, with known camera poses, our goal is to synthesize coherent, geometrically consistent novel views and simultaneously recover the underlying scene geometry. We specifically design our method for robustness and geometric consistency under the challenging sparse-view, wide-baseline setup.

\inparagraph{Overview.}
To achieve this, we propose a two-stage design, integrating a multi-view latent diffusion model (MV-LDM) with a feed-forward 3D Gaussian splatting representation. We provide an overview of our architecture in~\cref{fig:method_fig}. First, we refactor a state-of-the-art multi-view 3DGS prediction backbone \cite{chen2024mvsplat} to output a 3D variational latent field \cite{wewer2024latentsplat} for view-dependent appearance and geometry representation (\cref{sec:ff-3dgs}).
Different from previous work \cite{wewer2024latentsplat}, we rasterize the 3D variational latent field into 2D (\cref{sec:mv_latent_field}) and fine-tune a variational auto-encoder (VAE) to reconstruct the target views and relative depth from the rasterized latent field in pixel space (\cref{subsec:learning_mv_latent_features}). Finally, our MV-LDM (\cref{subsec:mv_latent_diffusion}) refines these preliminary latents, which serve as conditioning signal to the MV-LDM, enforcing both cross-modality consistency (appearance \vs geometry) and coherence across observed as well as unobserved viewpoints. This design encourages view-consistent synthesis for novel target views and coherent geometry across each context and predicted view frustum.
\input{figures/fig_method_new}

\inparagraph{Pre-processing.}
Our method assumes known camera extrinsics $\mathbf{T} = [\mathbf{R}|\mathbf{t}], \mathbf{T} \in \mathbb{R}^{3\times4}$ and intrinsics $\mathbf{K} \in \mathbb{R}^{3\times3}$ for both context and target views. These are obtained through an offline pre-processing stage, where we reconstruct the scene and extract camera parameters together with dense multi-view relative depth pseudo-labels $\mathbf{D}$, which are used to supervise geometry during training. To this end, we rely on the state-of-the-art dense feed-forward reconstruction model VGGT \cite{wang2025vggt}. We refer to the estimated camera parameters through their full projection matrix $\mathbf{P} = \mathbf{K} [\mathbf{R} | \mathbf{t}]$ for brevity. In the following, we use $\mathbf{P}^c_i$ to denote the projection matrix of the $i^\text{th}$ context view and $\mathbf{P}^{\odot}_j$ for the $j^\text{th}$ target view.

\subsection{Feed-forward 3D Latent Field}
\label{sec:ff-3dgs}
We predict pixel-aligned 3D Gaussians in a single feed-forward pass to obtain a consistent 3D scene representation.
In practice, we use the MVSplat multi-view encoder~\cite{chen2024mvsplat} to predict Gaussian splat parameters $\{(\mathbf{X}_k, \mathbf{\Sigma}_k, \alpha_k, \mathbf{c}_k)\}_{k=1}^{N\cdot H\cdot W}$ from the $N$ context views (\cf  \cref{fig:method_fig}, Stage 1).
Here, $\mathbf{X}_k\in\mathbb{R}^3$ is the 3D position of the Gaussian with covariance $\mathbf{\Sigma}_k\in\mathbb{R}^{3\times 3}$, representing its orientation and shape, $\alpha_k$ defines the opacity, and $\mathbf{c}_k\in\mathbb{R}^h$ the spherical harmonics of the color representation. We extend the representation through latent fields for \textit{appearance} and \textit{geometry}, building upon the \textit{variational Gaussians} formulation of latentSplat \cite{wewer2024latentsplat}.
To that end, our multi-view encoder additionally predicts, for each Gaussian, appearance and geometry latent distributions parameterized by $\boldsymbol{\mu}_k^a,\boldsymbol{\sigma}_k^a\in\mathbb{R}^4$ and $\boldsymbol{\mu}_k^g,\boldsymbol{\sigma}_k^g\in\mathbb{R}^4$, with diagonal covariances $\diag((\boldsymbol{\sigma}_k^l)^2)$, $l\in\{a,g\}$.
Each latent distribution is 4-dimensional to be consistent with standard LDM latents \cite{rombach2022high}. Our 3D representation is thus defined as the collection of per-pixel 3D Gaussian primitives $\mathcal{G}=\{(\mathbf{X}_k, \mathbf{\Sigma}_k, \alpha_k, \mathbf{c}_k, \boldsymbol{\mu}^a_k, \boldsymbol{\sigma}_k^a, \boldsymbol{\mu}_k^g, \boldsymbol{\sigma}_k^g)\}_{k=1}^{N \cdot H \cdot W}$.

\subsection{Stage 1: Rasterizing Variational Parameters}
\label{sec:mv_latent_field}
Although our \textit{variational} Gaussian representation shares the feed-forward prediction of appearance distributional parameters with~\cite{wewer2024latentsplat}, we use it differently. Rather than following a \textit{sample-then-rasterize} scheme, \ours adopts a \textit{rasterize-then-sample} formulation: we first rasterize the variational parameters of the 3D latent field into target-view latent distributions, and only then sample from the resulting 2D distributions. 
This formulation allows us to regularize the rasterized distributions directly in the 2D latent space. Specifically, we apply a KL regularization~\cite{kingma2013auto} that encourages the rasterized preliminary latents to follow the prior distribution of the pre-trained Stable Diffusion VAE (SD-VAE)~\cite{rombach2022high}, making a lightweight adaptation \cite{hu2022lora} sufficient for decoding. 
In detail, each 3D Gaussian primitive encodes a Gaussian distribution over latent features. When multiple primitives contribute to a pixel $p$, $\alpha$-blending induces a weighted mixture of these latent distributions along the projection ray. Since this mixture is generally not Gaussian, we \textit{approximate} it with a Gaussian obtained by matching its first and second moments.
We generalize standard $\alpha$-blending to operate on variational latent distributions and derive the per-pixel variational mean $\bar{\boldsymbol{\mu}}_p$ and (diagonal) covariance $\bar{\boldsymbol{\sigma}}_p^2$ in closed form as:
\begin{subequations}\label{eq:gauss-raster}
    \begin{align}
        \bar{\boldsymbol{\mu}}_p &= \mathbf{m}_{1,p} & \mathbf{m}_{1,p} &= \frac{1}{A_p}\sum\nolimits_{k}\alpha_{p,k} T_{p,k} \boldsymbol{\mu}_k, \label{eq:1a}\\
\bar{\boldsymbol{\sigma}}_p^2 & = \mathbf{m}_{2,p} - \mathbf{m}_{1,p}^2 & \mathbf{m}_{2,p} &= \frac{1}{A_p}\sum\nolimits_{k}\alpha_{p,k} T_{p,k} \big(\boldsymbol{\sigma}_k^2 + \boldsymbol{\mu}_k^2\big). \label{eq:1b}
    \end{align}
\end{subequations}
Here, $A_p=\sum\nolimits_k \alpha_{p,k} T_{p,k}$ is the accumulated opacity at pixel $p$, $\boldsymbol{\mu}_k$ and $\boldsymbol{\sigma}^2_k$ are the mean and (diagonal) covariance of the latent distribution encoded by the $k^\text{th}$ 3D Gaussian, and $\alpha_{p,k}$ its opacity contribution to pixel $p$ given its projected footprint. The index $k$ ranges over the Gaussians intersecting the ray through pixel $p$.
$T_{p,k} = \prod_{t<k}(1-\alpha_{p,t})$ is the transmittance up to the $k^\text{th}$ 3D Gaussian along the ray during front-to-back compositing~\cite{kerbl20233d}. A more detailed derivation of the per-pixel mean and covariance expressions is provided in \cref{app:rasterizer_math}.
We apply~\cref{eq:gauss-raster} for both appearance and geometry latents, performing $\alpha$-blending at each pixel of the latent image grid. 
Given a camera view $\mathbf{P}$, rasterization produces 2D latent variational parameters $\mathbf{r} = \big(\bar{\boldsymbol{\mu}}^a, \bar{\boldsymbol{\sigma}}^a, \bar{\boldsymbol{\mu}}^g, \bar{\boldsymbol{\sigma}}^g\big)$ at resolution $H/8\times W/8$. We denote this as $\mathbf{r} = \mathcal{R}(\mathcal{G}\mid \mathbf{P})$, where $\mathcal{R}$ renders the variational parameters of $\mathcal{G}$ into the latent image grid of view $\mathbf{P}$.

\subsection{Learning the Multi-view Latent Field}
\label{subsec:learning_mv_latent_features}

A naïve approach to training our 3D multi-view-consistent latent field would be to train in \emph{2D latent space} by matching our rasterized 2D latents to those encoded by the pre-trained SD-VAE~\cite{rombach2022high}.
In practice, however, we found that this causes training to saturate early, leading to blurry reconstructions at the VAE decoder output. We attribute this behavior to two factors. First, the SD-VAE encoder is not multi-view-consistent, leading to inconsistent supervision across views. Second, its latent statistics differ from those induced by rasterizing variational Gaussians. Thus, to enable the SD-VAE decoder to process our rasterized latents, we utilize parameter-efficient LoRA fine-tuning~\cite{hu2022lora}, optimizing only a small fraction of parameters ($\sim2\,$\%) instead of fine-tuning the entire decoder. 
We adapt two separate decoders  $\mathcal{D}_a, \mathcal{D}_g$ for appearance and geometry latents, respectively.
We sample our preliminary multi-view-consistent 2D latents using the reparameterization trick~\cite{kingma2013auto} as $\tilde{\mathbf{z}}^{\ell}=\bar{\boldsymbol{\mu}}^{\ell}+\bar{\boldsymbol{\sigma}}^{\ell}\odot\boldsymbol{\epsilon}^{\ell}$, with $\boldsymbol{\epsilon}^{\ell}\sim\mathcal{N}(\mathbf{0},\mathbf{1})$ and $\ell\in\{a,g\}$, and obtain RGB and depth reconstructions as $\hat{\mathbf{I}} = \mathcal{D}_a(\tilde{\mathbf{z}}^{a})$ and $\hat{\mathbf{D}} = \mathcal{D}_g(\tilde{\mathbf{z}}^{g})$.
We optimize the 3D latent field and fine-tune our decoders via LoRA by minimizing the following training objective in pixel space:
\begin{align}
    \mathcal{L}_{f}(\hat{\mathbf{I}}, \hat{\mathbf{D}})
    &= \mathcal{L}_{a}(\hat{\mathbf{I}}) + \mathcal{L}_{g}(\hat{\mathbf{D}})+ \nonumber \\ &\quad + \omega_{d, a}\,\text{D}_\text{KL}\big(q_{a}(\tilde{\mathbf{z}}^a | \mathcal{G})\|\, p(\tilde{\mathbf{z}}^a)\big) + \omega_{d, g}\,\text{D}_\text{KL}\big(q_{g}(\tilde{\mathbf{z}}^g | \mathcal{G})\|\, p(\tilde{\mathbf{z}}^g)\big),
\end{align}
where $\mathcal{L}_{a}(\hat{\mathbf{I}})$ and $\mathcal{L}_{g}(\hat{\mathbf{D}})$ are appearance and geometry losses, respectively. %
We regularize using the KL-divergence 
$\text{D}_\text{KL}(\cdot \,|\, p(\tilde{\mathbf{z}}^\ell))$ between the predicted variational posterior $q(\tilde{\mathbf{z}}^\ell|\mathcal{G})$, and a latent prior $p(\tilde{\mathbf{z}}^\ell)=\mathcal{N}(\mathbf{0}, \mathbf{I})$ where $\ell\in\{a,g\}$.

\inparagraph{Learning Multi-view Appearance.}
For appearance, we follow well-known practices for VAEs and  train using a combination of reconstruction, perceptual \cite{lpips}, and adversarial losses \cite{gan} between the reconstruction $\hat{\mathbf{I}}$ and the ground-truth target view $\mathbf{I}$:
\begin{align}
    \mathcal{L}_{a}(\hat{\mathbf{I}})
    &= \mathcal{L}_1(\mathbf{I}, \hat{\mathbf{I}})
    + \omega_\text{per}\, \text{LPIPS}(\mathbf{I}, \hat{\mathbf{I}}) + \omega_\text{adv}\, \lambda\, \text{GAN}(\mathbf{I}, \hat{\mathbf{I}}),
\end{align}
where $\lambda$ is the adaptive weight introduced by \cite{esser2021taming}, which balances the reconstruction and adversarial objectives during optimization.

\inparagraph{Learning Multi-view Geometry.}
To learn a multi-view latent representation for geometry, we train the decoder $\mathcal{D}_g$ to regress \textit{relative} pseudo-depth $\hat{\mathbf{D}}$. 
We empirically found a combination of regression, perceptual, and gradient-matching losses to perform best. We use the ``generalized'' Charbonnier loss $\rho_\alpha(x)$ \cite{barron2019general,sun2010secrets} as our regression term, where we set $\alpha = 1$. For the perceptual term, we penalize structural \textit{dissimilarity} between predicted and target depth maps, treated as images, \ie, after normalization to the image range $[0,1]$. For the gradient matching loss, $\mathcal{L}_{\text{gm}}$, we follow \cite{ranftl2020midas,yang2024depthv2}.
The full geometric loss is given as:
\begin{equation}
\begin{split}
    \mathcal{L}_g(\hat{\mathbf{D}})
    &= \rho_{\alpha}(\mathbf{D}, \hat{\mathbf{D}})
    + \omega_\text{dis}\, (1 - \text{SSIM}(\mathbf{D}, \hat{\mathbf{D}})) + \omega_{\text{gm}}\,\mathcal{L}_\text{gm}(\mathbf{D}, \hat{\mathbf{D}}), \\
    &\quad\text{where}~\mathcal{L}_\text{gm}(\mathbf{D}, \hat{\mathbf{D}}) = |\nabla_{x}\mathbf{D} - \nabla_{x}\hat{\mathbf{D}}|
    + |\nabla_{y}\mathbf{D} - \nabla_y\hat{\mathbf{D}}|.
    \end{split}
\end{equation}

Further, to stabilize training, we employ an additional auxiliary photometric loss $\mathcal{L}_\text{aux}$ between the color renderings $\hat{\mathbf{I}}_{\mathcal{G}}$ obtained by rasterizing the Gaussian parameters $\{(\mathbf{X}_k, \mathbf{\Sigma}_k, \alpha_k, \mathbf{c}_k)\}_{k=1}^{N\cdot H\cdot W}$ and the ground-truth target view $\mathbf{I}$, following prior works \cite{wewer2024latentsplat,chen2024mvsplat360}.
$\mathcal{L}_\text{aux}$ acts as the only source of gradients to the structural parameters of the Gaussian primitives. 
Thus, the complete training objective for our 3D latent field is defined as follows:
\begin{equation}
    \mathcal{L}(\hat{\mathbf{I}},\hat{\mathbf{I}}_{\mathcal{G}},\hat{\mathbf{D}}) = \mathcal{L}_{f}(\hat{\mathbf{I}},\hat{\mathbf{D}}) + \omega_\text{aux} \,\mathcal{L}_\text{aux}(\hat{\mathbf{I}}_{\mathcal{G}}).
\end{equation}

\subsection{Stage 2: Multi-View Latent Diffusion}
\label{subsec:mv_latent_diffusion}
After training our multi-view encoder-decoder architecture, we can infer a \textit{3D latent field} from $N$ context views and rasterize it into multi-view-consistent 2D latent variational parameters $\mathbf{r}_j$ for any target viewpoint  $\mathbf{P}^{\odot}_j$: $\mathbf{r}_j = \mathcal{R}(\mathcal{G}~|\mathbf{P}^{\odot}_j)$.
However, the initial scene representation $\mathcal{G}$ is constructed as the union of Gaussian primitives predicted within the context-view frustums, and therefore only covers regions observed by the input cameras. As a result, when rendering a novel viewpoint, pixels corresponding to previously unseen scene regions receive no latent features, since no Gaussian primitives project to them.

To address this limitation, we use the learned latent field as a coarse conditioning signal for observed regions and train a multi-view latent diffusion model (MV-LDM) to act as both a \emph{refiner} and \emph{inpainter} for appearance and depth.
Given a \textit{partial} reconstruction $\mathcal{G}$, obtained from $N$ calibrated context images $\mathcal{I}^{c}$ with cameras $\mathcal{P}^{c} = \{\mathbf{P}^{c}_i\}^N_{i=1}$, and $M$ target cameras $\mathcal{P}^{\odot} = \{\mathbf{P}_j^{\odot}\}_{j=1}^{M}$, we first rasterize $\mathcal{G}$ into 2D variational parameters and sample \textit{preliminary} target latents: $\tilde{\mathbf{z}}^{\odot}_j \sim \mathcal{R}(\mathcal{G} \mid \mathbf{P}^{\odot}_j)$.
We denote the rasterized preliminary latents 
as $\mathcal{\tilde{Z}}^{\odot} = \{\tilde{\mathbf{z}}^{\odot}_j\}^{M}_{j=1}$.
Context RGB images are encoded with the pre-trained SD-VAE into appearance latents $\mathcal{Z}^{c} = \{\mathbf{z}^{c}_i\}^N_{i=1}$; no context depth is used, as we assume only RGB inputs are available at test time. 
Our MV-LDM then samples refined target latents $\mathcal{Z}^{\odot} = \{\mathbf{z}^{\odot}_j\}^{M}_{j=1}$ from the conditional distribution
\begin{equation}
    p(\mathcal{Z}^{\odot}  | \mathcal{\tilde{Z}}^{\odot} , \mathcal{P}^{\odot}, \mathcal{Z}^{c}, \mathcal{P}^{c}),
\label{eq:residual_cond_dist}
\end{equation}
enabling joint appearance and geometry generation from sparse RGB inputs and the partial 3DGS representation. 

\inparagraph{Architecture.}
For our MV-LDM, we build upon the U-Net~\cite{ronneberger2015unet} architecture of SD2.1~\cite{rombach2022high} and adapt it for multi-view generation (\cf \cref{fig:method_fig}, Stage 2). 
To preserve its pre-trained generative prior, we introduce multi-view interactions through zero-initialized residual branches, rather than modifying the original network pathways. A schematic overview of the modified U-Net block is shown in \cref{fig:unet_block}. Specifically, we augment each U-Net block with cross-view 3D self-attention~\cite{shi2023mvdream, gao2024cat3d, asim2025met3r} to enable cross-view communication between tokens, while retaining the original frame-wise processing path of the U-Net. To condition on cameras, we apply PRoPE~\cite{li2025prope} within each 3D self-attention block.
We \textit{optionally} introduce 3D convolutional branches conditioned on camera parameters through adaLN~\cite{peebles2023scalable}, inspired by temporal blocks in video diffusion models~\cite{blattmann2023align,blattmann2023stable}.
\input{figures/fig_unet_block}
We implement them as additional zero-initialized residual pathways within each U-Net block and train them in a subsequent fine-tuning stage, while keeping the rest of the U-Net frozen to preserve its flexibility for unordered target views (\cf \cref{sec:ablating_3dconvs}). These branches promote locally smooth and consistent predictions across neighboring views in dense camera trajectories.
Finally, following the same design philosophy of minimizing disruption to the pre-trained single-view prior throughout optimization, we inject the rasterized latent features through a ControlNet-style~\cite{zhang2023adding} conditioning scheme. Specifically, the features are projected via $\phi$ to the corresponding U-Net decoder dimensions and added to the middle block and decoder skip connections, rather than modifying the U-Net input layers to accommodate additional channels.

\inparagraph{Learning Multi-view Refinement.} We adopt the DDPM  formulation~\cite{ho2020ddpm} with $\boldsymbol{v}$-prediction~\cite{salimans2022v} for the diffusion objective:
\begin{equation}
    \mathcal{L}_{\boldsymbol{v}}(\theta) := \mathbb{E}_{t, \boldsymbol{\epsilon}, \mathcal{Z}_0} \| \boldsymbol{v} - \boldsymbol{v}_{\theta} (\mathcal{Z}_t; \phi(\mathcal{\tilde{Z}}^{\odot}) , \mathcal{P}^{\odot}, \mathcal{Z}^{c}, \mathcal{P}^{c})\|^2_2,~ \boldsymbol{v} = \{\alpha_t \boldsymbol{\epsilon}_j -  \rho_t \mathbf{z}_{0,j}\}^{M}_{j=1}, 
\end{equation}
where $t\sim \mathcal{U}[0,T]$, $\boldsymbol{\epsilon}_j \sim\mathcal{N}(\mathbf{0}, \boldsymbol{I})$, and $\mathcal{Z}_t = \{\alpha_t \mathbf{z}_{0,j} + \rho_t \boldsymbol{\epsilon}_j\}^{M}_{j=1}$ denotes the noisy target latents, with $\alpha_t$ and $\rho_t$ given by the noise schedule~\cite{ho2020ddpm}.
During training, we use annealed multi-resolution noise~\cite{ke2024marigold}, which perturbs latents at multiple spatial scales to promote both global structure and fine detail recovery. 
Context views are provided as clean SD-VAE latents, without perturbing them with noise. The loss is applied only to target latents~\cite{asim2025met3r, gao2024cat3d}. 

\inparagraph{Classifier-free Guidance.}
We employ classifier-free guidance \cite{ho2022cfg} by defining two denoisers: the conditional model $\boldsymbol{v}_{\theta} (\mathcal{Z}_t | \mathbf{c})$, with $\mathbf{c} = (\tilde{\mathcal{Z}^{\odot}}, \mathcal{Z}^c)$, and the unconditional model $\boldsymbol{v}_{\theta} (\mathcal{Z}_t | \varnothing)$, which we obtain by dropping conditioning on context images and preliminary latents. At training time, we drop conditioning with a probability $p_\text{uncond}=0.1$. During sampling, we combine predictions as:
\begin{equation}
    \tilde{\boldsymbol{v}}_{\theta}(\mathcal{Z}_t | \mathbf{c})= (1-\gamma_\text{CFG})\,\boldsymbol{v}_{\theta}(\mathcal{Z}_t | \varnothing) + \gamma_\text{CFG}\,\boldsymbol{v}_{\theta}(\mathcal{Z}_t | \mathbf{c}).
\end{equation}

%% file: figures/fig_method_new.tex
\begin{figure*}[t]
    \centering
    \includegraphics[width=0.99\linewidth]{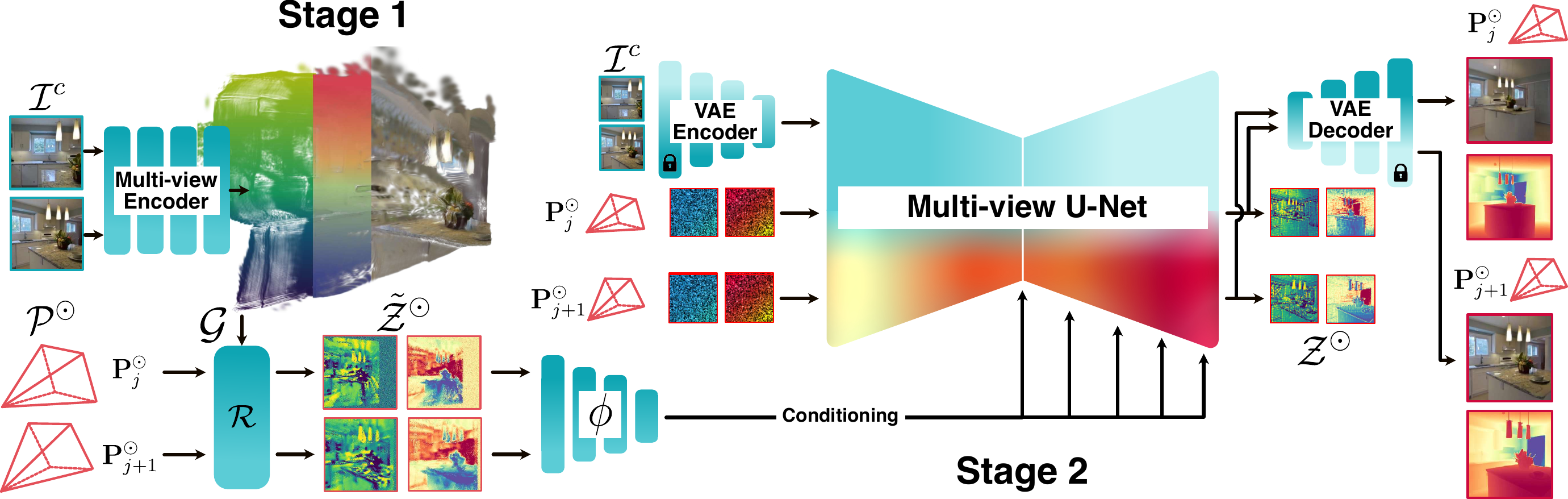}
    \caption{\textbf{Overview of \ours{}.} 
    From sparse context views $\mathcal{I}^{c}$, we regress a 3DGS  representation $\mathcal{G}$ encoding variational appearance and geometry 3D latent fields.
    Given target viewpoints $\mathcal{P}^{\odot}$, we rasterize the variational distributions from 3D into 2D, allowing us to sample preliminary 2D latents $\tilde{\mathcal{Z}}^{\odot}$, \ie, $\tilde{\mathcal{Z}}^{\odot} \sim \mathcal{R}(\mathcal{G}\mid \mathcal{P}^{\odot})$. 
    These are projected via $\phi$ and refined by a multi-view diffusion model %
    into target latents $\mathcal{Z}^{\odot}$, encoding high-quality novel views and depth maps, conditioning on context views $\mathcal{I}^{c}$.
    }
    \label{fig:method_fig}
    \vspace{-1em}
\end{figure*}

%% file: figures/fig_unet_block.tex
\begin{wrapfigure}{r}{0.33\linewidth}
    \centering
    \vspace{-2em}
    \includegraphics[width=\linewidth]{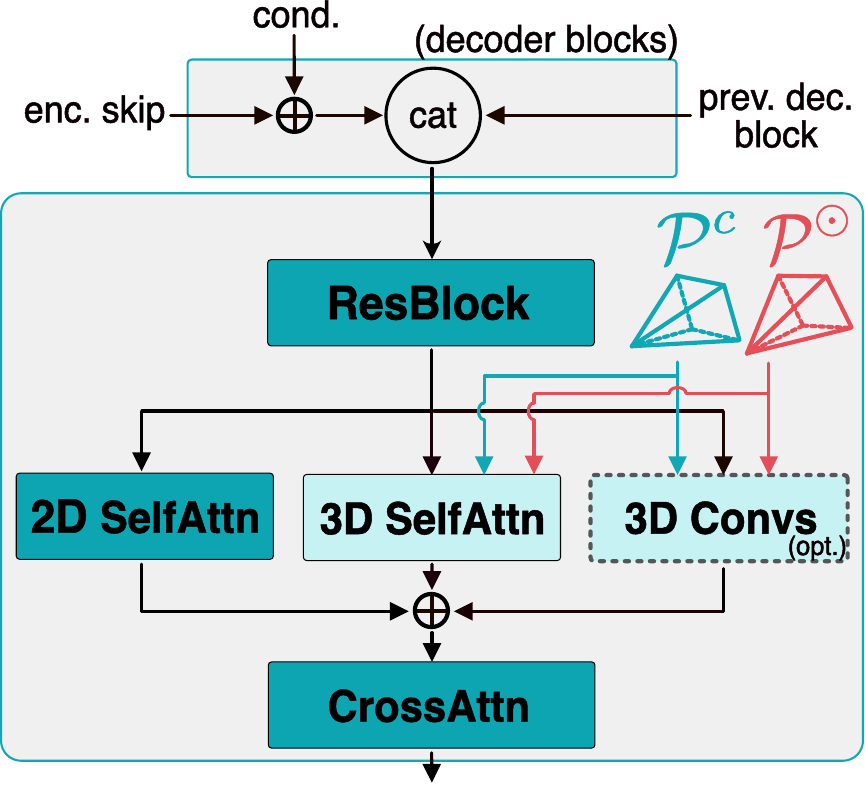}
    \vspace{-2em}
    \caption{\textbf{U-Net block.} We add two camera-conditioned branches: 3D self-attention for cross-view communication and \textit{optional} 3D convolutions for smooth nearby-view transitions.}
    \label{fig:unet_block}
    \vspace{-2em}
\end{wrapfigure}

%% file: sec/experiments.tex
\section{Experiments}

\input{figures/fig_qualitive_re10k}
\input{tables/tab_re10k_eccv}
\input{figures/fig_qualitive_dl3dv}
\input{sec/experimental_setup}

\subsection{Quantitative Results}

\subsubsection{Results on RealEstate10K.}
As can be seen in \cref{tab:results_re10k}, our method achieves competitive results on RE10K in the interpolation setting, while clearly outperforming all baselines in the extrapolation setting across the majority of metrics (except for SSIM). 
Extrapolation is inherently ill-posed: large portions of the target view lie outside the observed scene content, and multiple plausible reconstructions may exist. In such scenarios, models must balance two competing objectives: producing visually plausible completions while maintaining consistency with the underlying scene geometry. 
In this challenging setting, our results demonstrate that \ours effectively bridges this gap, achieving the best performance on perceptual and generative metrics, as well as PSNR. The results establish a new state of the art in the extrapolation setup for RE10K. 

\input{tables/tab_dl3dv_eccv}
\inparagraph{Results on DL3DV-10K.}
On DL3DV-10K, we report results under two trajectory spans~\cite{chen2024mvsplat360} in \cref{tab:results_dl3dv_new}: $n=300$, covering the full camera trajectory, which is typically two rounds, and $n=150$, covering 
a single scene round with a smaller frame-distance span. Across both settings, \ours demonstrates consistently strong performance on pixel-aligned, perceptual, and generative metrics. In particular, our method achieves the
strongest overall results among directly comparable approaches. While DepthSplat~\cite{xu2025depthsplat} reports competitive results, it is trained on the full DL3DV-10K dataset ($\sim$5$\times$ more data than other baselines), making a direct comparison less straightforward. Despite this disparity in training data, \ours obtains a lower FID in both settings.

\input{tables/tab_depth_and_cross_data}
\inparagraph{Depth Estimation.}
\label{sec:novel_view_depth_estimation}
We measure the novel-view depth-synthesis capability of \ours on ScanNet++~\cite{yeshwanth2023scannet++}, which provides GT depth. We conduct this evaluation using our DL3DV-10K model weights, in a cross-dataset generalization setting. We evaluate on the \texttt{nvs\_test\_iphone} split, comprising 12 RGB-D videos of indoor scenes captured with an iPhone. For each scene, we sample $N=4$ context views and $M=90$ target views, using the same sampling strategy as for DL3DV-10K (\cf \cref{app:sampling_protocol}). We report standard \cite{ranftl2020midas,yang2024depthv2} depth estimation metrics: threshold accuracy $\delta_1$, mean absolute relative error (AbsRel), RMSE, and RMSE$_\text{log}$. As shown in~\cref{tab:scannet}, \ours outperforms MVSplat360~\cite{chen2024mvsplat360} and DepthSplat~\cite{xu2025depthsplat} on all metrics except $\delta_1$.
Notably, improvements in AbsRel and RMSE indicate that \ours produces relative depth estimates that are geometrically closer to the ground truth, while gains in RMSE$_\text{log}$ reflect more accurate depth predictions over varying depth scales.

\subsection{Qualitative Results}
\input{figures/fig_pc}
\subsubsection{Comparisons with Baselines.} 
We present qualitative comparisons on RE10K in \cref{fig:qualitive_re10k}. DepthSplat~\cite{xu2025depthsplat} and pixelSplat~\cite{charatan2024pixelsplat} fail to synthesize content outside of the observed regions, as they lack a generative component. While latentSplat~\cite{wewer2024latentsplat} and MVSplat360~\cite{chen2024mvsplat360} can perform extrapolation of novel views beyond the input observations, they fail to generate plausible results when generating target viewpoints far from the observed ones. Most notably, our \ours is not only able to generate new, realistic visual content, but also produces yet-unobserved geometry, which is consistent with the generated appearance. 
\cref{fig:qualitive_dl3dv} further illustrates qualitative comparisons on the more challenging DL3DV-10K. Our \ours produces high-quality renderings from difficult viewpoints, with noticeably fewer artifacts. \cref{fig:qualitive_dl3dv} also highlights the stronger geometric perception of the scene from \ours, reflected in sharper depth estimates, compared to all prior methods, which often yield blurry or inconsistent geometry. This improved geometric understanding directly translates into an increased visual quality of our renderings, particularly in extrapolated regions.

\inparagraph{Assessing Multi-view Consistency.}
As \ours jointly predicts novel views with their corresponding depth maps, we can assess its geometric consistency by unprojecting predictions to 3D. 
As shown in~\cref{fig:pointclouds}, unprojecting predictions from multiple target views produces coherent and well-aligned point clouds on both DL3DV-10K (\cref{fig:pointclouds}, \textit{left} \& \textit{center}) and RE10K (\cref{fig:pointclouds}, \textit{right}), providing evidence of the multi-view consistency of our method. 

\subsection{Analysis and Ablations}
\label{sec:ablations}

\input{tables/table_68_views}
\inparagraph{Number of Context Views.}
\label{main:number_of_views}
Although our \ours is trained with a fixed number of context views ($N_\text{RE10K}=\num{2}$ and $N_\text{DL3DV}=\num{4}$), its architecture naturally supports variable numbers of inputs. In \cref{tab:results_68_views}, we compare against other generative baselines~\cite{wewer2024latentsplat,chen2024mvsplat360} 
under varying input configurations.
Notably, \ours consistently outperforms both baselines and shows clear gains as the number of context views increases ($N_{\uparrow}=6,8$), confirming that our architecture effectively exploits additional geometric evidence from more input observations.
latentSplat~\cite{wewer2024latentsplat} also benefits from more inputs, but scales poorly beyond six views due to the computational overhead of epipolar attention~\cite{he2020epipolar}.

\input{tables/tab_components_cross}

\inparagraph{Cross-dataset Generalization.}
\label{main:cross_data}
We evaluate cross-dataset generalization by directly testing the RE10K model on DL3DV-10K, without additional fine-tuning. We use $N=\num{2}$ and center-crop to a resolution of $256^2$ to match the training resolution of RE10K. 
We adopt the same view-sampling protocol of~\cite{xu2025depthsplat}, which selects views closer to the inputs and thus mimics the narrow-baseline setting on RE10K. As shown in~\cref{tab:cross_dataset}, \ours outperforms other generative baselines, using their RE10K checkpoints when evaluated on DL3DV-10K. 

\inparagraph{Understanding Design Choices.}
\label{app:model_components}
We ablate key components of ReconSplat on RE10K in the extrapolation setting, \textit{before} the fine-tuning stage (\textbf{ft}) that uses $2\times$ bilinearly upsampled inputs (\cf. \cref{sec:ex_setup}).
We study three design choices: conditioning on rasterized   latent features from our 3DGS representation ($\mathcal{G}$), modeling a joint appearance-depth distribution ($\mathbf{D}$), and training with annealed multi-resolution noise ($\boldsymbol{\epsilon}_\text{mr}$). We report ablation results in \cref{tab:re10k_abl_main}.
Conditioning on $\mathcal{G}$ consistently improves all metrics, indicating a better pixel-level agreement and geometric alignment, and confirming the benefit of our dual-stage design. Modeling appearance alone can slightly improve image-wise metrics, as it defines an easier learning problem, under the same model capacity and training budget; however, our joint appearance-depth formulation encourages multi-view geometric consistency, which is not fully captured by these metrics. 
Finally, annealed multi-resolution noise ($\boldsymbol{\epsilon}_\text{mr}$) improves FID while leaving image-wise metrics nearly unchanged, suggesting better perceptual realism, consistent with~\cite{ke2024marigold}.

\inparagraph{3D Convolutions.}
\label{sec:ablating_3dconvs}
For dense camera trajectories, temporal inductive biases can help reduce artifacts such as flickering. 
We therefore add optional 3D convolution blocks ($\mathcal{T}$)~\cite{blattmann2023align} and train them using sequences ($\mathcal{S}$) with $M_{\mathcal{S}}=10$ target frames and $N=2$ context frames, such that $N + M_{\mathcal{S}} = 12$. Training is performed on RE10K for 100K steps, while keeping the base network frozen.
As shown in \cref{tab:ablating_3d_convs}, we ablate this design on 100 RE10K test scenes rendering videos of 50 target frames (inter- or extra-polated). Training only $\mathcal{T}$ blocks matches multi-view consistency of full fine-tuning for the same number of steps, as per MEt3R~\cite{asim2025met3r}, while slightly improving image quality with far fewer trainable parameters ($77$M \vs $\sim$0.9B). These blocks can be enabled for continuous trajectories and omitted for unordered views; a 50-frame video takes $\sim$10s on a single H100 GPU.

%% file: figures/fig_qualitive_re10k.tex
\begin{figure*}[t]
\fontsize{6pt}{7.2pt}
\sffamily
    \centering
    \input{figures/qualitative/qualitative_realestate10k}
    \vspace{-0.75em}
    \caption{\textbf{Qualitative comparison on RE10K} \cite{realestate10k} for the challenging extrapolation setting.
    \ours predicts high-quality novel views and sharp depth estimates for unobserved viewpoints given only two context views, clearly surpassing prior work.
    }
    \label{fig:qualitive_re10k}
    \vspace{-1em}
\end{figure*}
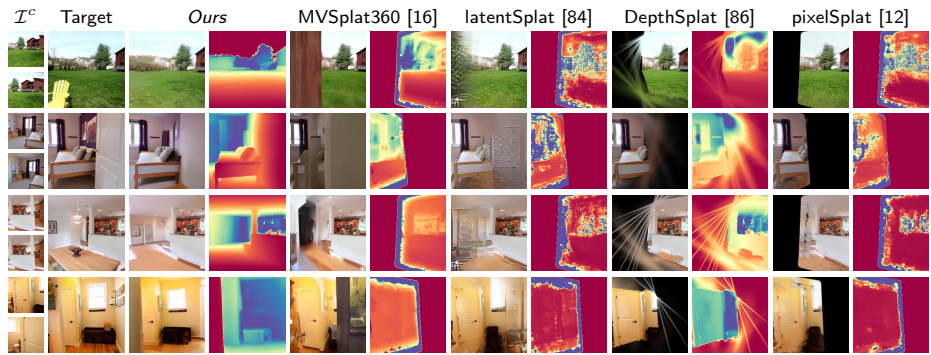

%% file: figures/qualitative/qualitative_realestate10k.tex
\scriptsize
\sffamily
\setlength{\tabcolsep}{1pt}
\renewcommand{\arraystretch}{1.0}

\newcommand{\imgwidth}{0.0814} %

\begin{tabular}{
    >{\centering\arraybackslash}m{0.038\textwidth} %
    >{\centering\arraybackslash}m{\imgwidth\textwidth}
    >{\centering\arraybackslash}m{\imgwidth\textwidth}
    >{\centering\arraybackslash}m{\imgwidth\textwidth}
    >{\centering\arraybackslash}m{\imgwidth\textwidth}
    >{\centering\arraybackslash}m{\imgwidth\textwidth}
    >{\centering\arraybackslash}m{\imgwidth\textwidth}
    >{\centering\arraybackslash}m{\imgwidth\textwidth}
    >{\centering\arraybackslash}m{\imgwidth\textwidth}
    >{\centering\arraybackslash}m{\imgwidth\textwidth}
    >{\centering\arraybackslash}m{\imgwidth\textwidth}
    >{\centering\arraybackslash}m{\imgwidth\textwidth}
}

$\mathcal{I}^c$ 
& Target %
& \multicolumn{2}{c}{\textit{Ours}}
& \multicolumn{2}{c}{MVSplat360~\cite{chen2024mvsplat360}}
& \multicolumn{2}{c}{latentSplat~\cite{wewer2024latentsplat}}
& \multicolumn{2}{c}{DepthSplat~\cite{xu2025depthsplat}}
& \multicolumn{2}{c}{pixelSplat~\cite{charatan2024pixelsplat}} \\[1pt]

\shortstack{\includegraphics[width=\linewidth]{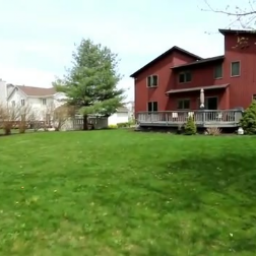}\\[-1pt]
\includegraphics[width=\linewidth]{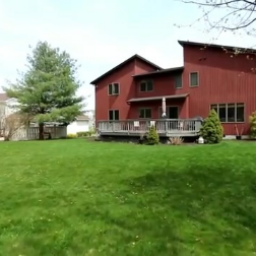}}
& \includegraphics[width=\linewidth]{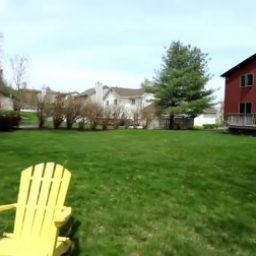} 
& \includegraphics[width=\linewidth]{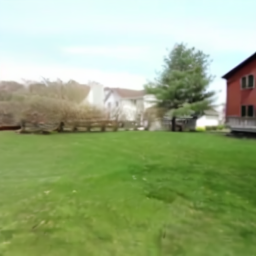} 
& \includegraphics[width=\linewidth]{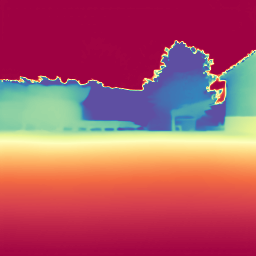}
& \includegraphics[width=\linewidth]{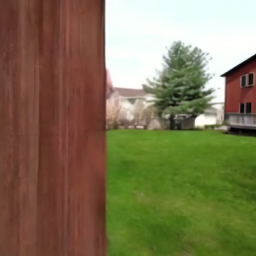}
& \includegraphics[width=\linewidth]{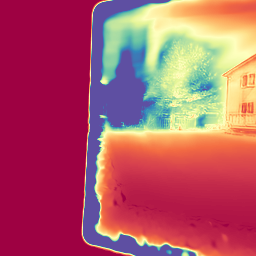} 
& \includegraphics[width=\linewidth]{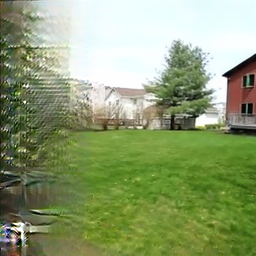} 
& \includegraphics[width=\linewidth]{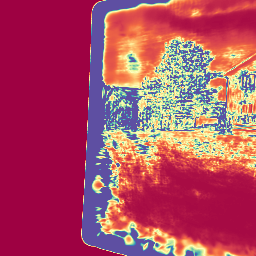} 
& \includegraphics[width=\linewidth]{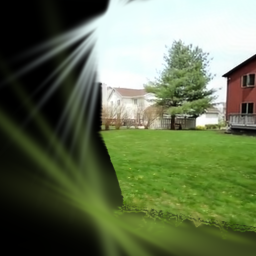} 
& \includegraphics[width=\linewidth]{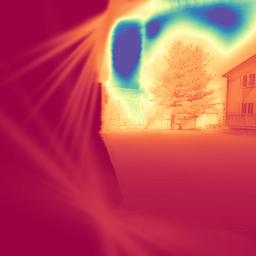} 
& \includegraphics[width=\linewidth]{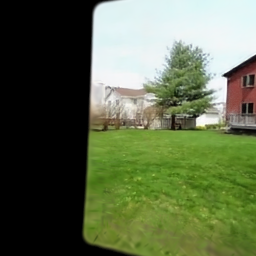} 
& \includegraphics[width=\linewidth]{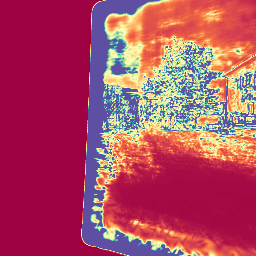} \\

\shortstack{\includegraphics[width=\linewidth]{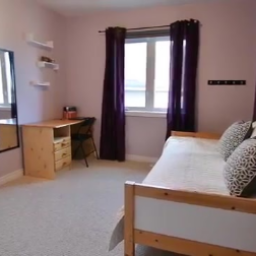}\\[-1pt]
\includegraphics[width=\linewidth]{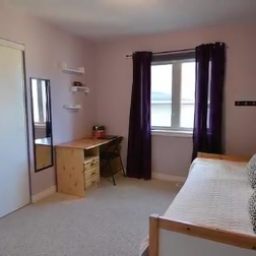}}
& \includegraphics[width=\linewidth]{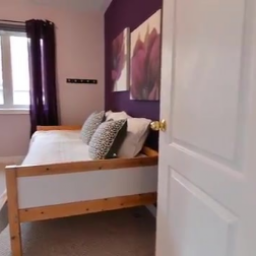} 
& \includegraphics[width=\linewidth]{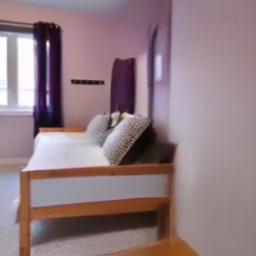}
& \includegraphics[width=\linewidth]{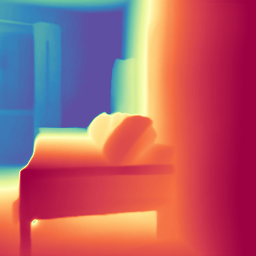}
& \includegraphics[width=\linewidth]{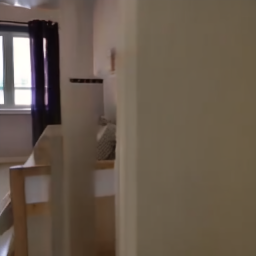} 
& \includegraphics[width=\linewidth]{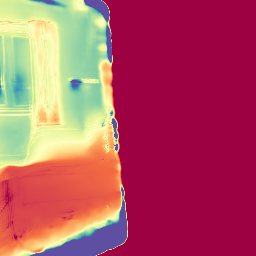} 
& \includegraphics[width=\linewidth]{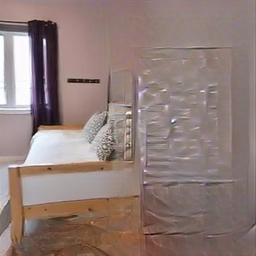} 
& \includegraphics[width=\linewidth]{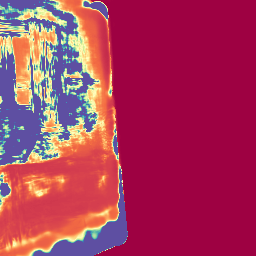} 
& \includegraphics[width=\linewidth]{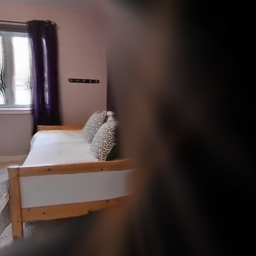} 
& \includegraphics[width=\linewidth]{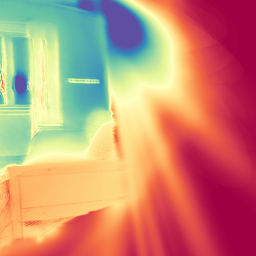} 
& \includegraphics[width=\linewidth]{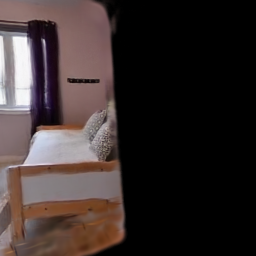} 
& \includegraphics[width=\linewidth]{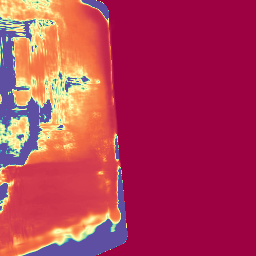} \\

\shortstack{\includegraphics[width=\linewidth]{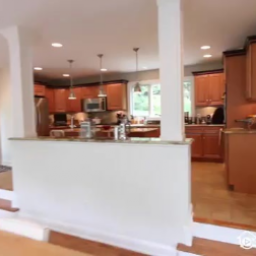}\\[-1pt]
\includegraphics[width=\linewidth]{figures/qualitative/re10k_eccv/row3_old/context_image_000075.png}}
& \includegraphics[width=\linewidth]{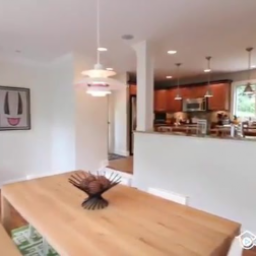} 
& \includegraphics[width=\linewidth]{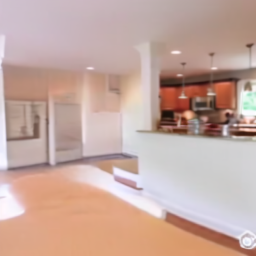}
& \includegraphics[width=\linewidth]{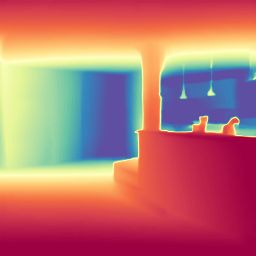}
& \includegraphics[width=\linewidth]{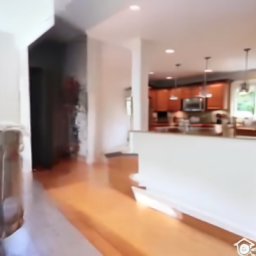} 
& \includegraphics[width=\linewidth]{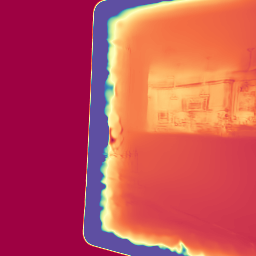} 
& \includegraphics[width=\linewidth]{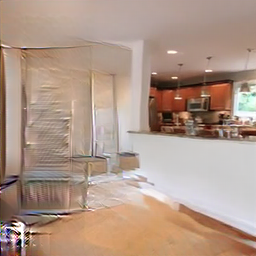} 
& \includegraphics[width=\linewidth]{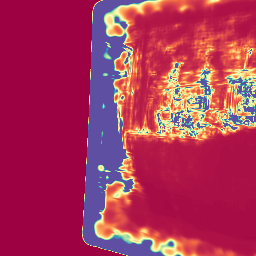} 
& \includegraphics[width=\linewidth]{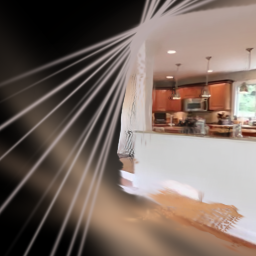} 
& \includegraphics[width=\linewidth]{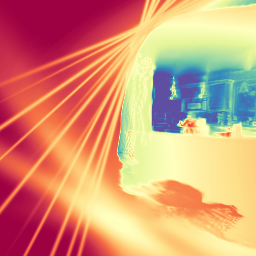} 
& \includegraphics[width=\linewidth]{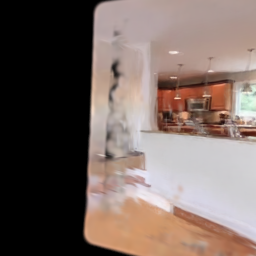} 
& \includegraphics[width=\linewidth]{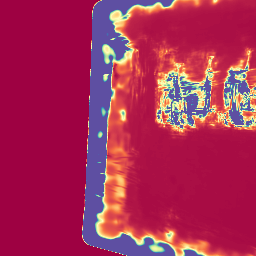} \\

\shortstack{\includegraphics[width=\linewidth]{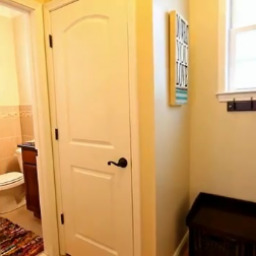}\\[-1pt]
\includegraphics[width=\linewidth]{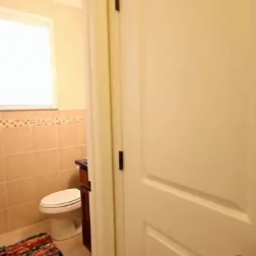}}
& \includegraphics[width=\linewidth]{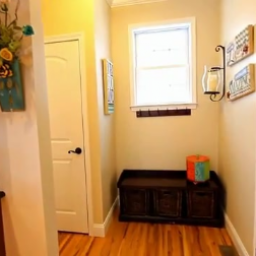} 
& \includegraphics[width=\linewidth]{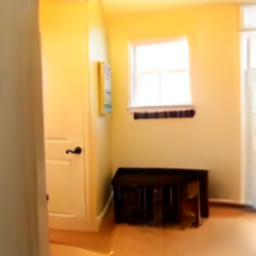} 
& \includegraphics[width=\linewidth]{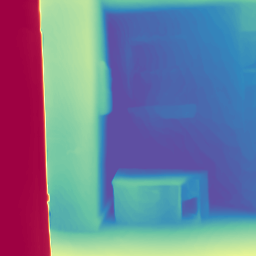}
& \includegraphics[width=\linewidth]{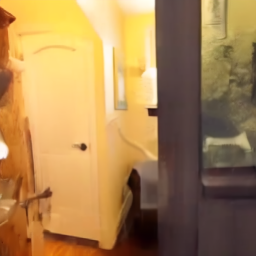}
& \includegraphics[width=\linewidth]{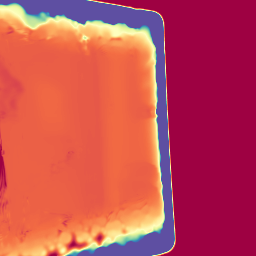} 
& \includegraphics[width=\linewidth]{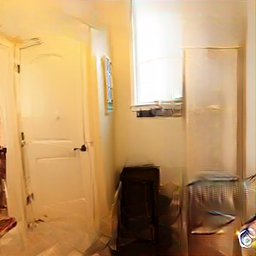} 
& \includegraphics[width=\linewidth]{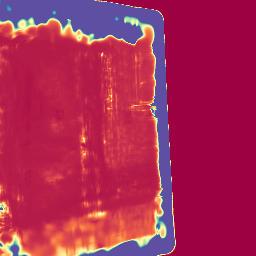} 
& \includegraphics[width=\linewidth]{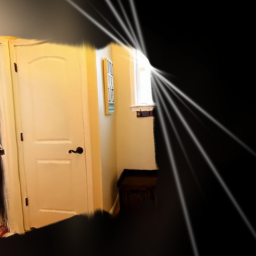} 
& \includegraphics[width=\linewidth]{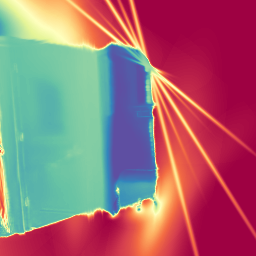} 
& \includegraphics[width=\linewidth]{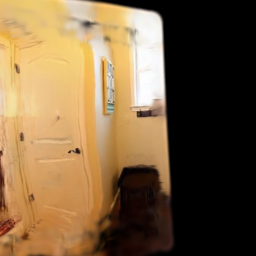} 
& \includegraphics[width=\linewidth]{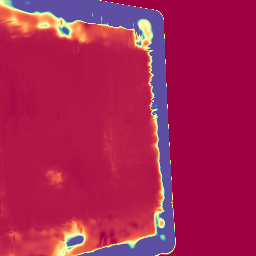} \\

\end{tabular}

%% file: tables/tab_re10k_eccv.tex
\begin{table*}[b]
\vspace{-1em}
\centering
\caption{\textbf{Novel-view synthesis on RE10K.} \ours outperforms both \emph{regression-based} and \emph{generative-based} prior methods, often substantially, in the challenging extrapolation setting (except in SSIM), demonstrating the generalizability of our approach. \ours also outperforms generative approaches \emph{(bottom)} in the interpolation setting across all metrics (except SSIM, where all gen.\ methods are close).}
\label{tab:results_re10k}
\begingroup
\vspace{-0.25em}
\setlength{\tabcolsep}{0.5pt}
\tablesize
\begin{tabularx}{\linewidth}{>{\raggedright\columncolor{white}[\tabcolsep][\tabcolsep]}X
S[table-format=1.2]
S[table-format=1.3]
S[table-format=1.3]
S[table-format=2.2]
S[table-format=1.3]
p{6mm}
S[table-format=1.2]
S[table-format=1.3]
S[table-format=1.3]
S[table-format=2.2]
S[table-format=1.3]} 
\toprule
\raisebox{-2.5pt}[0pt][0pt]{\multirow{2}{*}{\textbf{Method}}} 
& \multicolumn{5}{c}{\textbf{Interpolation}} 
& 
& \multicolumn{5}{c}{\textbf{Extrapolation}} \\ 
\cmidrule(lr){2-6}\cmidrule(lr){8-12}
                        & {FID\,$\downarrow$}  & {LPIPS\,$\downarrow$} & {DISTS\,$\downarrow$} & {PSNR\,$\uparrow$} & {SSIM\,$\uparrow$} && {FID\,$\downarrow$} & {LPIPS\,$\downarrow$} & {DISTS\,$\downarrow$} & {PSNR\,$\uparrow$} & {SSIM\,$\uparrow$}  \\ 
\midrule
\multicolumn{12}{@{}l}{\emph{Regression-based}} \\

pixelSplat~\cite{charatan2024pixelsplat}  &   
4.58   & 0.178    & 0.109   &  24.29   &   0.820  &
&  11.18  &  0.259 & 0.154   & 20.65  &0.728     \\
MVSplat~\cite{chen2024mvsplat}               &    \underline{3.49}    &   0.161  & 0.097  & 24.12  & \underline{0.845}     &
&  9.83   & 0.245     &  0.145     &  20.49    & \underline{0.752}    \\ 
DepthSplat~\cite{xu2025depthsplat}               & \bfseries 2.70   &   \bfseries 0.109  &  \bfseries 0.069     &  \bfseries 27.65     &   \bfseries 0.894 &
&  11.81   & \underline{0.227}  &0.139      &    20.71    & \bfseries0.776 \\
\midrule
\multicolumn{12}{@{}l}{\emph{Generative-based}} \\
latentSplat~\cite{wewer2024latentsplat}             &  3.84   &   0.169 &   0.097  &  23.83    &  0.807   &
& 6.93    &  0.234   &   \underline{0.127}     &  \underline{21.60}    &   0.732     \\
MVSplat360~\cite{chen2024mvsplat360}              &  4.94    &   0.180   &  0.120     &    22.77  &   0.813    &
&  \underline{5.49}   &    0.245  &   0.148    &   20.46   &   0.748     \\
\rowcolor{rowhighlight}
\textbf{ReconSplat} \emph{(Ours)}   &  4.26 &  \underline{0.155}  &   \underline{0.093}     & \underline{24.56}    &  0.808       &&  \bfseries 4.89 &  \bfseries 0.222  & \bfseries 0.120   & \bfseries 21.92 & 0.735      \\
\bottomrule
\end{tabularx}
\endgroup
\end{table*}

%% file: figures/fig_qualitive_dl3dv.tex
\begin{figure*}[t]
    \centering
    \input{figures/qualitative/qualitative_dl3dv}
    \vspace{-0.75em}
    \caption{\textbf{Qualitative comparison on DL3DV-10K} \cite{ling2024dl3dv}.
    \ours predicts high-quality novel views and sharp depth estimates for unobserved viewpoints compared to prior work for the challenging wide-baseline setup on DL3DV-10K given only four context views.
    }
    \label{fig:qualitive_dl3dv}
    \vspace{-1em}
\end{figure*}
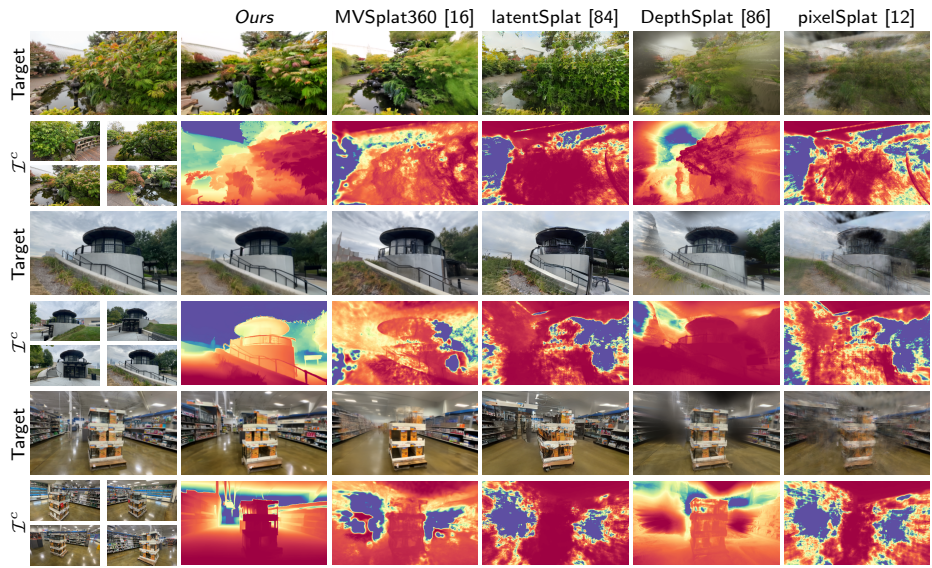

%% file: figures/qualitative/qualitative_dl3dv.tex
\scriptsize
\sffamily
\setlength{\tabcolsep}{1pt}
\renewcommand{\arraystretch}{1.0}
\newcommand{\imgwidthhalf}{0.0715} %
\newcommand{\imgwidth}{0.1575} %

\begin{tabular}{
    >{\centering\arraybackslash}m{0.015\textwidth}
    >{\centering\arraybackslash}m{\imgwidth\textwidth}
    >{\centering\arraybackslash}m{\imgwidth\textwidth}
    >{\centering\arraybackslash}m{\imgwidth\textwidth}
    >{\centering\arraybackslash}m{\imgwidth\textwidth}
    >{\centering\arraybackslash}m{\imgwidth\textwidth}
    >{\centering\arraybackslash}m{\imgwidth\textwidth}
}

& %
& \textit{Ours}
& MVSplat360~\cite{chen2024mvsplat360}
& latentSplat~\cite{wewer2024latentsplat}
& DepthSplat~\cite{xu2025depthsplat}
& pixelSplat~\cite{charatan2024pixelsplat} \\[1pt]

\rotatebox[origin=lB]{90}{\hspace{-0.15em}Target}
& \includegraphics[width=\linewidth]{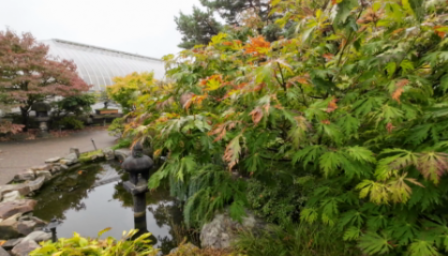} 
& \includegraphics[width=\linewidth]{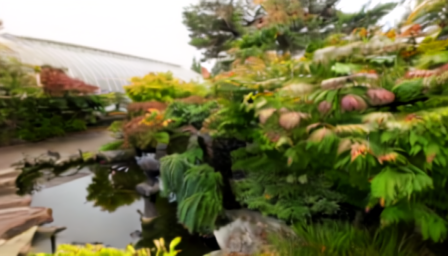} 
& \includegraphics[width=\linewidth]{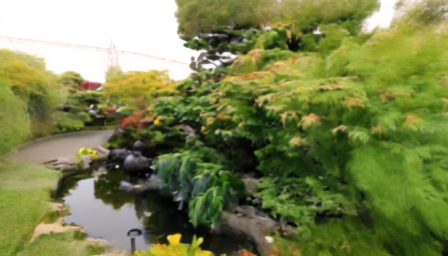} 
& \includegraphics[width=\linewidth]{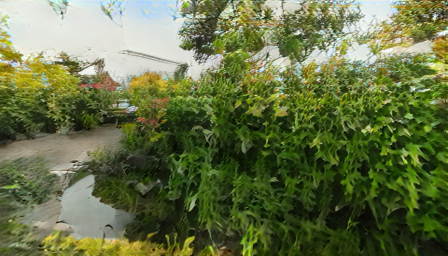} 
& \includegraphics[width=\linewidth]{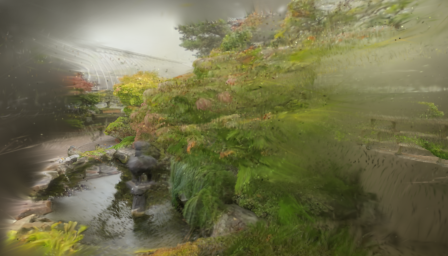}  
& \includegraphics[width=\linewidth]{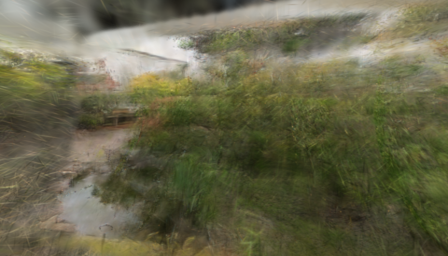}  \\

\rotatebox[origin=lB]{90}{\hspace{0.15em}$\mathcal{I}^c$}
& \shortstack{\includegraphics[width=0.475\linewidth]{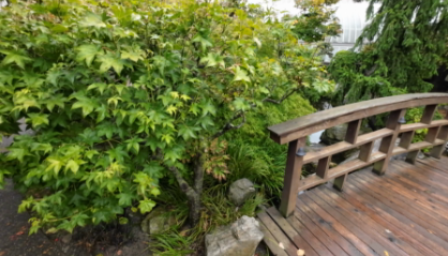}\hspace{0.5pt}
\includegraphics[width=0.475\linewidth]{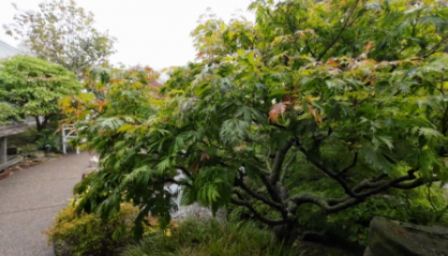}\\[-1pt]
\includegraphics[width=0.475\linewidth]{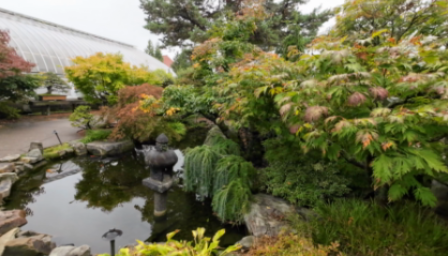}\hspace{0.5pt}
\includegraphics[width=0.475\linewidth]{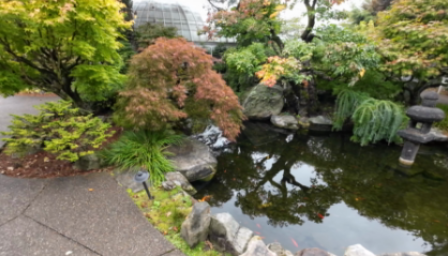}}
& \includegraphics[width=\linewidth]{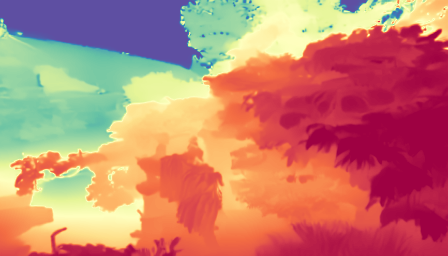} 
& \includegraphics[width=\linewidth]{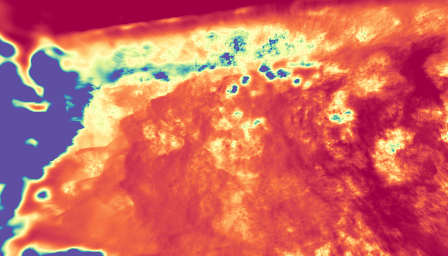} 
& \includegraphics[width=\linewidth]{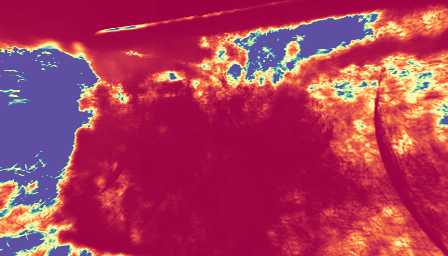} 
& \includegraphics[width=\linewidth]{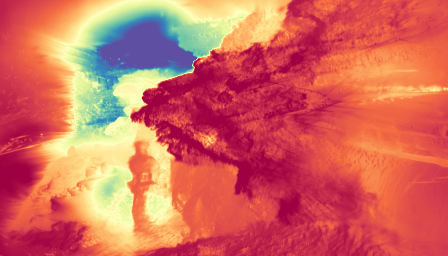}  
& \includegraphics[width=\linewidth]{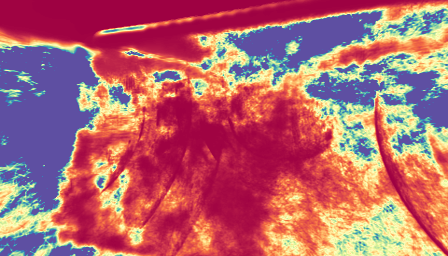}  \\

\rotatebox[origin=lB]{90}{\hspace{-0.15em}Target}
& \includegraphics[width=\linewidth]{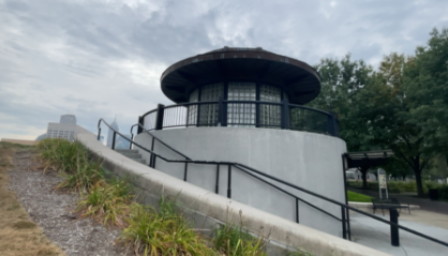} 
& \includegraphics[width=\linewidth]{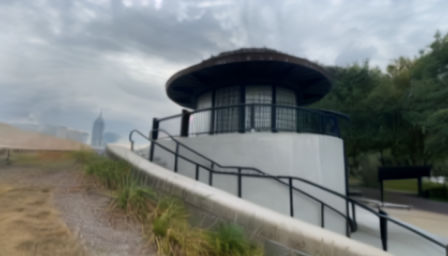} 
& \includegraphics[width=\linewidth]{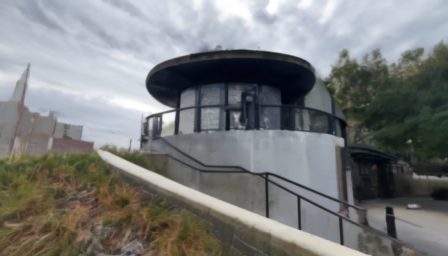} 
& \includegraphics[width=\linewidth]{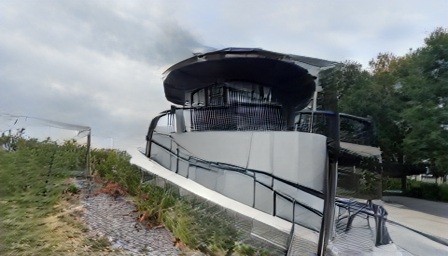} 
& \includegraphics[width=\linewidth]{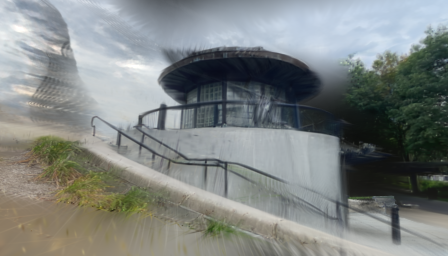}  
& \includegraphics[width=\linewidth]{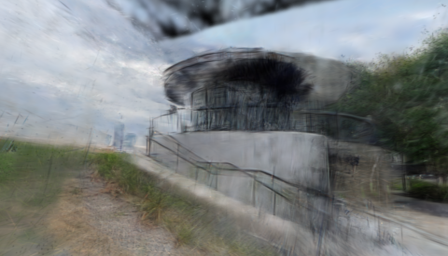}  \\

\rotatebox[origin=lB]{90}{\hspace{0.15em}$\mathcal{I}^c$}
& \shortstack{\includegraphics[width=0.475\linewidth]{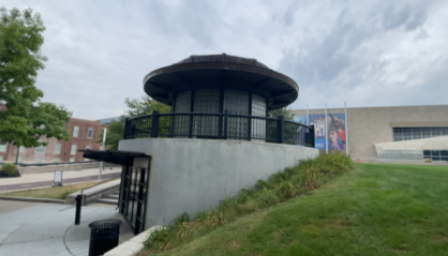}\hspace{0.5pt}
\includegraphics[width=0.475\linewidth]{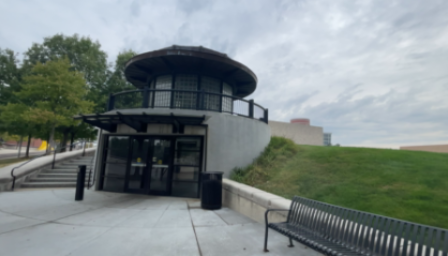}\\[-1pt]
\includegraphics[width=0.475\linewidth]{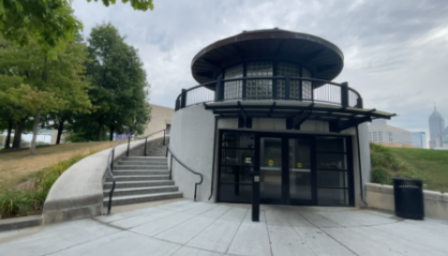}\hspace{0.5pt}
\includegraphics[width=0.475\linewidth]{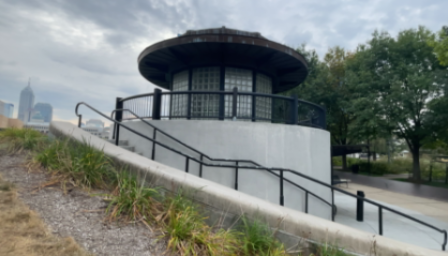}}
& \includegraphics[width=\linewidth]{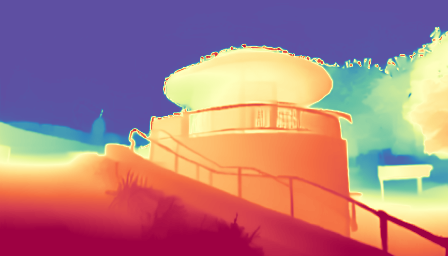} 
& \includegraphics[width=\linewidth]{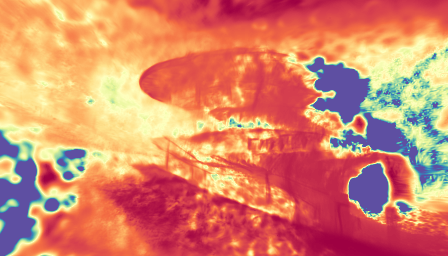} 
& \includegraphics[width=\linewidth]{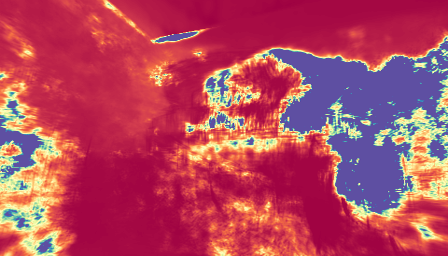} 
& \includegraphics[width=\linewidth]{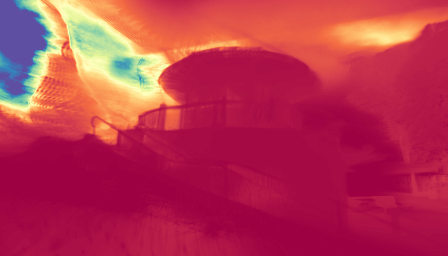}  
& \includegraphics[width=\linewidth]{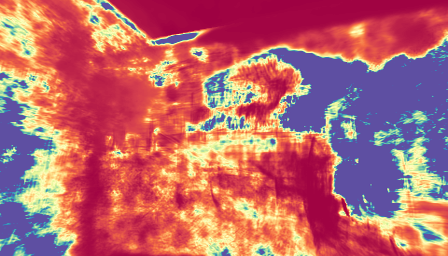}  \\

\rotatebox[origin=lB]{90}{\hspace{-0.15em}Target}
& \includegraphics[width=\linewidth]{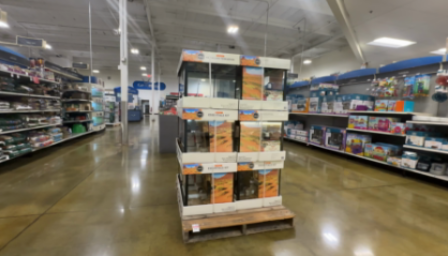} 
& \includegraphics[width=\linewidth]{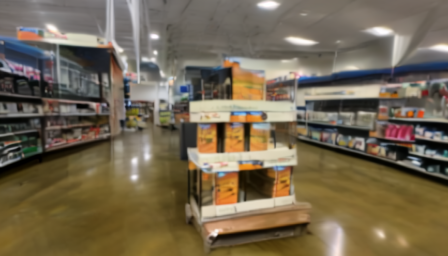} 
& \includegraphics[width=\linewidth]{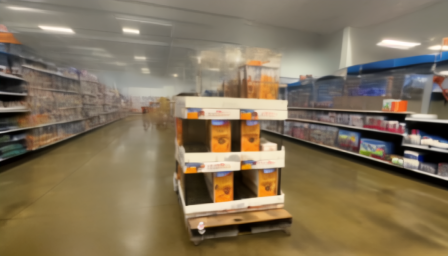} 
& \includegraphics[width=\linewidth]{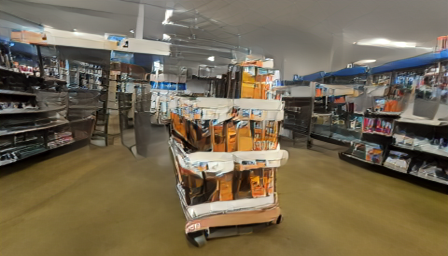} 
& \includegraphics[width=\linewidth]{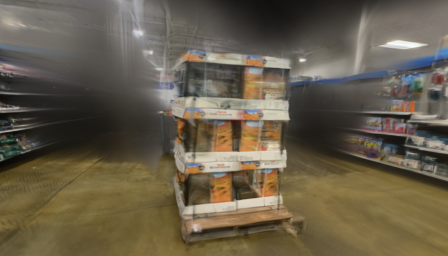}  
& \includegraphics[width=\linewidth]{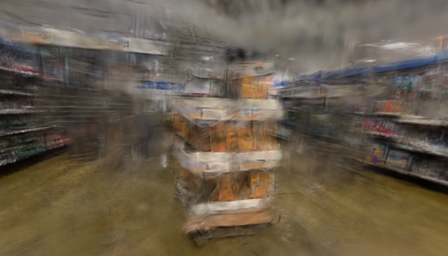}  \\

\rotatebox[origin=lB]{90}{\hspace{0.15em}$\mathcal{I}^c$}
& \shortstack{\includegraphics[width=0.475\linewidth]{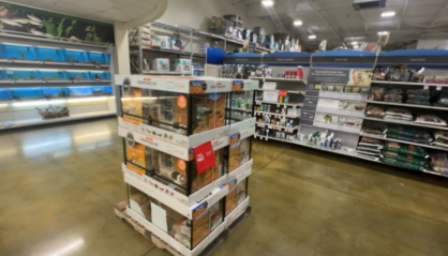}\hspace{0.5pt}
\includegraphics[width=0.475\linewidth]{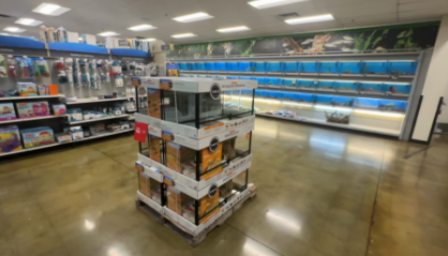}\\[-1pt]
\includegraphics[width=0.475\linewidth]{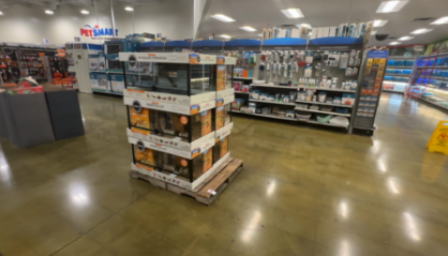}\hspace{0.5pt}
\includegraphics[width=0.475\linewidth]{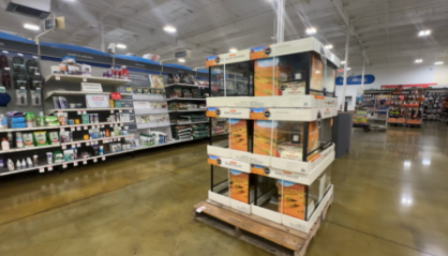}}
& \includegraphics[width=\linewidth]{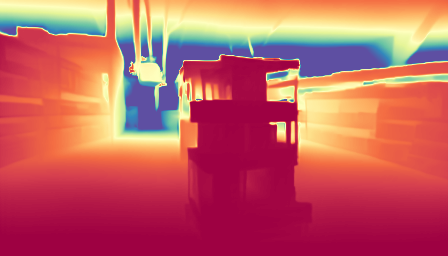} 
& \includegraphics[width=\linewidth]{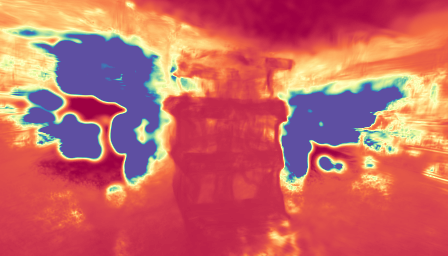} 
& \includegraphics[width=\linewidth]{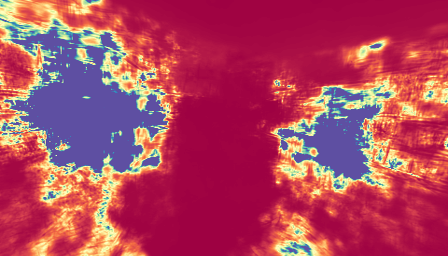} 
& \includegraphics[width=\linewidth]{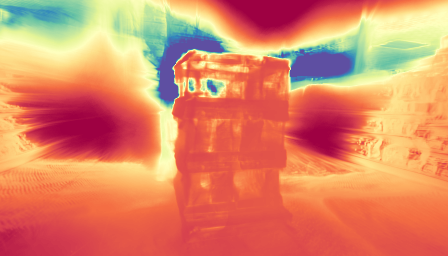}  
& \includegraphics[width=\linewidth]{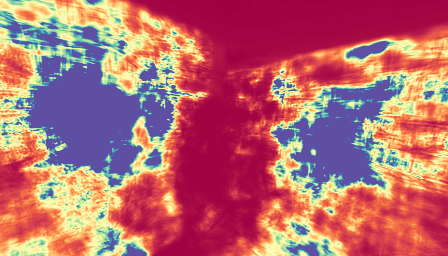}  \\

\end{tabular}

%% file: sec/experimental_setup.tex
\subsection{Experimental Setup}
\label{sec:ex_setup}

\subsubsection{Datasets.}
We conduct experiments on two scene-centric real-world datasets: RealEstate10K (RE10K)~\cite{realestate10k} and the recent DL3DV-10K~\cite{ling2024dl3dv}, using the same training split as MVSplat360~\cite{chen2024mvsplat360}. For evaluation on RE10K, we increase the maximum distance between context and target frames to better assess wide-baseline and far-view extrapolation (\cf \cref{app:sampling_protocol}). For DL3DV-10K, we follow the evaluation protocol of~\cite{chen2024mvsplat360}.

\inparagraph{Training Details.}
\label{training_details}
We first train our model on RE10K for 200\,K steps at $256^2$ resolution, followed by 50\,K fine-tuning steps with $2\times$ bilinearly upsampled inputs. This reduces the effective image-to-latent downscaling factor from $f=8$ to $f=4$, 
while retaining the efficiency of large-batch pre-training on low-resolution latents.
We find that this fine-tuning consistently improves all image-quality metrics. Crucially, as optimization is done in latent space, this strategy ensures that the generated latents remain aligned with the original SD-VAE~\cite{rombach2022high} training resolution. 
A low latent resolution, \ie, $(256/f)^2$, creates a mismatch with the SD-VAE training resolution and degrades decoding quality. Upsampling the inputs effectively mimics a smaller downscaling factor, \ie, $f=4$, mitigating this mismatch, consistent with~\cite{chen2024mvsplat360}.
For DL3DV-10K, we initialize from the RE10K $f=8$ checkpoint, train for 140\,K steps at $256\times480$, and apply the same 50\,K $f=4$ fine-tuning steps. All experiments are conducted on 8$\times$H100 GPUs.

\inparagraph{Baselines.}
\label{baselines}
We conduct a thorough comparison of our proposed \ours against recent feed-forward 3DGS regression-based approaches, including pixelSplat \cite{charatan2024pixelsplat}, MVSplat \cite{chen2024mvsplat}, and  DepthSplat \cite{xu2025depthsplat}, as well as generative methods, including latentSplat \cite{wewer2024latentsplat} and MVSplat360~\cite{chen2024mvsplat360}. For RE10K, we rely on the official checkpoints for all baselines. On DL3DV-10K, we train baselines ourselves whenever no checkpoint is publicly available (\ie, pixelSplat, latentSplat, and MVSplat), initializing from their respective RE10K weights. 
Notably, DepthSplat~\cite{xu2025depthsplat} is trained on the entire DL3DV-10K training split, containing $\sim$5$\times$ more data, and is therefore less directly comparable. For quantitative evaluation, we report standard pixel-aligned and perceptual image quality metrics, including PSNR in dB, SSIM~\cite{ssim}, LPIPS~\cite{lpips}, and DISTS~\cite{dists}. As our method leverages a generative model, we additionally report the Fréchet Inception Distance (FID)~\cite{fid} to measure the realism and distributional alignment of the generated images with respect to the ground-truth data distribution.

%% file: tables/tab_dl3dv_eccv.tex
\begin{table*}[b]
\vspace{-1em}
\centering
\caption{\textbf{Novel-view synthesis on DL3DV-10K.} \ours outperforms both 
\emph{regression-based} and \emph{generative-based} methods on reconstruction, perceptual, and FID metrics, demonstrating its generalizability.  \textbf{Note}: \textcolor{gray}{DepthSplat*} is trained on the full DL3DV-10K dataset ($\sim$5$\times$ more data), making direct comparison less meaningful.} 
\label{tab:results_dl3dv_new}
\begingroup
\sisetup{
  table-align-uncertainty=true,
  separate-uncertainty=true,
}
\setlength{\tabcolsep}{0.5pt}
\vspace{-0.25em}
\tablesize
\begin{tabularx}{\linewidth}{>{\raggedright\columncolor{white}[\tabcolsep][\tabcolsep]}X
S[table-format=2.2]
S[table-format=1.3]
S[table-format=1.3]
S[table-format=2.2]
S[table-format=1.3]
p{6mm}
S[table-format=2.2]
S[table-format=1.3]
S[table-format=1.3]
S[table-format=2.2]
S[table-format=1.3]
}
\toprule
\raisebox{-2.5pt}[0pt][0pt]{\multirow{2}{*}{\textbf{Method}}} 
& \multicolumn{5}{c}{\textbf{Two-round} ($n = 300$)} 
& 
& \multicolumn{5}{c}{\textbf{One round} ($n = 150$)} \\ 
\cmidrule(lr){2-6}\cmidrule(lr){8-12}
 & {FID\,$\downarrow$} & {LPIPS\,$\downarrow$} & {DISTS\,$\downarrow$} & {PSNR\,$\uparrow$} & {SSIM\,$\uparrow$} &
 & {FID\,$\downarrow$} & {LPIPS\,$\downarrow$} & {DISTS\,$\downarrow$} & {PSNR\,$\uparrow$} & {SSIM\,$\uparrow$} \\ 
\midrule
\multicolumn{12}{l}{\emph{Regression-based}}\\
 pixelSplat~\cite{charatan2024pixelsplat}
 & 132.79 & 0.472 & 0.287 &  \underline{17.15} & \underline{0.494} &
 & 108.63 & 0.424 & 0.255 & \underline{18.46} &  \underline{0.551} \\

MVSplat~\cite{chen2024mvsplat}
 & 97.73 & 0.476 & 0.235 & 15.59 & 0.358 &
 & 70.23 & 0.429 & 0.191 & 16.14 & 0.356 \\

\textcolor{gray}{DepthSplat*~\cite{xu2025depthsplat}}
 & \color{gray}55.84 & \color{gray}0.297 & \color{gray}0.155 & \color{gray}19.41 & \color{gray}0.633 &
 & \color{gray}43.67 & \color{gray}0.221 & \color{gray}0.115 & \color{gray}21.51 & \color{gray}0.716 \\

\midrule
\multicolumn{12}{l}{\emph{Generative-based}}\\
latentSplat~\cite{wewer2024latentsplat}
 & 73.08 & \underline{0.411} & \underline{0.182} & 16.32 & 0.472 &
 & 65.04 & \underline{0.363} & \underline{0.164} & 17.50 & 0.535 \\

MVSplat360~\cite{chen2024mvsplat360}
 & \underline{52.86} & 0.461 & 0.191 & 14.17 & 0.332 &
 & \underline{44.16} & 0.419 & 0.170 & 14.84 & 0.346 \\

\rowcolor{rowhighlight}
\textbf{ReconSplat} \emph{(Ours)}
 & \bfseries 43.87 & \bfseries 0.351 & \bfseries 0.137 & \bfseries 17.58 & \bfseries 0.508 &
 & \bfseries 35.83 & \bfseries 0.274 & \bfseries 0.112 & \bfseries 19.71 & \bfseries 0.598 \\

\bottomrule
\end{tabularx}
\endgroup
\end{table*}

%% file: tables/tab_depth_and_cross_data.tex
\begin{figure}[t]
    \centering
    \vspace{-1em}
    \begin{minipage}[t]{0.52\columnwidth}
        \input{tables/scannet_smaller}
    \end{minipage}%
    \hfill%
    \begin{minipage}[t]{0.44\columnwidth}
        \input{tables/tab_cross_dl3dv_from_re10k_smaller}
    \end{minipage}
    \vspace{-1em}
\end{figure}

%% file: tables/scannet_smaller.tex
{
\centering
\captionof{table}{\textbf{Novel-view depth synthesis on ScanNet++}~\cite{yeshwanth2023scannet++}, with DL3DV models. Our generative depth prior yields more accurate depth estimates than baselines.}
\label{tab:scannet}
\vspace{0.6em}
\begingroup 
\sisetup{ 
detect-weight=true, 
detect-inline-weight=math, 
} 
\setlength{\tabcolsep}{2.0pt}
\tablesize
\begin{tabular}{
@{}l
S[table-format=2.2]
S[table-format=1.3]
S[table-format=1.3]
S[table-format=1.3]
@{}}
\toprule
\textbf{Method} 
 &  {$\delta_1\uparrow$} & 
 {AbsRel\,$\downarrow$} & {RMSE\,$\downarrow$} & {RMSE$_\text{log}$\,$\downarrow$} \\ 
\midrule
MVSplat360~\cite{chen2024mvsplat360}  & 42.17    &  0.410 & 0.753 & 0.404  \\
DepthSplat~\cite{xu2025depthsplat}  & \textbf{70.58}   &  \underline{0.255} & \underline{0.492}  & \underline{0.266}    \\
\rowcolor{rowhighlight}
\emph{Ours}  & \underline{67.32} & \bfseries 0.206 & \bfseries 0.467 & \bfseries 0.238   \\ 
\bottomrule
\end{tabular}
\endgroup
}

%% file: tables/tab_cross_dl3dv_from_re10k_smaller.tex
\centering

\captionof{table}{\textbf{Cross-dataset generalization.} \ours outperforms generative baselines when generalizing from RE10K to DL3DV-10K.}
\label{tab:cross_dataset}
\vspace{0.6em}
\sisetup{detect-weight=true, mode=math} 
\setlength{\tabcolsep}{0.5pt}
\tablesize
\begin{tabularx}{\linewidth}{@{}X
    S[table-format=2.2]
    S[table-format=1.3]
    S[table-format=2.2]
    S[table-format=1.3]@{}} 
\toprule 
\textbf{Method} & {FID\,$\downarrow$} & {LPIPS\,$\downarrow$} & {PSNR\,$\uparrow$} & {SSIM\,$\uparrow$} \\ 
\midrule
latentSplat~\cite{wewer2024latentsplat} & 43.03 & 0.214 & 22.21 & 0.735 \\
MVSplat360~\cite{chen2024mvsplat360} & \underline{38.75} & \underline{0.189} & \underline{22.75} & \bfseries 0.786 \\
\rowcolor{rowhighlight}
\emph{Ours} & \bfseries 27.24 & \bfseries 0.150 & \bfseries 23.89 & \underline{0.768} \\ 
\bottomrule
\end{tabularx}

%% file: figures/fig_pc.tex
\begin{figure}[t]
    \centering
    \includegraphics[width=\columnwidth]{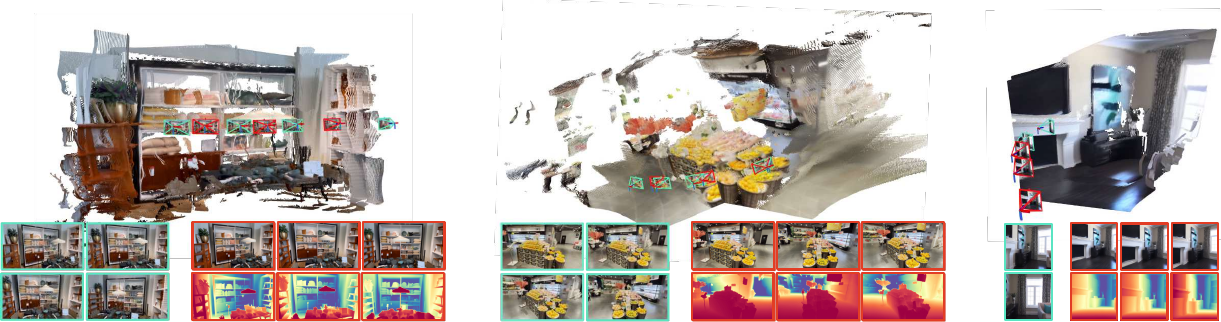}
    \caption{\textbf{Dense point clouds.} Given a set of $N$ context views (\textit{cyan}), \ours predicts $M$
    novel views and their geometry (\textit{red}), for both observed and unobserved regions, as demonstrated by unprojecting depth and appearance into dense point clouds for DL3DV-10K \textit{(left, center)} and for RE10K \textit{(right)}.}
    \label{fig:pointclouds}
    \vspace{-0.5em}
\end{figure}

%% file: tables/table_68_views.tex
\begin{table*}[t]
\centering
\caption{\textbf{Increased number of context views.} On DL3DV-10K, \ours consistently outperforms generative baselines when given additional inputs ($N_{\uparrow}=6,8$), despite being trained with $N_\text{DL3DV}=\num{4}$. OOM denotes ``out of memory''.}
\label{tab:results_68_views}
\begingroup
\sisetup{
  table-align-uncertainty=true,
  separate-uncertainty=true,
}
\vspace{-0.30em}
\setlength{\tabcolsep}{0.5pt}
\tablesize
\begin{tabularx}{\linewidth}{
    X
    S[table-format=2.2]
    S[table-format=1.3]
    S[table-format=1.3]
    S[table-format=2.2]
    S[table-format=1.3]
    p{4mm}
    S[table-format=2.2]
    S[table-format=1.3]
    S[table-format=1.3]
    S[table-format=2.2]
    S[table-format=1.3]
}
\toprule
\raisebox{-2.5pt}[0pt][0pt]{\multirow{2}{*}{\textbf{Method}}}
& \multicolumn{5}{c}{\textbf{6 Context Views}}
& 
& \multicolumn{5}{c}{\textbf{8 Context Views}} \\
\cmidrule(lr){2-6}\cmidrule(lr){8-12}
& {FID$\downarrow$} & {LPIPS$\downarrow$} & {DISTS$\downarrow$} & {PSNR$\uparrow$} & {SSIM$\uparrow$}
&
& {FID$\downarrow$} & {LPIPS$\downarrow$} & {DISTS$\downarrow$} & {PSNR$\uparrow$} & {SSIM$\uparrow$} \\
\midrule
latentSplat~\cite{wewer2024latentsplat}
& 63.87 & \underline{0.341} & \underline{0.158} & \underline{18.05} & \underline{0.571}
&
& \multicolumn{1}{c}{(OOM)} & \multicolumn{1}{c}{(OOM)} & \multicolumn{1}{c}{(OOM)} & \multicolumn{1}{c}{(OOM)} & \multicolumn{1}{c}{(OOM)} \\
MVSplat360~\cite{chen2024mvsplat360}
& \underline{41.80} & 0.403 & 0.162 & 14.98 & 0.348
&
& 41.11 & 0.399 & 0.161 & 14.98 & 0.348 \\
\rowcolor{rowhighlight}
\textbf{ReconSplat} \emph{(Ours)}
& \bfseries 30.13 & \bfseries 0.218 & \bfseries 0.095 & \bfseries 21.55 & \bfseries 0.671
&
& \bfseries 27.40 & \bfseries 0.188 & \bfseries 0.087 & \bfseries 22.71 & \bfseries 0.715 \\
\bottomrule
\end{tabularx}
\endgroup
\end{table*}

%% file: tables/tab_components_cross.tex
\begin{figure}[t]
    \centering
    \vspace{-1em}
    \begin{minipage}[t]{0.48\columnwidth}
        \input{tables/tab_re10k_abl_main}
    \end{minipage}%
    \hfill%
    \begin{minipage}[t]{0.47\columnwidth}
        \input{tables/tab_ablating_3d_convs}
    \end{minipage}
    \vspace{-1em}
\end{figure}

%% file: tables/tab_re10k_abl_main.tex
{
\centering 
\captionof{table}{\textbf{Ablations on RE10K}. We ablate conditioning on 3DGS features ($\mathcal{G}$), learning a generative depth prior ($\mathbf{D}$), and multi-resolution noise ($\boldsymbol{\epsilon}_{\mathrm{mr}}$), before fine-tuning (\textbf{ft}) on $2\times$ upsampled inputs.} 
\label{tab:re10k_abl_main} 
\vspace{0.6em}
\begingroup 
\sisetup{ 
detect-weight=true, 
detect-inline-weight=math, 
} 
\setlength{\tabcolsep}{2.0pt} \renewcommand{\arraystretch}{0.84} \tablesize 
\begin{tabularx}{\linewidth}{
>{\raggedright\arraybackslash}X
S[table-format=2.2]
S[table-format=1.3]
S[table-format=1.3]
S[table-format=1.2]}
\toprule 
\textbf{Variant} & {PSNR$\uparrow$} & {SSIM$\uparrow$} & {LPIPS$\downarrow$} & {FID$\downarrow$} \\ 
\midrule 
w/o ft, $\mathcal{G}$ & 20.75 & 0.693 & 0.249 & 6.07 \\ 
w/o ft, $\mathbf{D}$ & 21.61 & 0.718 & 0.232 & 5.84 \\ 
w/o ft, $\epsilon_{\mathrm{mr}}$ & 21.27 & 0.705 & 0.241 & 6.02 \\
w/o ft & 21.26 & 0.704 & 0.241 & 5.84 \\ 
\midrule 
\rowcolor{rowhighlight}
\textbf{Full} & \bfseries 21.92 & \bfseries 0.735 & \bfseries 0.222 & \bfseries 4.89 \\ 
\bottomrule 
\end{tabularx} 
\endgroup 
}

%% file: tables/tab_ablating_3d_convs.tex
{
\centering 
\captionof{table}{\textbf{Effect of 3D convolutions}. With the same 100K-step fine-tuning budget on sequences $\mathcal{S}$ ($N_{\mathcal{S}}=12$), training only $\mathcal{T}$ blocks matches or slightly outperforms full model fine-tuning, with far fewer updated parameters.} 
\label{tab:ablating_3d_convs} 
\vspace{0.6em}
\begingroup 
\sisetup{ 
detect-weight=true, 
detect-inline-weight=math
} 
\setlength{\tabcolsep}{2.0pt} 
\renewcommand{\arraystretch}{0.84} 
\tablesize 
\begin{tabularx}{\linewidth}{
c
c
>{\raggedright\arraybackslash}X
S[table-format=2.2]
S[table-format=1.3]
S[table-format=1.3]}
\toprule 
$\mathcal{S}$ 
& $\mathcal{T}$ 
& Trainable
& \multicolumn{1}{c}{PSNR$\uparrow$} 
& \multicolumn{1}{c}{LPIPS$\downarrow$} 
& \multicolumn{1}{c}{MEt3R~\cite{asim2025met3r}$\downarrow$} \\ 
\midrule 
\xmark & \xmark & --                 & 20.16           & 0.252           & 0.058 \\
\cmark & \xmark & All                & 20.70           & 0.240           & \bfseries 0.050 \\
\cmark & \cmark & Only $\mathcal{T}$ & \bfseries 20.98 & \bfseries 0.232 & \bfseries 0.050 \\
\bottomrule 
\end{tabularx} 
\endgroup 
}

%% file: sec/conclusion.tex
\section{Conclusion}
We present \ours, a generalizable feed-forward 3D scene reconstruction model that takes an important step toward addressing the fundamental ambiguities of sparse-view 3D reconstruction.
By leveraging 3D Gaussian splatting, we provide a latent variational distribution in 3D that is consistently rasterized into 2D. These geometrically-grounded latents guide the subsequent multi-view diffusion process, enabling the synthesis of geometrically consistent content for unobserved viewpoints. 
This integration allows \ours to perform NVS and depth extrapolation in wide-baseline scenarios while simultaneously producing dense, explicit 3D point clouds in a single feed-forward pass.

\inparagraph{Limitations \& Future Work.} While \ours achieves state-of-the-art novel view synthesis, it relies on sufficiently accurate geometric cues from the feed-forward reconstruction prior and inherits a latent space learned for single-image reconstruction rather than natively for 3D-aware, multi-view generation. Future work could explore stronger 3D reconstruction priors and native multi-view latent representations.

%% file: sec/appendix.tex
\title{\ours: Generalizable 3D Scene Reconstruction Beyond Observed Views\\[6pt]\large -- Supplementary Material --}

\titlerunning{\ours}

\author{Giuseppe Stracquadanio\inst{1}\orcidlink{0009-0005-5086-4856} \and
Kevin Raj\inst{1,2}\orcidlink{0009-0007-3271-5990} \and
Julia Grabinski\inst{1}\orcidlink{0000-0002-8371-1734} \and\\
Stefan Roth\inst{1,2,3}\orcidlink{0000-0001-9002-9832}}

\authorrunning{G.\ Stracquadanio et al.}

\institute{\textsuperscript{1}TU Darmstadt,  \textsuperscript{2}Zuse School ELIZA, 
\textsuperscript{3}hessian.AI \\
\email{\{name.surname\}@visinf.tu-darmstadt.de}}

\maketitle
\renewcommand{\theHsection}{appendix.\Alph{section}}
\setcounter{section}{0}
\renewcommand\thesection{\Alph{section}}
\setcounter{page}{1}
\pagenumbering{roman}
\setcounter{figure}{6}
\setcounter{table}{7}
\setcounter{equation}{8}

In the following, we provide additional information and details that accompany the main paper. This appendix is structured as follows:
\begin{itemize}
    \item \Cref{app:qual} presents extended qualitative comparisons of \ours against other baselines, on both RealEstate10K and DL3DV-10K. We report qualitative results on sparse and unordered target views (\textit{Set} NVS). Results on dense camera trajectories (\textit{Trajectory} NVS), which are better visualized as videos, are available on the \href{https://visinf.github.io/reconsplat/}{project website}, together with interactive point-cloud visualizations. We additionally report a comparison with LVSM~\cite{jin2025lvsm}, a state-of-the-art NVS transformer operating without an explicit 3D representation and directly \textit{regressing} pixels. 
    \item \Cref{app:suppl_experiments} analyzes the effect of the classifier-free guidance scale ($\gamma_\text{CFG}$) and the number of sampling steps (\ie, sampling budget) on image quality metrics.
    \item \Cref{app:rasterizer_math} presents the mathematical details for our modified (\ie variational) 3DGS rasterizer (\cf \cref{sec:mv_latent_field}).
    \item \Cref{app:ex_setup} provides a more detailed description of the experimental setup and additional implementation details.
    \item \Cref{app:limit} discusses the main limitations of our method.
    \item \Cref{app:impact} provides a broader societal impact statement.
\end{itemize}

\section{Additional Results}
\label{app:qual}
\subsubsection{Set NVS.}
We provide additional novel-view and depth synthesis results in \cref{fig:app_qualitive_re10k} for RE10K and in \cref{fig:app_qualitive_dl3dv,fig:app_qualitive_dl3dv2,fig:app_qualitive_dl3dv3,fig:app_qualitive_dl3dv4,fig:app_qualitive_dl3dv5} for DL3DV-10K. We compare with other generative methods, MVSplat360 \cite{chen2024mvsplat360} and latentSplat \cite{wewer2024latentsplat}, focusing on the sparse-view wide-baseline (DL3DV-10K) and extrapolation (RE10K) setups. These qualitative results further demonstrate that our \ours produces high-quality renderings for challenging viewpoints while generating plausible reconstructions for unseen regions. Additionally, \ours is the only generative model compared to our baselines that consistently provides sharp depth estimates due to its geometric consistency.

\inparagraph{Trajectory NVS.}
Side-by-side video comparisons between \ours and other baselines \cite{xu2025depthsplat,chen2024mvsplat360,wewer2024latentsplat} are available on our \href{https://visinf.github.io/reconsplat/}{project website}. We observe that our model can jointly denoise substantially more views (up to $60$ target views at $256^2$ resolution on a  80GB H100 GPU) without a noticeable degradation in quality relative to the sparse target-view configurations used during training. Despite not relying on a large-scale pre-trained video diffusion model~\cite{blattmann2023stable}, \ours produces significantly more temporally consistent novel views compared to MVSplat360~\cite{chen2024mvsplat360}. We attribute the inconsistencies observed in MVSplat360 at every 14 frames to the fixed temporal horizon of the underlying video diffusion model, which is inherently limited to generating sequences of at most 14 frames. While latentSplat~\cite{wewer2024latentsplat} achieves reasonable multi-view consistency with a GAN-based decoder~\cite{gan}, this comes at the cost of overall generation quality. Furthermore, while DepthSplat \cite{xu2025depthsplat} produces high-quality novel views in interpolation settings and benefits from the inherent multi-view consistency of its 3DGS representation, without any generative refinement, its regression-based formulation does not support plausible extrapolation beyond the observed scene content and remains limited in modeling complex geometric structures.

\inparagraph{Point Clouds.} We provide interactive point-cloud visualizations of our scene reconstruction and generation results on our \href{https://visinf.github.io/reconsplat/}{project website}. A stand-alone web viewer is also available \href{https://visinf.github.io/reconsplat/static/pcd/index.html}{here}. 

\inparagraph{Comparison to  LVSM~\cite{jin2025lvsm}.}
We compare \ours against LVSM~\cite{jin2025lvsm}, a state-of-the-art novel-view-synthesis transformer that regresses pixels directly, without an explicit 3D representation. On RE10K extrapolation, LVSM yields better image-quality metrics 
($\mathbf{0.172}$ \vs $0.222$ LPIPS), which favor blurry averages rather than plausible reconstructions. In contrast, \ours achieves lower FID ($5.20$ \vs $\mathbf{4.89}$), indicating stronger scene modeling under uncertainty and \textit{beyond the observed views}.
See \cref{fig:compare_lvsm} for a qualitative comparison.
\input{figures/fig_compare_lvsm}

\section{Additional Analysis}
\label{app:suppl_experiments}
In the following, we further analyze how classifier-free guidance scales and different sampling budgets can affect our results.

\inparagraph{Classifier-free Guidance.}
\Cref{fig:cfg_choice} analyzes the influence of the classifier-free guidance scale $\gamma_\text{CFG}$ on both RE10K and DL3DV-10K. Across both datasets, we observe a consistent pattern: moderate guidance ($\gamma_\text{CFG}=3$) consistently yields the best overall results. This finding is in line with observations from prior work \cite{gao2024cat3d}. Both weak to no guidance (\ie, $\gamma_\text{CFG}\approx1$) and stronger guidance ($\gamma_\text{CFG}\geq5$) lead to noticeable degradation. This effect is most pronounced for perceptual metrics. 
\input{figures/fig_cfg_metrics}
\input{figures/fig_timestep_analysis}

\inparagraph{Sampling Budget.}
Further, we also analyze how a different number of sampling steps affects pixel-aligned metrics and perceptual metrics, by measuring them at different timesteps during sampling and studying their evolution. We conduct this analysis on the DL3DV-10K benchmark.  
As the number of denoising or sampling steps is directly proportional to time and compute resources, it is crucial to assess whether our model consistently improves samples over time, given a fixed sampling budget. 
As shown in \cref{fig:timestep_analysis_dl3dv10k}, when sampling without classifier-free guidance (\textit{i.e.,} $\gamma_\text{CFG} = 1$), distortion metrics such as PSNR and SSIM saturate after only a single sampling step. In contrast, perceptual metrics (LPIPS, DISTS) continue to improve steadily throughout the sampling process. We attribute this behavior to our conditioning scheme with preliminary latents from the 3DGS representation, which allows the model to rapidly capture the coarse geometry and low-frequency structure of the scene. Interestingly, after $\sim20$ sampling steps, we observe a slight decline in distortion metrics, suggesting that additional steps might gradually steer the prediction away from pixel-aligned agreement with the ground truth while improving perceptual quality, revealing a trade-off between distortion and perceptual fidelity. 
When sampling with classifier-free guidance ($\gamma_\text{CFG} > 1$), predictions from the unconditional model are initially detrimental for these metrics in a lower sampling-budget regime, but eventually lead to slightly better results on perceptual metrics when $T=70$. 

\section{Mathematical Details for the Variational Rasterizer}
\label{app:rasterizer_math}
As presented in \cref{sec:ff-3dgs}, we define our 3D representation $\mathcal{G}$ as the union of per-pixel 3D primitives (\ie, 3D Gaussians) predicted from the $N$ context views:
\begin{equation*}
    \mathcal{G}=\big\{(\mathbf{X}_k, \mathbf{\Sigma}_k, \alpha_k, \mathbf{c}_k, \boldsymbol{\mu}^a_k, \boldsymbol{\sigma}_k^a, \boldsymbol{\mu}_k^g, \boldsymbol{\sigma}_k^g)\big\}_{k=1}^{N \cdot H \cdot W}.
\end{equation*}
Beyond the typical Gaussian splat representation \cite{chen2024mvsplat}, each primitive carries two 4-dimensional diagonal Gaussian latent distributions, one for appearance and one for geometry, parameterized by $(\boldsymbol{\mu}_k^a, \boldsymbol{\sigma}_k^a)$ and $(\boldsymbol{\mu}_k^g, \boldsymbol{\sigma}_k^g)$, respectively. Since the same derivation applies independently to the appearance and geometry latents, we use $\ell\in\{a,g\}$ to denote either branch. For each primitive $k$, let 
\begin{equation}
    f_k^\ell(\mathbf{x}) = \mathcal{N}\Big( \mathbf{x};\boldsymbol{\mu}_k^\ell, \diag \big((\boldsymbol{\sigma}_k^\ell)^2\big)\!\Big)
\end{equation} 
be the density of its latent distribution. We omit $\ell$ below for notational simplicity.
The variational parameters $(\boldsymbol{\mu}_k, \boldsymbol{\sigma}_k)$ are predicted \textit{per primitive} by our multi-view encoder, similarly to the encoder of a VAE, which instead predicts \textit{per-pixel} variational parameters in latent space. 

To project this variational representation to pixels of a (target) view, we extend standard front-to-back $\alpha$-compositing~\cite{kerbl20233d} from deterministic attributes to Gaussian distributions. For a target pixel $p$, let the contributing primitives be sorted in front-to-back order, and define their compositing weights as 
\begin{equation} 
    w_{p,k} = \alpha_{p,k} T_{p,k}, \qquad T_{p,k} = \prod_{t<k}(1-\alpha_{p,t}), 
\label{eq:raster_weights} 
\end{equation} 
where $\alpha_{p,k}$ denotes the effective opacity contribution of primitive $k$ at pixel $p$, given its projected footprint. The accumulated opacity at pixel $p$ is then 
\begin{equation} 
    A_p = \sum_k w_{p,k}. 
\end{equation}

Interpreting the rasterized latent as a Gaussian mixture, we obtain the normalized per-pixel density 
\begin{equation} 
 g_p(\mathbf{x}) = \frac{1}{A_p}\sum_k w_{p,k} f_k(\mathbf{x}), \label{eq:raster_mixture} 
\end{equation} 
for pixels with $A_p>0$. Its mean is given by the weighted average of the component means: 
\begin{subequations} \label{eq:mean_explained}
\begin{align} 
\bar{\boldsymbol{\mu}}_p &= \mathbb{E}_{g_p}[\mathbf{x}] \\ &= \frac{1}{A_p}\sum_k w_{p,k}\,\mathbb{E}_{f_k}[\mathbf{x}] \\ &= \frac{1}{A_p}\sum_k w_{p,k}\,\boldsymbol{\mu}_k. \label{eq:raster_mean} 
\end{align}
\end{subequations}

The variance follows from the second raw moment of the mixture. Since the second moment of a diagonal Gaussian is 
\begin{equation} \mathbb{E}_{f_k}[\mathbf{x}^2] = \boldsymbol{\sigma}_k^2 + \boldsymbol{\mu}_k^2, 
\end{equation} 
where squares are applied element-wise, we have 
\begin{subequations} \label{eq:variance_explained} 
\begin{align} 
\bar{\boldsymbol{\sigma}}_p^2 &= \mathbb{E}_{g_p}[\mathbf{x}^2] - \big(\mathbb{E}_{g_p}[\mathbf{x}]\big)^2 \\ &= \frac{1}{A_p}\sum_k w_{p,k}\,\mathbb{E}_{f_k}[\mathbf{x}^2] - \bar{\boldsymbol{\mu}}_p^2 \\ 
&= \frac{1}{A_p}\sum_k w_{p,k} \left(\boldsymbol{\sigma}_k^2 + \boldsymbol{\mu}_k^2\right) - \bar{\boldsymbol{\mu}}_p^2. 
\end{align} 
\end{subequations}

\section{Experimental Setup}
In the following sections, we provide comprehensive details regarding our experimental setup. Our code is available at \href{https://github.com/visinf/reconsplat}{https://github.com/visinf/reconsplat}.
\label{app:ex_setup}
\input{sec/exp_setup_app}

\section{Limitations and Discussions}
\label{app:limit} 

Although our model achieves state-of-the-art results on challenging novel-view and depth synthesis settings as measured by image and depth metrics, it also inherits some limitations from our dual-stage architecture. We describe the main limitation of our \ours in the following.

\inparagraph{Feed-Forward Geometry.} Our architecture conditions the diffusion process on sparse scene reconstructions by a feed-forward multi-view backbone \cite{chen2024mvsplat}. While \ours learns to refine novel views rendered from the Gaussian primitives, its output quality is still tied to the quality of the initial underlying reconstruction. Inaccuracies in the Gaussian primitives, such as floating splats, missing structures, or imprecise geometry in weakly observed regions, can bias the conditioning signal and lead to localized artifacts or view inconsistencies in the final renderings. Improving the robustness of the reconstruction backbone, or jointly refining the underlying scene representation, remains an interesting direction for future work. 

\inparagraph{Latent Representation.} 
Our multi-view latent diffusion model builds on the latent space of a pre-trained Stable Diffusion VAE~\cite{rombach2022high}, which was originally learned for single-image reconstruction rather than 3D-aware, multi-view generation. Although we introduce cross-view communication in the denoising U-Net and condition the model on rasterized 3D Gaussian features, the latent representation itself does not explicitly encode correspondences across views or enforce that the same 3D point has consistent latent structure when observed from different cameras. As a result, multi-view consistency is encouraged by the denoising architecture and 3D conditioning, but is not directly modeled by the latent space. This can limit performance under wide-baseline inputs, strong occlusions, or extrapolated regions requiring hallucinated content across multiple views. Future work could explore, for instance, 3D-aware latent tokenizers trained on multi-view data, or consistent latent representations that preserve geometric correspondences across views. 

\section{Broader Societal Impact}
\label{app:impact}

\ours{}'s dual capabilities, producing plausible novel-view synthesis for unobserved views and precise depth estimation, enable the creation of spatially accurate, navigable 3D digital twins, which offer significant societal benefits across several sectors. Positively, \ours enhances 3D consistency for novel-view and depth synthesis.
However, the model's ability to create views that are plausible but may not represent the actual ground truth presents risks, including the potential for misrepresentation. Therefore, precautions are necessary when using generated views in safety-critical applications like autonomous driving.
Additionally, the model's ability to reconstruct unobserved areas poses a risk of misuse in privacy-intrusive surveillance.
Lastly, the generative capabilities of \ours could be misused to create fake videos.
Responsible deployment demands mandatory transparency measures, such as watermarking all synthesized views \cite{yoo2022deep,zhang2024attack} and implementing anonymization protocols for sensitive data \cite{geppert2021privacy,moon2024efficient}.

\input{figures/fig_app_qualitative_re10k}

\input{figures/fig_app_qualitative_dl3dv}

%% file: figures/fig_compare_lvsm.tex
\begin{figure}[t]
\centering
\scriptsize
\sffamily
\setlength{\tabcolsep}{1pt}
\renewcommand{\arraystretch}{1.0}

\newcommand{\imgwidth}{0.18}

\begin{tabular}{
    >{\centering\arraybackslash}m{0.025\columnwidth}
    >{\centering\arraybackslash}m{\imgwidth\columnwidth}
    >{\centering\arraybackslash}m{\imgwidth\columnwidth}
    >{\centering\arraybackslash}m{\imgwidth\columnwidth}
    >{\centering\arraybackslash}m{\imgwidth\columnwidth}
    >{\centering\arraybackslash}m{\imgwidth\columnwidth}
}

\rotatebox[origin=lB]{90}{\hspace{-0.15em}GT}
& \includegraphics[width=\linewidth]{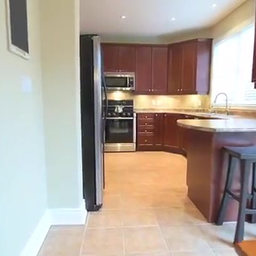}
& \includegraphics[width=\linewidth]{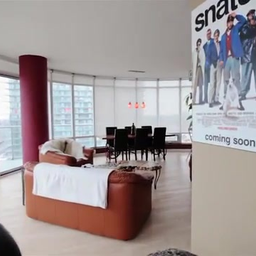}
& \includegraphics[width=\linewidth]{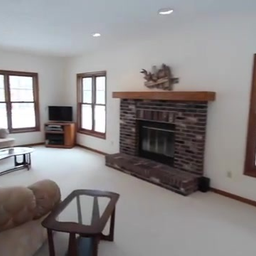}
& \includegraphics[width=\linewidth]{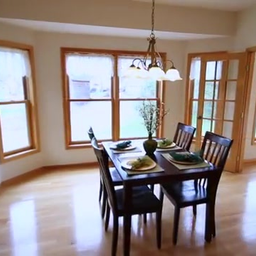}
& \includegraphics[width=\linewidth]{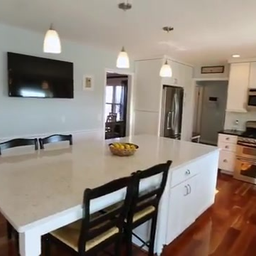} \\

\rotatebox[origin=lB]{90}{\hspace{-0.15em}LVSM}
& \includegraphics[width=\linewidth]{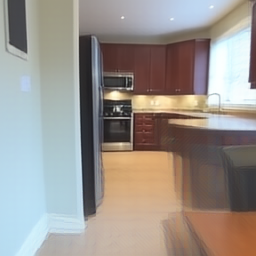}
& \includegraphics[width=\linewidth]{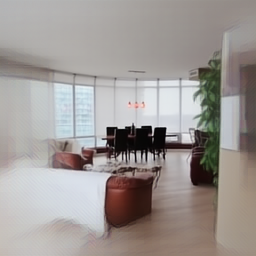}
& \includegraphics[width=\linewidth]{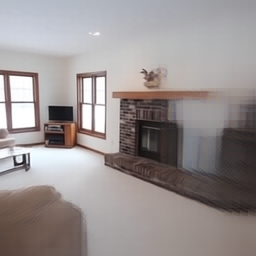}
& \includegraphics[width=\linewidth]{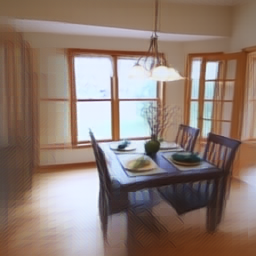}
& \includegraphics[width=\linewidth]{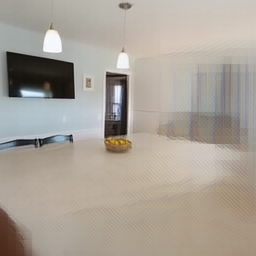} \\

\rotatebox[origin=lB]{90}{\hspace{-0.15em}Ours}
& \includegraphics[width=\linewidth]{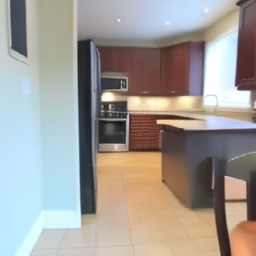}
& \includegraphics[width=\linewidth]{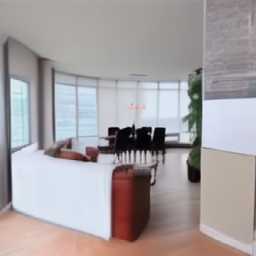}
& \includegraphics[width=\linewidth]{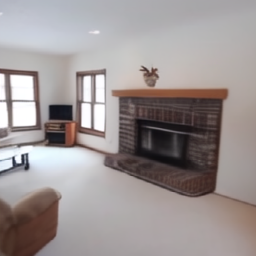}
& \includegraphics[width=\linewidth]{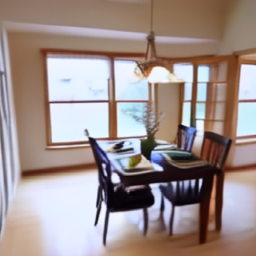}
& \includegraphics[width=\linewidth]{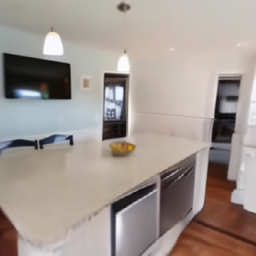}

\end{tabular}
\vspace{-1em}
\caption{\textbf{Qualitative comparison with LVSM~\cite{jin2025lvsm}}.}
\label{fig:compare_lvsm}
\vspace{-1.5em}
\end{figure}

%% file: figures/fig_cfg_metrics.tex
\begin{figure}[t]
    \centering
    \includegraphics[width=\columnwidth]{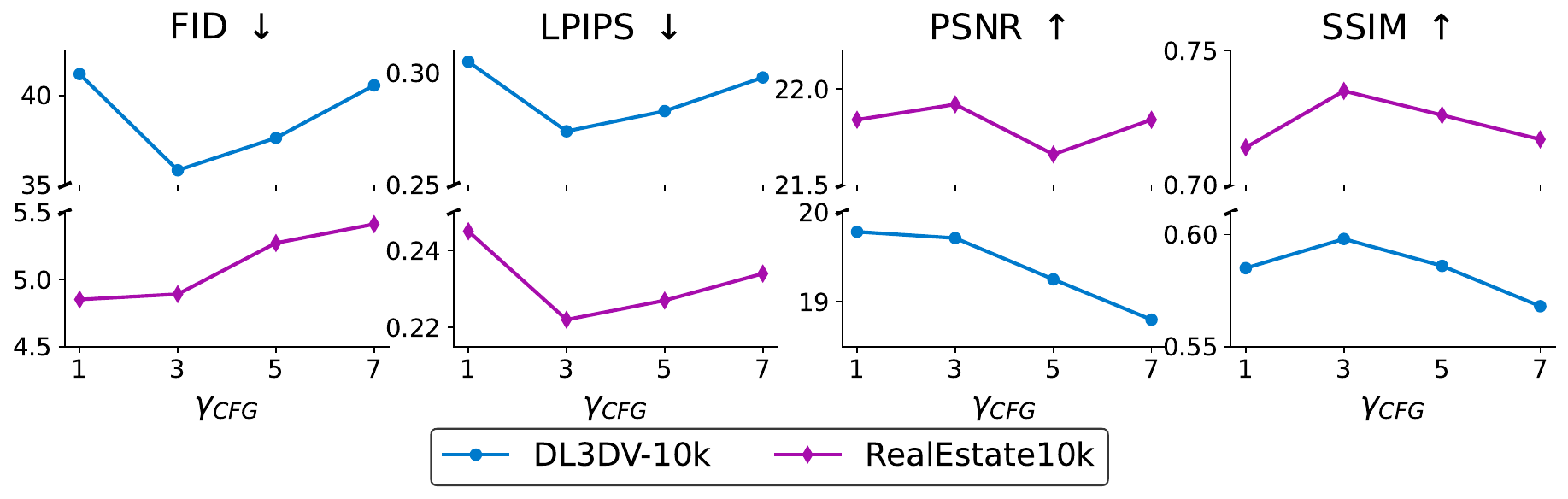}
    \caption{\textbf{Influence of classifier-free guidance.} Metrics over different choices of $\gamma_\text{CFG}$ for DL3DV-10K \cite{ling2024dl3dv} ($n=150$) and RealEstate10K \cite{realestate10k} (Extrapolation). Fixing $\gamma_\text{CFG} = 3$ yields improvements across all metrics and datasets.}
    \label{fig:cfg_choice}
\end{figure}

%% file: figures/fig_timestep_analysis.tex
\begin{figure}[t]
    \centering
    \includegraphics[width=\columnwidth]{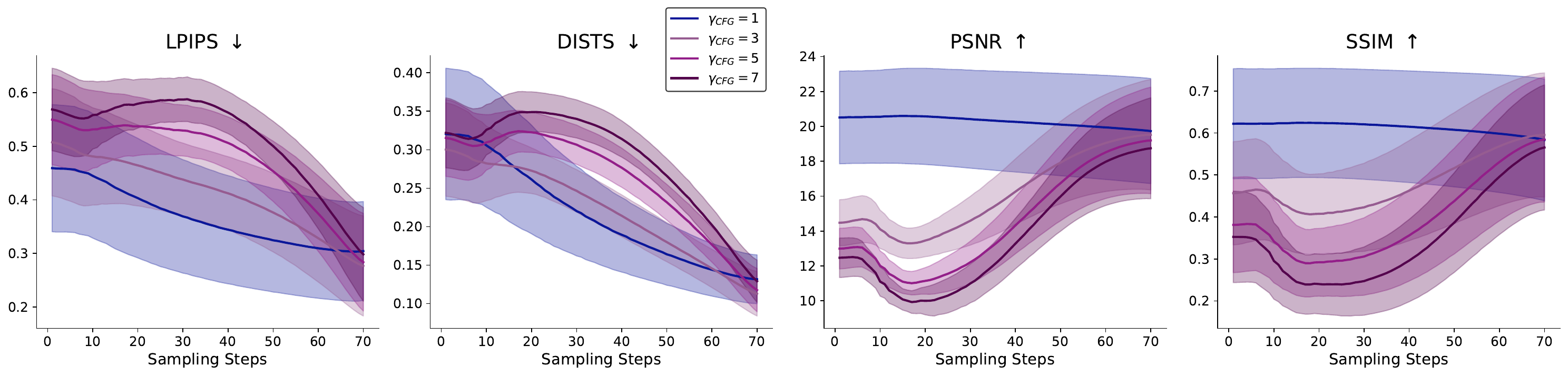}
    \caption{\textbf{Influence of sampling budget.} Distortion and perceptual metrics given different sampling budgets for DL3DV-10K.}
    \label{fig:timestep_analysis_dl3dv10k}
\end{figure}

%% file: sec/exp_setup_app.tex
\subsection{Datasets}
\label{app:datasets}
To evaluate ReconSplat, we conduct experiments on two scene-centric real-world datasets: RealEstate10K (RE10K)~\cite{realestate10k}, which primarily consists of indoor walk-through videos collected from YouTube, and the DL3DV-10K dataset~\cite{ling2024dl3dv}, which collects a diverse range of indoor and unbounded outdoor scenes. For experiments on RE10K, we adopt the official training/test split, in line with previous work \cite{charatan2024pixelsplat, wewer2024latentsplat,chen2024mvsplat, chen2024mvsplat360,xu2025depthsplat}, consisting of \num{67477} scenes for training and \num{7289} for testing. For DL3DV-10K, we follow the evaluation protocol introduced by \cite{chen2024mvsplat360} to ensure fair comparisons. Specifically, we construct our training set by combining the ``3K'' and ``4K'' splits, including \num{1581} scenes. We evaluate on the official DL3DV-10K Benchmark split \cite{ling2024dl3dv}, which consists of \num{140} scenes that are excluded from the training set to prevent data leakage. We perform our evaluation under the same settings of \cite{chen2024mvsplat360}: $n\!=\!\num{150}$ and $n\!=\!\num{300}$, where $n$ is the frame-distance span across sampled test views. Since most DL3DV-10K scenes contain two-round camera trajectories, restricting the frame distance to $n\!=\!\num{150}$ represents a simple heuristic to ensure covering of a single camera loop. Thus, in the main paper, we refer to the settings $n\!=\!\num{150}$ and $n\!=\!\num{300}$ as ``One round'' and ``Two round'' (\cf \Cref{tab:results_dl3dv_new}), respectively. 

\subsection{Data Pre-processing}
\label{app:data_preprocessing}
Our method requires paired multi-view images and dense depth maps for supervision. We rely on the state-of-the-art dense feed-forward reconstruction method VGGT \cite{wang2025vggt} to process each scene from RE10K and our DL3DV-10K training split, yielding images, (relative) depth maps, and corresponding camera poses in a shared, normalized coordinate frame. To further improve the quality of the relative depth maps before using them as pseudo-labels for training, particularly for the challenging DL3DV-10K dataset, we employ Video Depth Anything (VDA) \cite{chen2025video}, a state-of-the-art video depth estimator with strong domain-generalization capabilities. We align VDA predictions to the VGGT reference scale by fitting a single global scale and shift in disparity space, \ie, on inverse depth, using robust Huber regression over confidence-filtered pixels pooled from all frames of a scene. The aligned disparities are then inverted back to depth.
Before encoding to latents with the SD-VAE, depth maps are normalized per scene to $[-1,1]$ using the $2$nd and $98$th percentiles of valid depth values, with values outside that range clipped. 

\subsection{View-Sampling Protocol} 
\label{app:sampling_protocol}
For RE10K, we train with $N_\text{RE10K}=\num{2}$ context views and $M_\text{RE10K}=\num{4}$ target views. For DL3DV-10K, we use $N_\text{DL3DV}=\num{4}$ and $M_\text{DL3DV}=\num{4}$. We use the same setup for quantitative comparisons with our baselines. However, this is not a strict requirement, and our \ours can leverage a larger number of context views and jointly generate many more target views at once, \eg, for rendering a video of 50 frames (\cf \cref{sec:ablating_3dconvs}).

To sample input and target views for training and evaluation, we follow prior work \cite{charatan2024pixelsplat,chen2024mvsplat, chen2024mvsplat360,xu2025depthsplat} and use heuristics based on the temporal distance between frames in the video. 
More specifically, during training on RE10K, we sample input views and target views with the following criteria based on the distance between frame indices. Let $f_L$ and $f_R$ denote the indices of the two context frames (left and right, respectively), and $f^{\odot}$ the index of a target frame.
\begin{enumerate}
    \item We sample two input frames $f_L$ and $f_R$, $s.t.$ $\num{45} < f_R - f_L < \num{90}$.
    \item We sample each target frame $f^{\odot}$, $s.t.$ $f_L \num{-90} < f^{\odot} < f_R + \num{90}$.
\end{enumerate}

\smallskip
During an initial warm-up stage, we progressively increase both the distance between inputs and the maximum distance of the targets from the inputs, until reaching the described criteria. For evaluation, we enforce $f_L < f^{\odot} < f_R$ for the interpolation setup and $\text{max}(f_L - f^{\odot}, f^{\odot} - f_R) < \num{90}$ for extrapolation.
For training and evaluation on DL3DV-10K, we follow \cite{chen2024mvsplat360} and sample context views by applying farthest-point sampling to the set of target camera poses, ensuring good spatial coverage. The unused views are then randomly sampled as target views. 

\subsection{Training Details}
\label{app:training_protocol}

\begin{table}[t!]
\centering
\small
\caption{
\textbf{Stage 1 training configuration.} (a) Training setup for RE10K and DL3DV, including initialization, output resolution, number of context and target views, batch size, and optimization settings. (b) Loss terms and corresponding weights for the auxiliary Gaussian rendering $\mathcal{L}_{\text{aux}}$, appearance $\mathcal{L}_{\text{a}}$, geometry $\mathcal{L}_{\text{g}}$, and latent-regularization objectives. $\dagger$ The encoder's patch shim requires both image dimensions to be multiples of $64$
(patch size $16$ $\times$ downscale $4$). DL3DV inputs are therefore center-cropped from
$256\times480$ to $256\times448$, with intrinsics rescaled accordingly. $\ddagger$ We set the KL weight to $10^{-6}$, following Stable Diffusion / LDM $f=8$ autoencoder~\cite{rombach2022high}.}
\label{tab:stage1_config}

\textbf{(a) Training setup}
\par\vspace{2pt}
\begin{tabular*}{\columnwidth}{@{\extracolsep{\fill}}lcc@{}}
\toprule
Setting & RE10K & DL3DV \\
\midrule
\multicolumn{3}{@{}l}{\emph{Data and views}} \\
\quad Init.\ from       
    &   MVSplat~\cite{chen2024mvsplat} 
    & RE10K ckpt. \\
\quad Resolution           
    & $256^2$       
    & $256\times448^{\dagger}$ \\
\quad Context views ($N$)    
    & 2             
    & 4 \\
\quad Target views ($M$)     
    & 4             
    & 4 \\
\quad Effective batch size ($B_{\text{eff}}$) & 24            & 8 \\
\addlinespace[3pt]
\multicolumn{3}{@{}l}{\emph{Optimization}} \\
\quad Training steps       
    & $100\mathrm{K}$ 
    & $100\mathrm{K}$ \\
\quad Precision            
    & \texttt{fp32}          
    & \texttt{fp32} \\
\quad Learning rate        
    & $2\times10^{-4}$ 
    & $2\times10^{-4}$ \\
\quad Learning rate schedule         
    & cosine        
    & cosine \\
\quad Warm-up              
    & $2\mathrm{K}$ 
    & $2\mathrm{K}$ \\
\quad Gradient clipping    
    & 0.5           
    & 0.5 \\
\bottomrule
\end{tabular*}

\vspace{2em}

\textbf{(b) Loss weights}
\par\vspace{2pt}
\begin{tabular*}{\columnwidth}{@{\extracolsep{\fill}}llcc@{}}
\toprule
Loss & Term & RE10K & DL3DV \\
\midrule

$\mathcal{L}_{\mathrm{aux}}$
  & $\ell_2$                         & 10.0 & 10.0 \\

  & LPIPS                            & 0.5  & 0.5 \\

\addlinespace[3pt]

$\mathcal{L}_a$
  & $\ell_1$                         & 1.0  & 1.0 \\

  & LPIPS $(\omega_{\mathrm{per}})$  & 0.5  & 1.0 \\
  & Adversarial, gen. $(\omega_{\mathrm{adv}})$               & 0.75 & 0.75 \\
  & Adversarial, disc.               & 1.0  & 1.0 \\

\addlinespace[3pt]

$\mathcal{L}_g$
  & Charbonnier $\rho_\alpha$        & 1.0  & 0.5 \\

  & Gradient $(\omega_{\mathrm{gm}})$ & 5.0 & 10.0 \\
  & $1-\mathrm{SSIM}$ $(\omega_{\mathrm{dis}})$
                                      & 1.0  & 1.0 \\
\addlinespace[3pt]
Latent reg.
  & KL, appearance $(\omega_{d,a})$ and geometry
    $(\omega_{d,g})$              & $10^{-6\,\ddagger}$ & $10^{-6\,\ddagger}$ \\

\bottomrule
\end{tabular*}
\end{table}

\begin{table*}[t!]
\centering \small
\caption{\textbf{Stage 2 configuration.} (a) Shared optimization, diffusion-training, and inference settings. (b) Checkpoint-specific configurations for RE10K and DL3DV. $\dagger$ The encoder's patch shim requires both image dimensions to be multiples of $64$
(patch size $16$ $\times$ downscale $4$). DL3DV inputs are therefore center-cropped from
$256\times480$ to $256\times448$, with intrinsics rescaled accordingly. $\ddagger$ ``Only $\mathcal{T}$'' denotes fine-tuning only the 3D convolution blocks,
with all remaining parameters frozen.}
\label{tab:stage2_config}

\textbf{(a) Shared (RE10K \& DL3DV) optimization and diffusion settings}
\par\vspace{2pt}
\begin{tabular*}{\textwidth}{@{\extracolsep{\fill}}ll@{}}
\toprule
Setting & Value \\
\midrule

\multicolumn{2}{@{}l}{\emph{Optimization}} \\
\quad Precision
    & \texttt{fp32} \\
\quad Learning rate
    & $5\!\times\!10^{-5}$ \\
\quad Learning rate schedule
    & constant \\
\quad Warm-up
    & $1\text{K}$ steps, linear \\
\quad Gradient clipping
    & $0.5$ \\
\addlinespace[3pt]
\multicolumn{2}{@{}l}{\emph{Diffusion training}} \\
\quad Forward process
    & DDPM~\cite{ho2020ddpm} \\
\quad Diffusion timesteps 
    & $1000$ \\
\quad Parameterization
    & $v$-prediction~\cite{salimans2022v} \\
\quad Noise schedule
    & scaled linear, $\beta \in [8.5\!\times\!10^{-4},\,1.2\!\times\!10^{-2}]$ \\
    
\addlinespace[3pt]
\multicolumn{2}{@{}l}{\emph{Sampling}} \\
\quad Solver
    & DPM-Solver++~\cite{lu2025dpmsolverpp}\\
\quad Solver steps
    & $70$ \\
\quad CFG Scale ($\gamma_{\text{CFG}}$)
    & $3.0$ \\

\bottomrule
\end{tabular*}

\vspace{2em}

\textbf{(b) Checkpoint-specific training configuration}
\par\vspace{2pt}
\begin{tabular*}{\textwidth}{@{\extracolsep{\fill}}lcrrrrrc@{}}
\toprule
Checkpoint
    & Init.\ from
    & Latent res.
    & $N$
    & $M$
    & $B_{\mathrm{eff}}$
    & Steps
    & Trainable \\
\midrule
\multicolumn{8}{@{}l}{\emph{RE10K}, $256^2$} \\
Base ($\mathsf{B}_{\text{RE10K}}$)
    & SD 2.1~\cite{rombach2022high}
    & $32^2$
    & 2 & 4 & 80
    & $200\text{K}$
    & All \\
\quad$\hookrightarrow$ $+$ $f=8\rightarrow4$
    & $\mathsf{B}_{\text{RE10K}}$
    & $64^2$
    & 2 & 4 & 24
    & $50\text{K}$
    & All \\
\quad$\hookrightarrow$ $+$ Video ($\mathcal{S}$)
    & $\mathsf{B}_{\text{RE10K}}$
    & $32^2$
    & 2 & 10 & 32
    & $100\text{K}$
    & Only $\mathcal{T}^{\ddagger}$ \\
\addlinespace[3pt]
\midrule
\multicolumn{8}{@{}l}{\emph{DL3DV}, $256\times448^{\dagger}$} \\
Base ($\mathsf{B}_{\text{DL3DV}}$)
    & $\mathsf{B}_{\text{RE10K}}$
    & $32\times56$
    & 4 & 4 & 32
    & $140\text{K}$
    & All \\
\quad$\hookrightarrow$ $+$ $f=8\rightarrow4$
    & $\mathsf{B}_{\text{DL3DV}}$
    & $64\times112$
    & 4 & 4 & 8
    & $50\text{K}$
    & All \\
\quad$\hookrightarrow$ $+$ Video ($\mathcal{S}$)
    & $\mathsf{B}_{\text{DL3DV}}$
    & $32\times56$
    & 4 & 10 & 24
    & $100\text{K}$
    & Only $\mathcal{T}^{\ddagger}$ \\
\bottomrule
\end{tabular*}
\end{table*}

Our training proceeds in two stages. First, we train our feed-forward 3DGS backbone based on \cite{chen2024mvsplat}, with the full training objective described in \cref{subsec:learning_mv_latent_features} to learn the associated multi-view latent field (\cf~\cref{sec:ff-3dgs}). The corresponding training configurations are reported in~\cref{tab:stage1_config}. For rasterization, we use our custom CUDA rasterizer, based on \cite{wewer2024latentsplat}, which we modify to allow for rasterization of variational parameters (\cf \cref{sec:mv_latent_field}). We then train our multi-view latent diffusion model (MV-LDM) as described in the main paper (\cf~\cref{sec:ex_setup}), conditioning the denoising process on the preliminary latents sampled from the 2D variational distributions rasterized at the target views (\cf~\cref{subsec:mv_latent_diffusion}). Configurations and settings for our checkpoints are reported in ~\cref{tab:stage2_config}. All experiments are conducted on 8$\times$ H100 GPUs. 

\subsection{Implementation Details}

\label{app:implementation}
\subsubsection{Model Initialization.} We initialize our diffusion U-Net~\cite{ronneberger2015unet} with the official Stable Diffusion v2.1 \cite{rombach2022high} checkpoint. We rely on the same checkpoint to initialize the weights of our adapted \cite{hu2022lora} VAE decoders. 

\inparagraph{First-Stage Decoder Adaptation.} We adapt the first-stage autoencoder using parameter-efficient LoRA fine-tuning~\cite{hu2022lora} applied only to the VAE decoders. The LoRA adapters are inserted into every convolutional and linear layer of the appearance and geometry/depth decoder. We use rank $r=8$, scaling factor $\alpha=8$, and a dropout rate of $0.1$. This provides a lightweight decoder adaptation mechanism for the color and depth reconstruction heads, without fully fine-tuning the decoder.

\inparagraph{Adapting the Denoising U-Net.} We concatenate appearance and geometry latents along the feature dimension as the input of our denoising U-Net. We adapt the input and output convolutional layers of the U-Net, to make it able to process the expanded inputs. For the input layer, we duplicate the weight tensor and divide its values by two, to prevent an increase in the first-layer activation magnitudes \cite{ke2024marigold}. For the output layer, we \textit{zero}-initialize \cite{zhang2023adding} the
weight tensor for geometry channels. This preserves the original capabilities of the Stable Diffusion U-Net at initialization, and to progressively learn the task of decoding the additional geometry output jointly with appearance latents.
\par We further make the denoiser multi-view aware by adding a 3D self-attention branch alongside the existing 2D self-attention. Let $f$ be the first-stage downsampling factor, so the U-Net operates on latents of resolution $H/f \times W/f$, and its level $l \in \{0,1,2,3\}$ on features downsampled by $2^l f$ relative to the input. Since cross-view attention across all $N{+}M$ views scales quadratically with the token count, which quadruples with each finer level, we add the branch only where $2^l f > f$, that is $l \geq 1$, in both the encoder and the decoder and at the bottleneck; this means level $l = 0$ keeps 2D self-attention alone. When fine-tuning on sequences (\cf~\cref{sec:ablating_3dconvs}) 3D convolutional blocks are added as a further \textit{zero}-initialized residual branch inside these same blocks, therefore following the same rule.

\inparagraph{Conditioning the Denoising U-Net.} To condition the U-Net on preliminary latents obtained from the 3DGS representation, we implement $\phi$ (\cf \cref{subsec:mv_latent_diffusion} and \cref{fig:method_fig}) as a convolutional network composed of 12 Conv2D layers with GroupNorm \cite{wu2018gn} and SiLU \cite{ramachandran2018silu} activations, matching the number of decoder blocks in the Stable Diffusion U-Net. The intermediate feature maps produced by $\phi$ are added to the U-Net skip-connections from the encoder, which are concatenated with the inputs of the corresponding decoder layers.

\inparagraph{Pyramid Noise.} During training, we rely on annealed multi-resolution noise \cite{ke2024marigold} (or \textit{pyramid} noise), which combines multiple Gaussian noise images sampled at different resolutions and then upsampled to match the original U-Net input. Annealed multi-resolution noise has been shown to provide faster convergence and increased accuracy for depth estimation  \cite{ke2024marigold}. 
Following \cite{ke2024marigold}, we anneal the strength of low-resolution noise components as a function of the current diffusion timestep $t$, such that the resulting input noise matches the original Gaussian noise at timestep $t=\num{0}$. 

\inparagraph{Sampling.} We use classifier-free guidance \cite{ho2022cfg} for sampling. Unless otherwise specified, all our main results are obtained using $\gamma_\text{CFG} = \num{3.0}$. We ablate the influence of CFG in \cref{app:suppl_experiments}. To train the unconditional model, during training we randomly drop conditioning $\mathbf{c}$ with a probability of $p_\text{uncond}=0.1$.

%% file: figures/fig_app_qualitative_re10k.tex
\begin{figure*}[t]
    \centering
\input{figures/app_qualitative/qual_re10k}
    \caption{\textbf{Extensive qualitative comparisons on RE10K.} 
    Given $N=\num{2}$ context views \emph{(leftmost column)}, our \ours predicts extrapolated novel views and consistent depth \emph{(third and fourth columns)} for unobserved viewpoints. We show the corresponding ground-truth image for each target viewpoint \emph{(second column)}. We compare our predictions with MVSplat360 \cite{chen2024mvsplat360} and latentSplat \cite{wewer2024latentsplat} \emph{(fifth to eighth columns)}.  
    }
    \label{fig:app_qualitive_re10k}
\end{figure*}
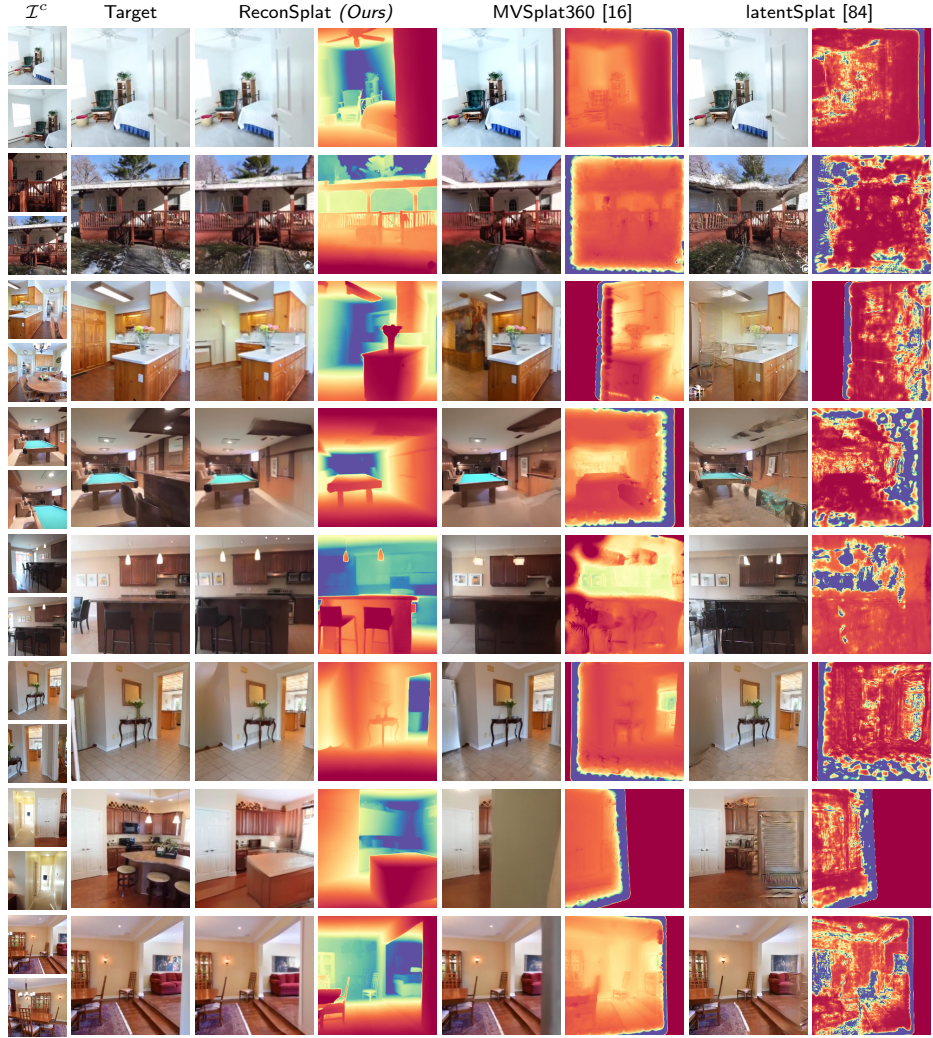

%% file: figures/app_qualitative/qual_re10k.tex
\scriptsize
\sffamily
\setlength{\tabcolsep}{1pt}
\renewcommand{\arraystretch}{1.0}

\newcommand{\imgwidth}{0.128}

\begin{tabular}{
    >{\centering\arraybackslash}m{0.0625\textwidth}
    >{\centering\arraybackslash}m{\imgwidth\textwidth}
    >{\centering\arraybackslash}m{\imgwidth\textwidth}
    >{\centering\arraybackslash}m{\imgwidth\textwidth}
    >{\centering\arraybackslash}m{\imgwidth\textwidth}
    >{\centering\arraybackslash}m{\imgwidth\textwidth}
    >{\centering\arraybackslash}m{\imgwidth\textwidth}
    >{\centering\arraybackslash}m{\imgwidth\textwidth}
}

$\mathcal{I}^c$
& Target %
& \multicolumn{2}{c}{\ours \textit{(Ours)}}
& \multicolumn{2}{c}{MVSplat360~\cite{chen2024mvsplat360}}
& \multicolumn{2}{c}{latentSplat~\cite{wewer2024latentsplat}}
\\[1pt]

\shortstack{\includegraphics[width=\linewidth]{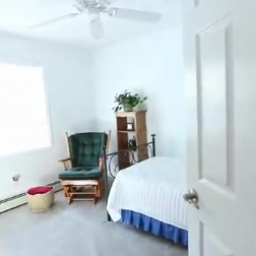}\\[-1pt]
\includegraphics[width=\linewidth]{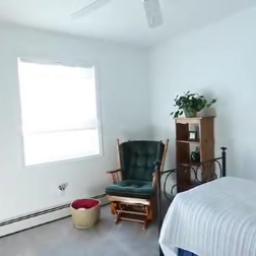}}
& \includegraphics[width=\linewidth]{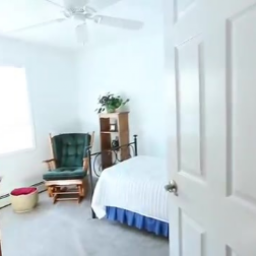} 
& \includegraphics[width=\linewidth]{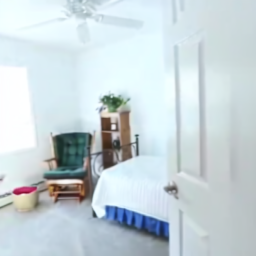}
& \includegraphics[width=\linewidth]{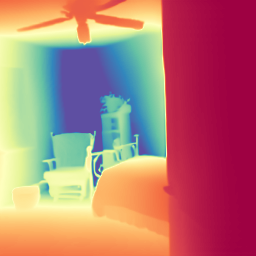}
& \includegraphics[width=\linewidth]{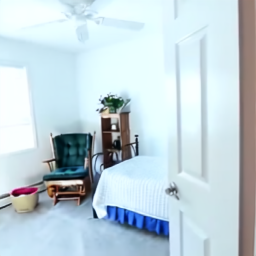} 
& \includegraphics[width=\linewidth]{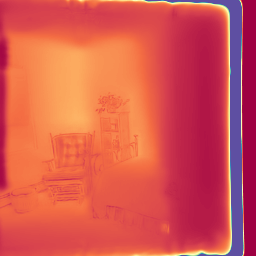} 
& \includegraphics[width=\linewidth]{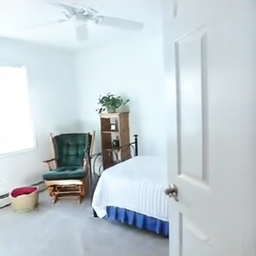} 
& \includegraphics[width=\linewidth]{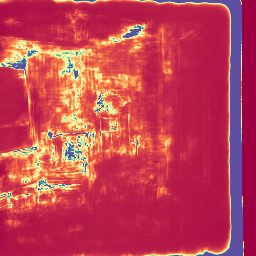} 
\\

\shortstack{\includegraphics[width=\linewidth]{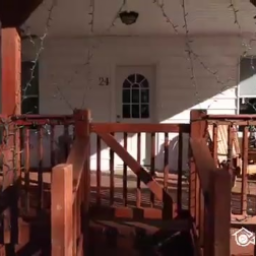}\\[-1pt]
\includegraphics[width=\linewidth]{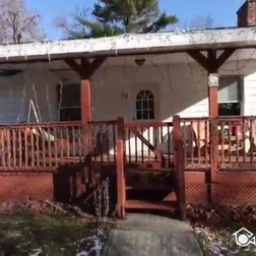}}
& \includegraphics[width=\linewidth]{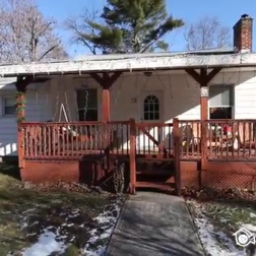} 
& \includegraphics[width=\linewidth]{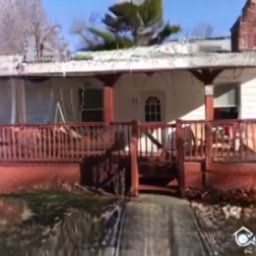}
& \includegraphics[width=\linewidth]{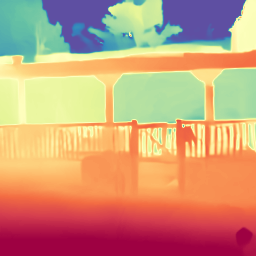}
& \includegraphics[width=\linewidth]{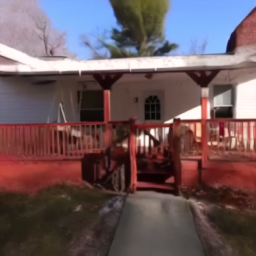} 
& \includegraphics[width=\linewidth]{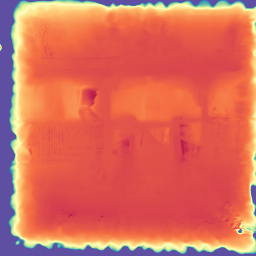} 
& \includegraphics[width=\linewidth]{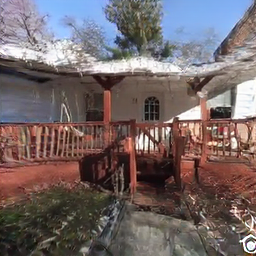} 
& \includegraphics[width=\linewidth]{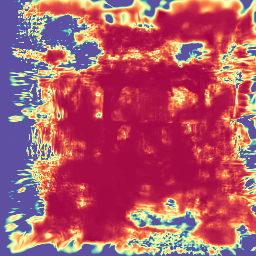} 
\\
\shortstack{\includegraphics[width=\linewidth]{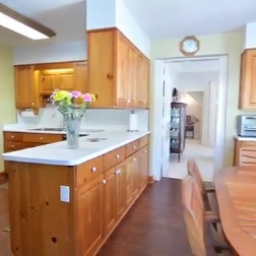}\\[-1pt]
\includegraphics[width=\linewidth]{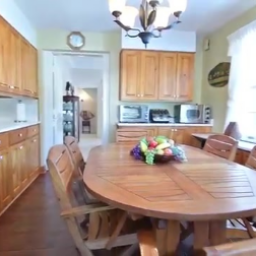}}
& \includegraphics[width=\linewidth]{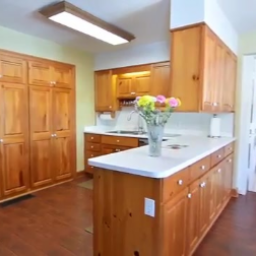} 
& \includegraphics[width=\linewidth]{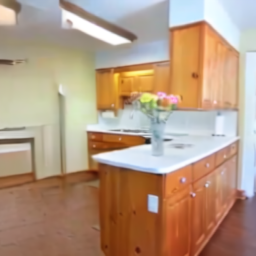}
& \includegraphics[width=\linewidth]{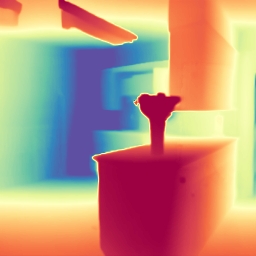}
& \includegraphics[width=\linewidth]{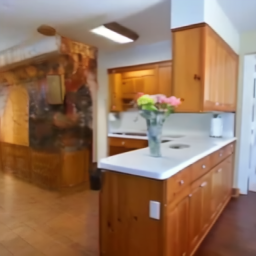} 
& \includegraphics[width=\linewidth]{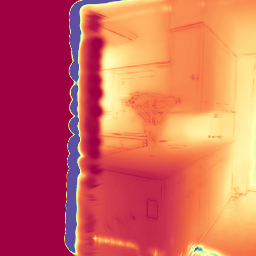} 
& \includegraphics[width=\linewidth]{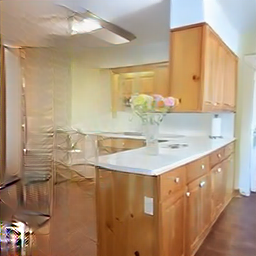} 
& \includegraphics[width=\linewidth]{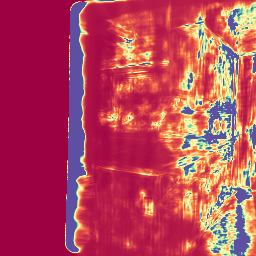} 
\\
\shortstack{\includegraphics[width=\linewidth]{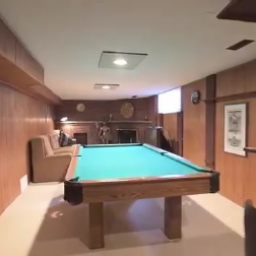}\\[-1pt]
\includegraphics[width=\linewidth]{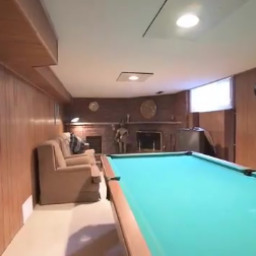}}
& \includegraphics[width=\linewidth]{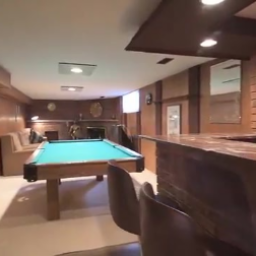} 
& \includegraphics[width=\linewidth]{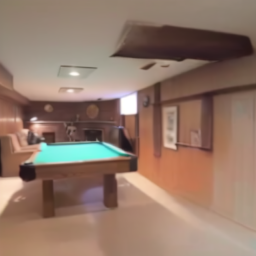}
& \includegraphics[width=\linewidth]{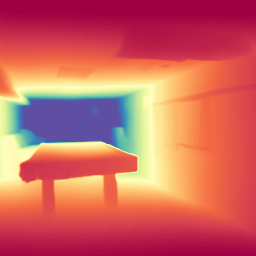}
& \includegraphics[width=\linewidth]{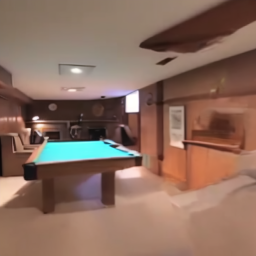} 
& \includegraphics[width=\linewidth]{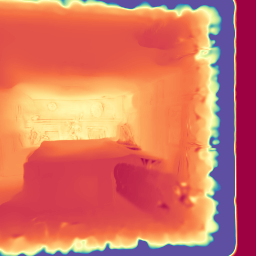} 
& \includegraphics[width=\linewidth]{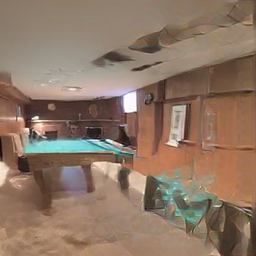} 
& \includegraphics[width=\linewidth]{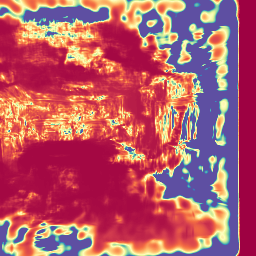} 
\\
\shortstack{\includegraphics[width=\linewidth]{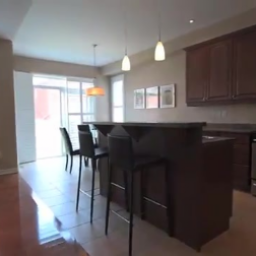}\\[-1pt]
\includegraphics[width=\linewidth]{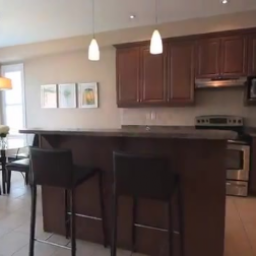}}
& \includegraphics[width=\linewidth]{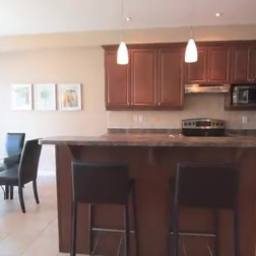} 
& \includegraphics[width=\linewidth]{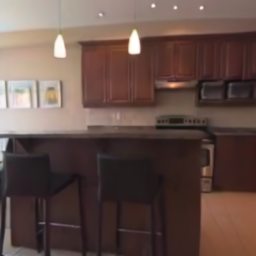}
& \includegraphics[width=\linewidth]{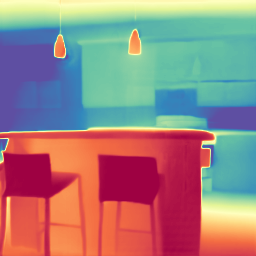}
& \includegraphics[width=\linewidth]{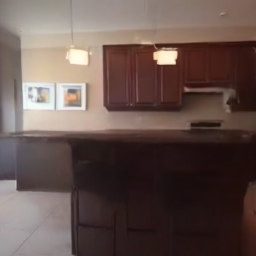} 
& \includegraphics[width=\linewidth]{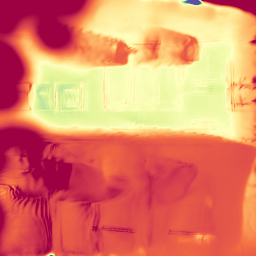} 
& \includegraphics[width=\linewidth]{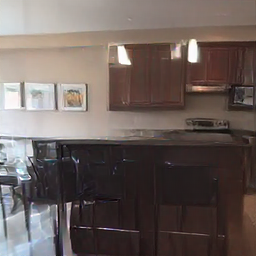} 
& \includegraphics[width=\linewidth]{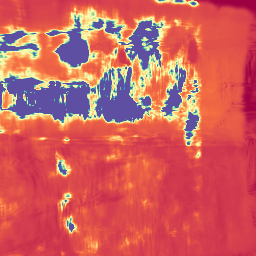} 
\\
\shortstack{\includegraphics[width=\linewidth]{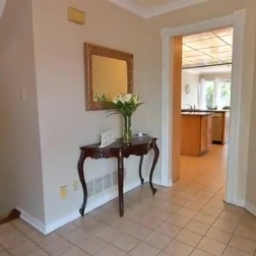}\\[-1pt]
\includegraphics[width=\linewidth]{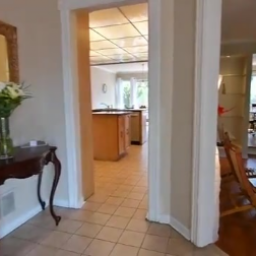}}
& \includegraphics[width=\linewidth]{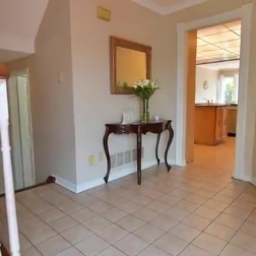} 
& \includegraphics[width=\linewidth]{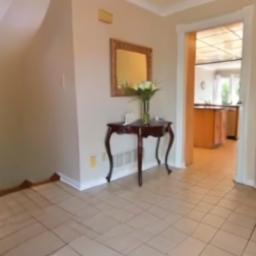}
& \includegraphics[width=\linewidth]{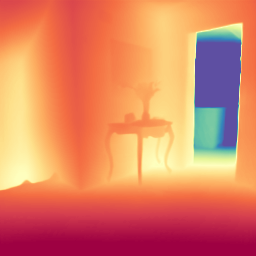}
& \includegraphics[width=\linewidth]{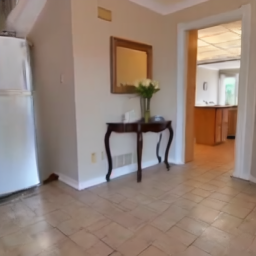} 
& \includegraphics[width=\linewidth]{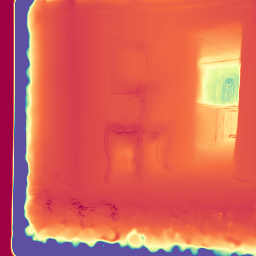} 
& \includegraphics[width=\linewidth]{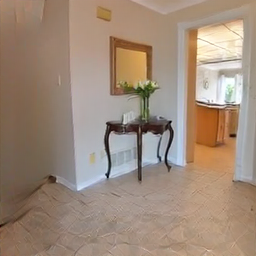} 
& \includegraphics[width=\linewidth]{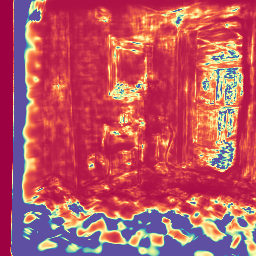} 
\\
\shortstack{\includegraphics[width=\linewidth]{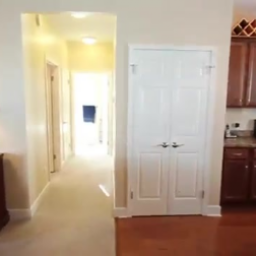}\\[-1pt]
\includegraphics[width=\linewidth]{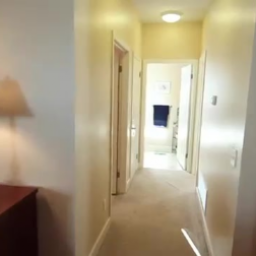}}
& \includegraphics[width=\linewidth]{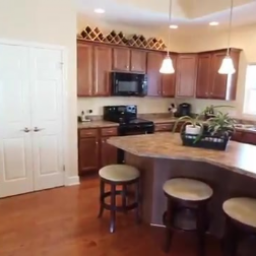} 
& \includegraphics[width=\linewidth]{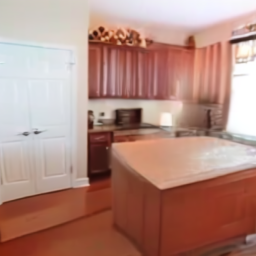}
& \includegraphics[width=\linewidth]{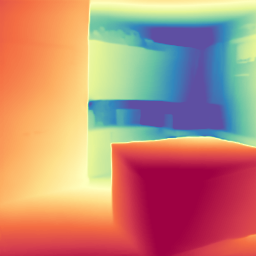}
& \includegraphics[width=\linewidth]{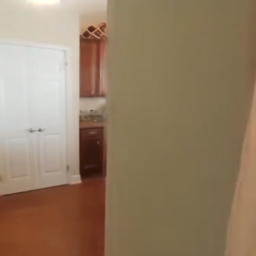} 
& \includegraphics[width=\linewidth]{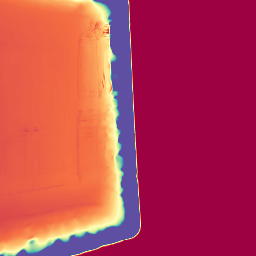} 
& \includegraphics[width=\linewidth]{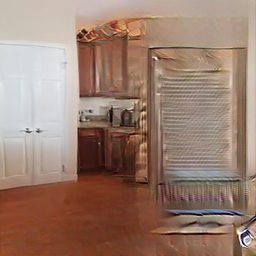} 
& \includegraphics[width=\linewidth]{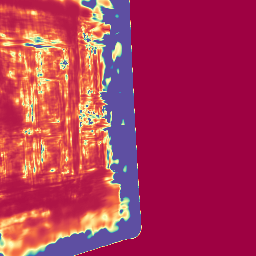} 
\\
\shortstack{\includegraphics[width=\linewidth]{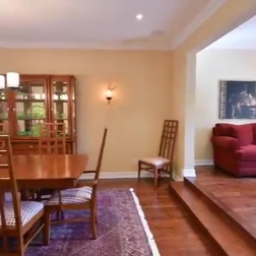}\\[-1pt]
\includegraphics[width=\linewidth]{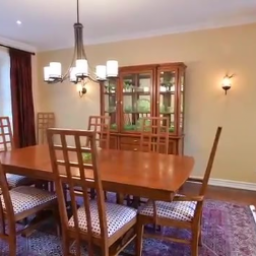}}
& \includegraphics[width=\linewidth]{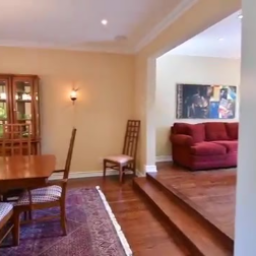} 
& \includegraphics[width=\linewidth]{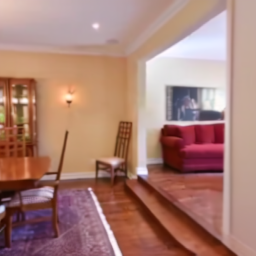}
& \includegraphics[width=\linewidth]{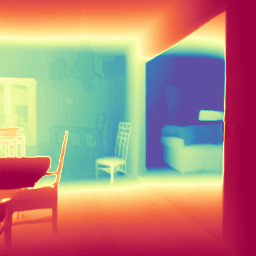}
& \includegraphics[width=\linewidth]{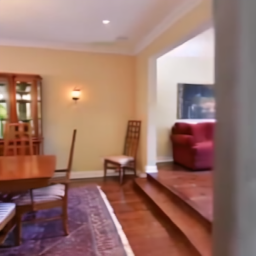} 
& \includegraphics[width=\linewidth]{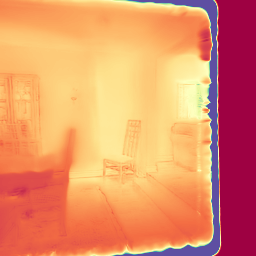} 
& \includegraphics[width=\linewidth]{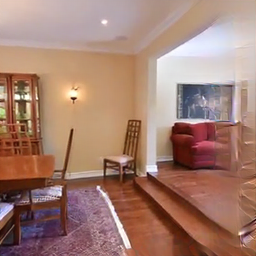} 
& \includegraphics[width=\linewidth]{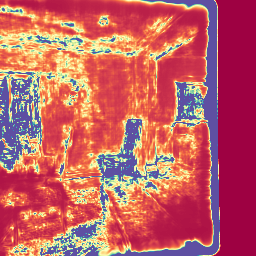} 
\\
\end{tabular}

%% file: figures/fig_app_qualitative_dl3dv.tex
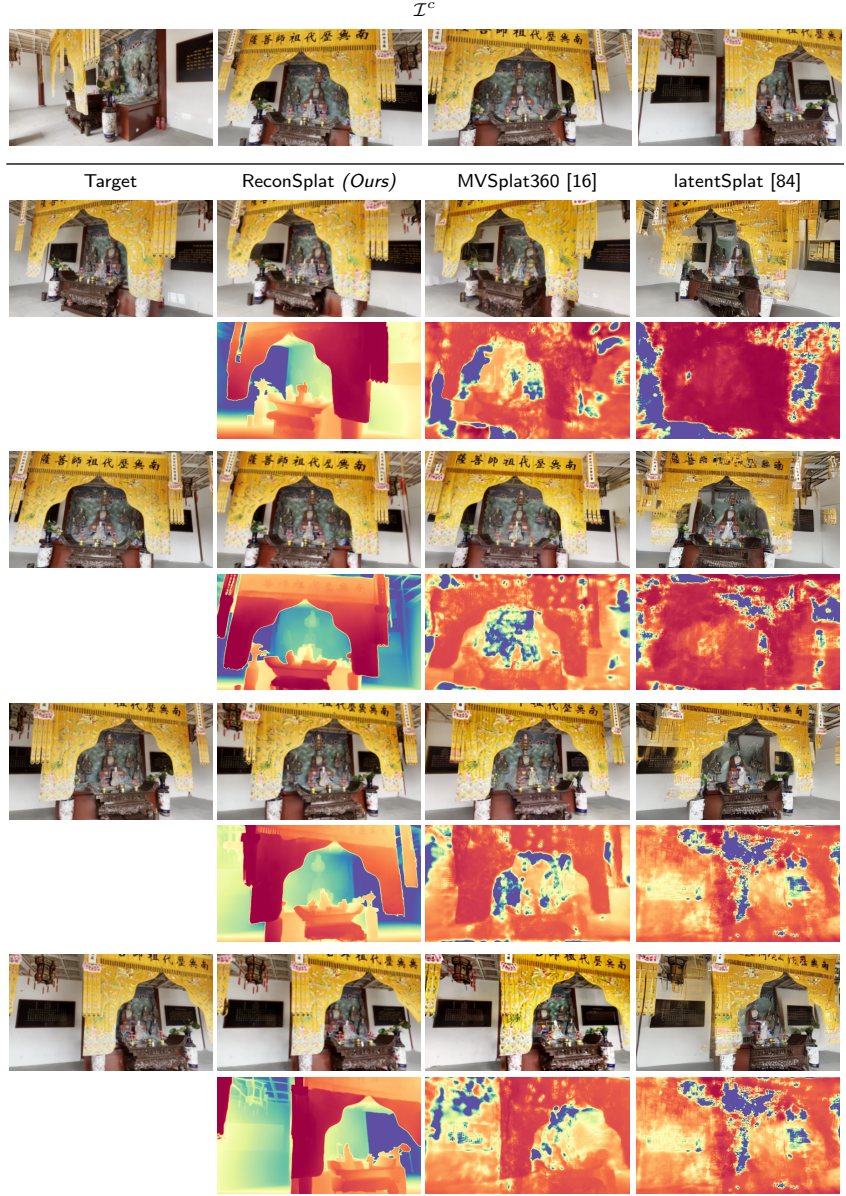
\begin{figure*}[t]
    \centering
    \input{figures/app_qualitative/qual_dl3dv}
    \caption{\textbf{Extensive qualitative comparisons on DL3DV-10K.} Given $N=\num{4}$ context views \emph{(top row)}, \ours predicts high-quality novel views \emph{(odd rows, second column)} and accurate depth predictions \emph{(even rows, second column)} for unobserved viewpoints compared to prior work for the challenging wide-baseline setup of DL3DV-10K. For each target viewpoint, we show the corresponding ground-truth image \emph{(first column)}, followed by predictions from \ours and other generative baselines.  
    }
    \label{fig:app_qualitive_dl3dv}
\end{figure*}

\begin{figure*}[t]
    \centering
    \input{figures/app_qualitative/qual_dl3dv5}
    \caption{\textbf{Extensive qualitative comparisons on DL3DV-10K.} Given $N=\num{4}$ context views \emph{(top row)}, \ours predicts high-quality novel views \emph{(odd rows, second column)} and accurate depth predictions \emph{(even rows, second column)} for unobserved viewpoints compared to prior work for the challenging wide-baseline setup of DL3DV-10K. For each target viewpoint, we show the corresponding ground-truth image \emph{(first column)}, followed by predictions from \ours and other generative baselines.  
    }
    \label{fig:app_qualitive_dl3dv2}
\end{figure*}
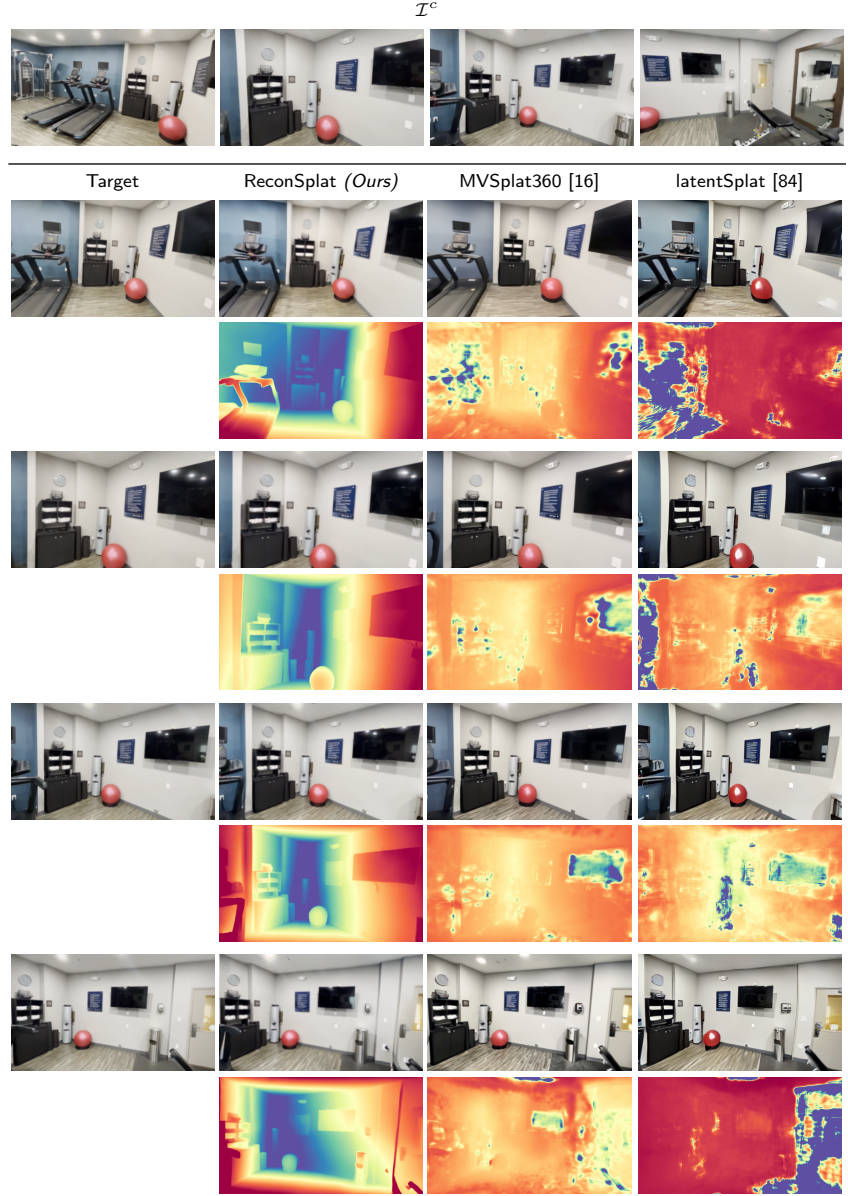

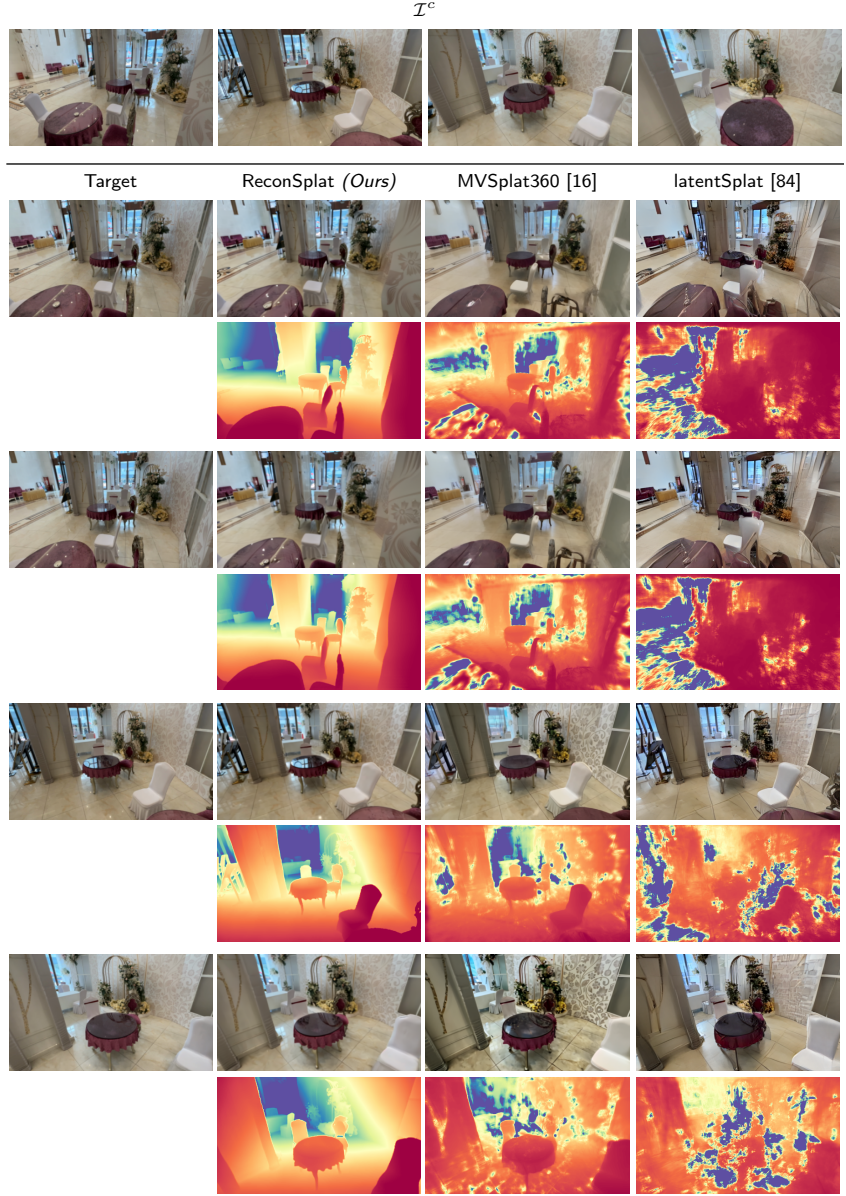
\begin{figure*}[t]
    \centering
    \input{figures/app_qualitative/qual_dl3dv3}
    \caption{\textbf{Extensive qualitative comparisons on DL3DV-10K.} Given $N=\num{4}$ context views \emph{(top row)}, \ours predicts high-quality novel views \emph{(odd rows, second column)} and accurate depth predictions \emph{(even rows, second column)} for unobserved viewpoints compared to prior work for the challenging wide-baseline setup of DL3DV-10K. For each target viewpoint, we show the corresponding ground-truth image \emph{(first column)}, followed by predictions from \ours and other generative baselines.  
    }
    \label{fig:app_qualitive_dl3dv3}
\end{figure*}

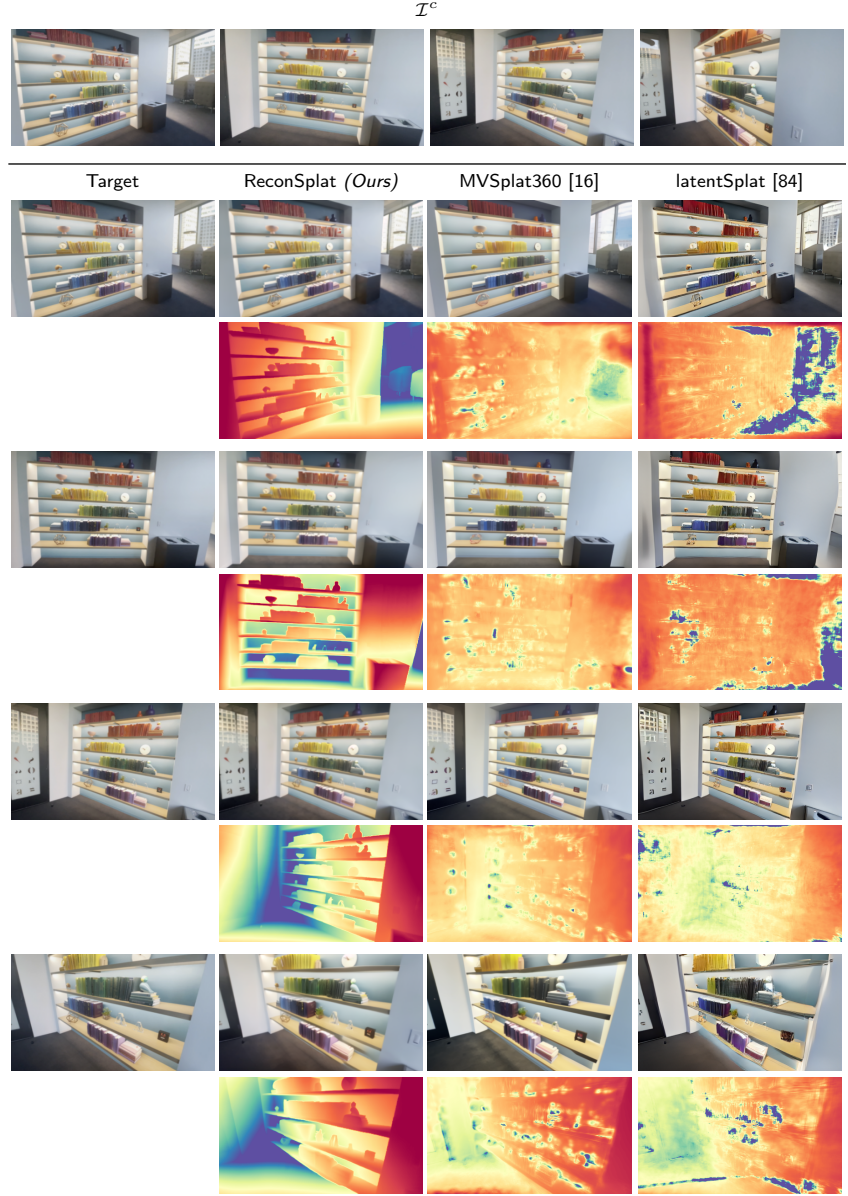
\begin{figure*}[t]
    \centering
    \input{figures/app_qualitative/qual_dl3dv6}
    \caption{\textbf{Extensive qualitative comparisons on DL3DV-10K.} Given $N=\num{4}$ context views \emph{(top row)}, \ours predicts high-quality novel views \emph{(odd rows, second column)} and accurate depth predictions \emph{(even rows, second column)} for unobserved viewpoints compared to prior work for the challenging wide-baseline setup of DL3DV-10K. For each target viewpoint, we show the corresponding ground-truth image \emph{(first column)}, followed by predictions from \ours and other generative baselines.  
    }
    \label{fig:app_qualitive_dl3dv4}
\end{figure*}

\begin{figure*}[t]
    \centering
    \input{figures/app_qualitative/qual_dl3dv4}
    \caption{\textbf{Extensive qualitative comparisons on DL3DV-10K.} Given $N=\num{4}$ context views \emph{(top row)}, \ours predicts high-quality novel views \emph{(odd rows, second column)} and accurate depth predictions \emph{(even rows, second column)} for unobserved viewpoints compared to prior work for the challenging wide-baseline setup of DL3DV-10K. For each target viewpoint, we show the corresponding ground-truth image \emph{(first column)}, followed by predictions from \ours and other generative baselines.  
    }
    \label{fig:app_qualitive_dl3dv5}
\end{figure*}
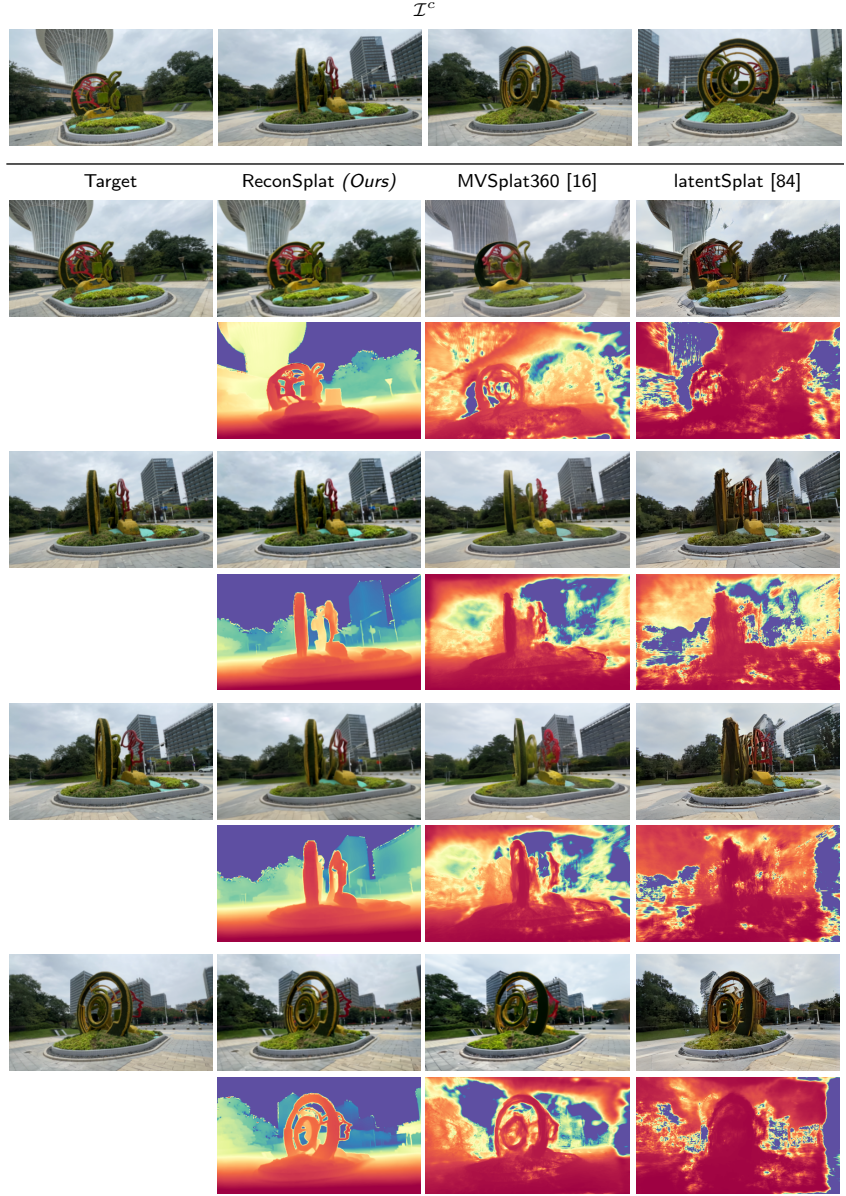

%% file: figures/app_qualitative/qual_dl3dv.tex
\scriptsize
\sffamily
\setlength{\tabcolsep}{1pt}
\renewcommand{\arraystretch}{1.0}

\newcommand{\imgwidth}{0.21}

\begin{tabular}{cccc}

\multicolumn{4}{c}{$\mathcal{I}^c$}\\[0.35em]

\multicolumn{4}{c}{
    \includegraphics[width=0.22\textwidth]{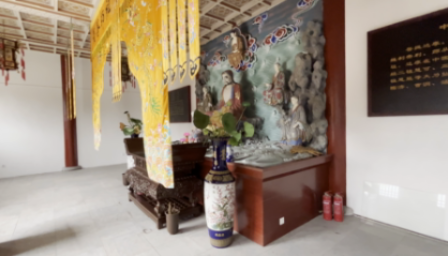}
    \includegraphics[width=0.22\textwidth]{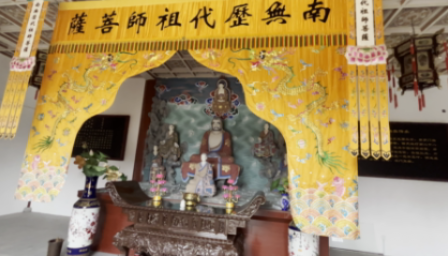}
    \includegraphics[width=0.22\textwidth]{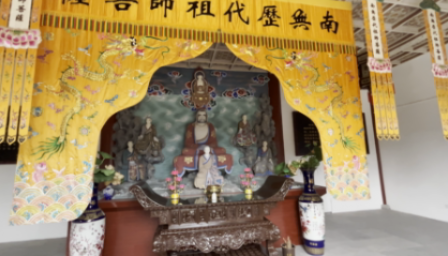}
    \includegraphics[width=0.22\textwidth]{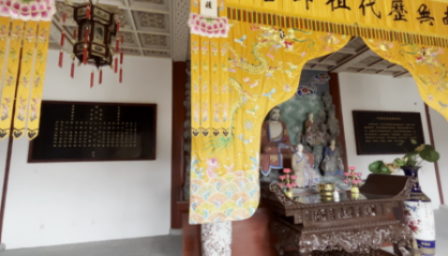}
}\\[0.35em]

\midrule

Target 
&\ours \textit{(Ours)}
& MVSplat360~\cite{chen2024mvsplat360}
& latentSplat~\cite{wewer2024latentsplat}
\\[0.35em]

\includegraphics[width=0.22\textwidth]{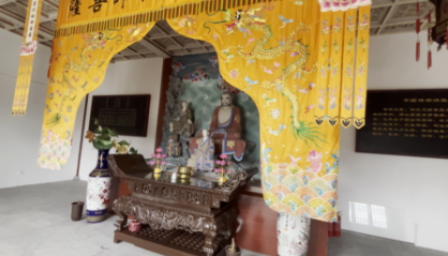}
&
\includegraphics[width=0.22\textwidth]{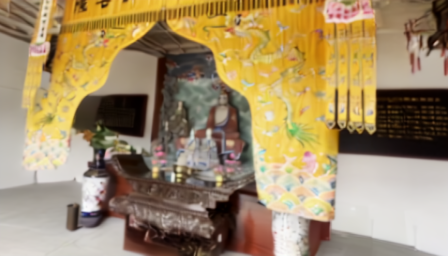}
&
\includegraphics[width=0.22\textwidth]{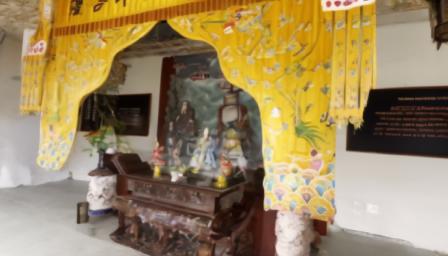}
&
\includegraphics[width=0.22\textwidth]{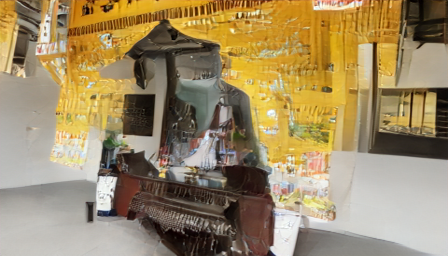}
\\

&
\includegraphics[width=0.22\textwidth]{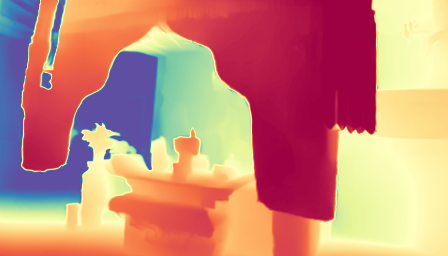}
&
\includegraphics[width=0.22\textwidth]{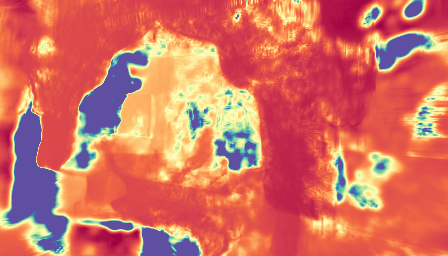}
&
\includegraphics[width=0.22\textwidth]{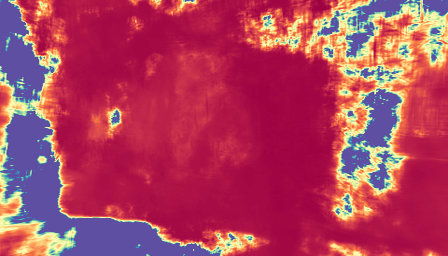}
\\[0.35em]

\includegraphics[width=0.22\textwidth]{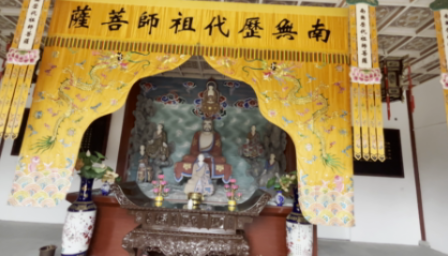}
&
\includegraphics[width=0.22\textwidth]{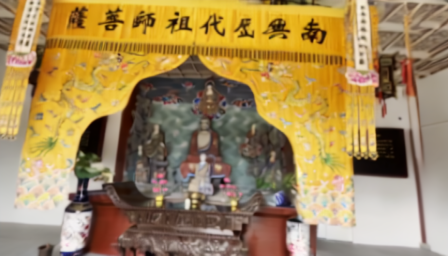}
&
\includegraphics[width=0.22\textwidth]{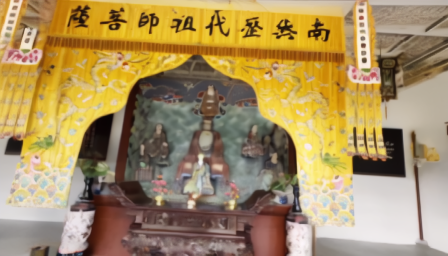}
&
\includegraphics[width=0.22\textwidth]{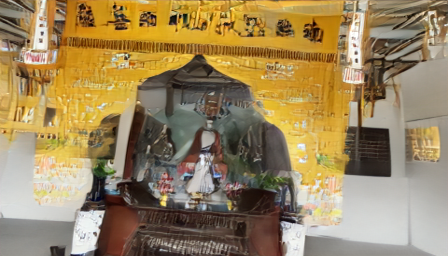}
\\

&
\includegraphics[width=0.22\textwidth]{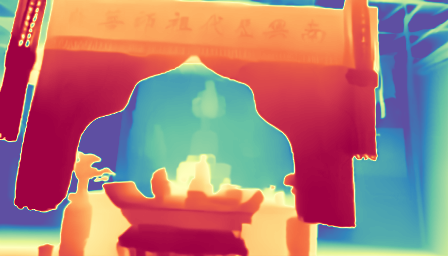}
&
\includegraphics[width=0.22\textwidth]{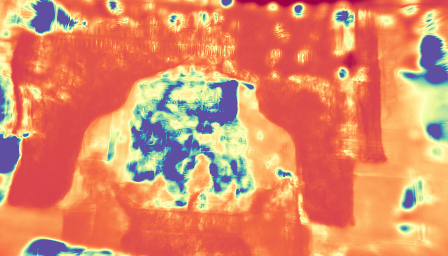}
&
\includegraphics[width=0.22\textwidth]{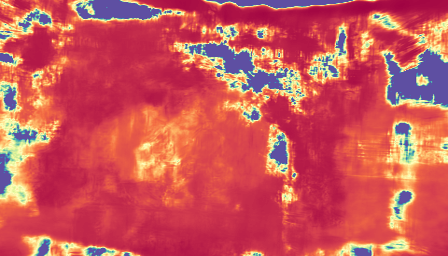}
\\[0.35em]

\includegraphics[width=0.22\textwidth]{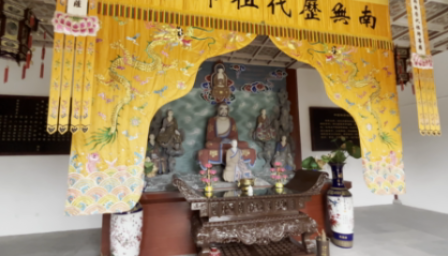}
&
\includegraphics[width=0.22\textwidth]{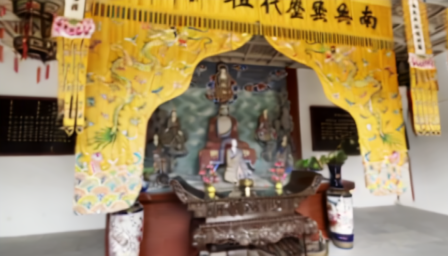}
&
\includegraphics[width=0.22\textwidth]{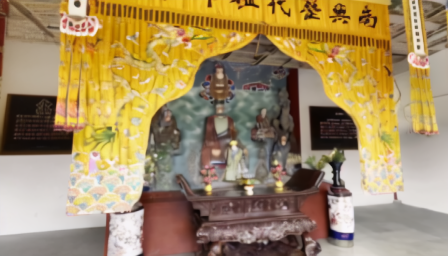}
&
\includegraphics[width=0.22\textwidth]{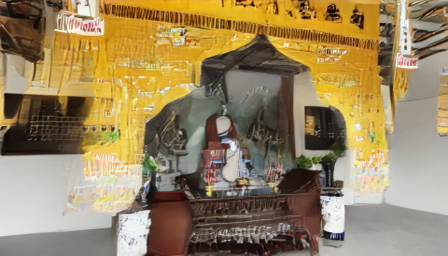}
\\

&
\includegraphics[width=0.22\textwidth]{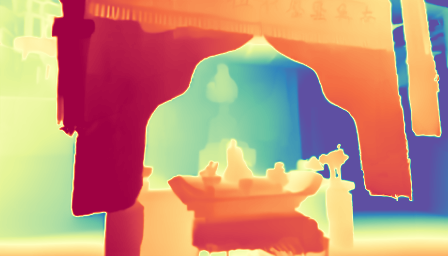}
&
\includegraphics[width=0.22\textwidth]{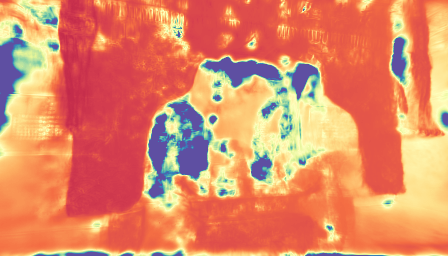}
&
\includegraphics[width=0.22\textwidth]{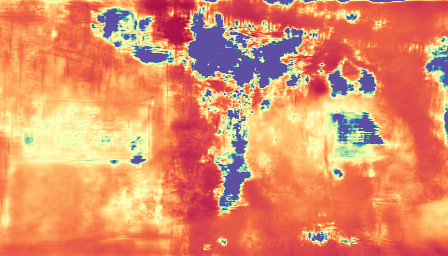}
\\[0.35em]

\includegraphics[width=0.22\textwidth]{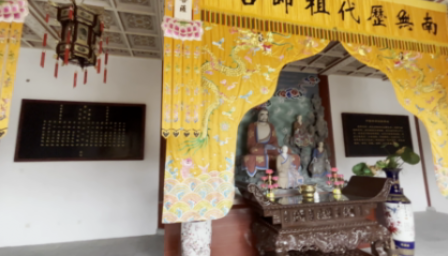}
&
\includegraphics[width=0.22\textwidth]{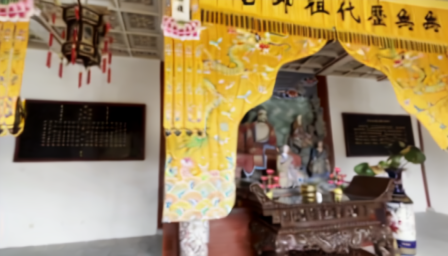}
&
\includegraphics[width=0.22\textwidth]{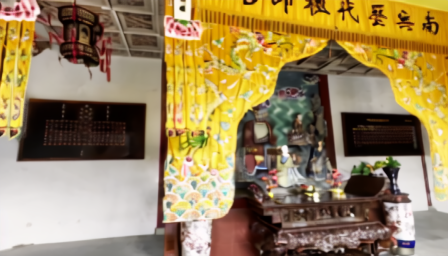}
&
\includegraphics[width=0.22\textwidth]{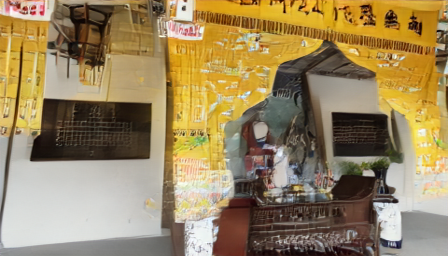}
\\

&
\includegraphics[width=0.22\textwidth]{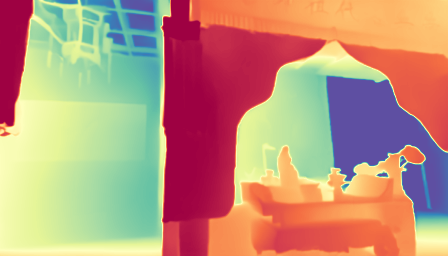}
&
\includegraphics[width=0.22\textwidth]{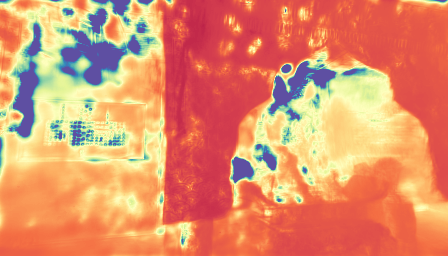}
&
\includegraphics[width=0.22\textwidth]{figures/app_qualitative/dl3dv_eccv_app/resize_1da88_ls/depth/000106.png}
\\

\end{tabular}

%% file: figures/app_qualitative/qual_dl3dv5.tex
\scriptsize
\sffamily
\setlength{\tabcolsep}{1pt}
\renewcommand{\arraystretch}{1.0}

\newcommand{\imgwidth}{0.21}

\begin{tabular}{cccc}

\multicolumn{4}{c}{$\mathcal{I}^c$}\\[0.35em]

\multicolumn{4}{c}{
    \includegraphics[width=0.22\textwidth]{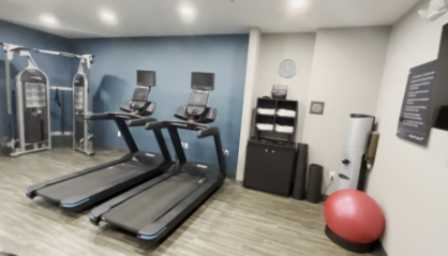}
    \includegraphics[width=0.22\textwidth]{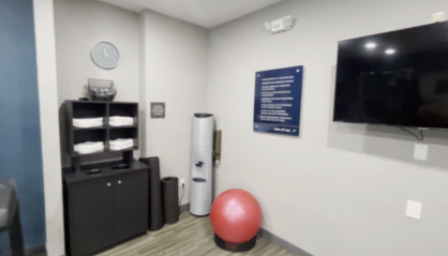}
    \includegraphics[width=0.22\textwidth]{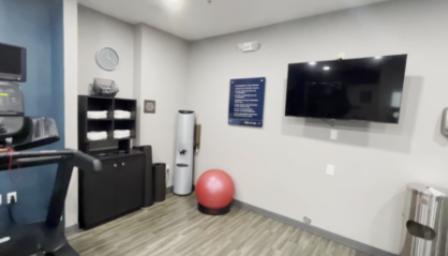}
    \includegraphics[width=0.22\textwidth]{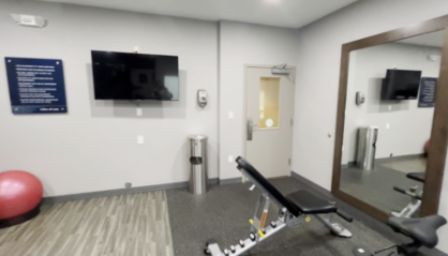}
}\\[0.35em]

\midrule

Target 
& \ours \textit{(Ours)}
& MVSplat360~\cite{chen2024mvsplat360}
& latentSplat~\cite{wewer2024latentsplat}
\\[0.35em]

\includegraphics[width=0.22\textwidth]{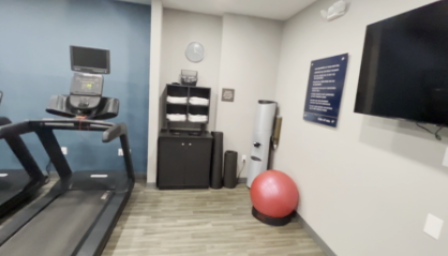}
&
\includegraphics[width=0.22\textwidth]{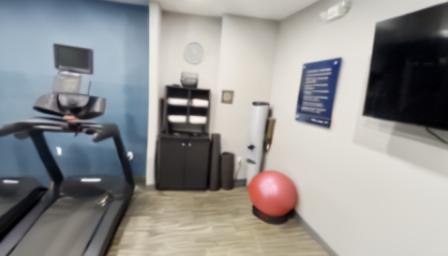}
&
\includegraphics[width=0.22\textwidth]{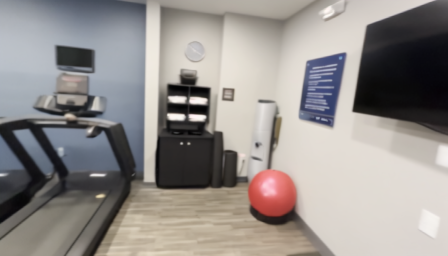}
&
\includegraphics[width=0.22\textwidth]{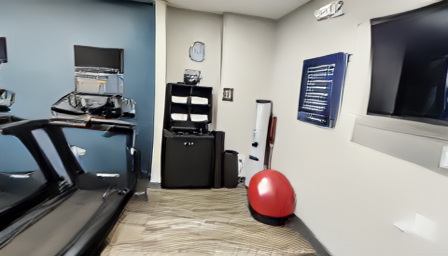}
\\

&
\includegraphics[width=0.22\textwidth]{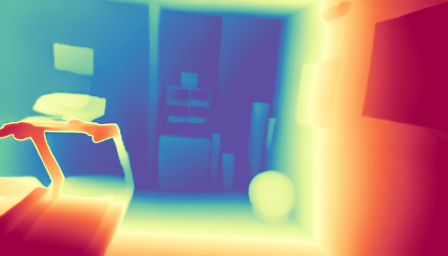}
&
\includegraphics[width=0.22\textwidth]{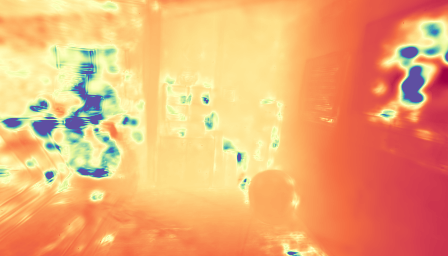}
&
\includegraphics[width=0.22\textwidth]{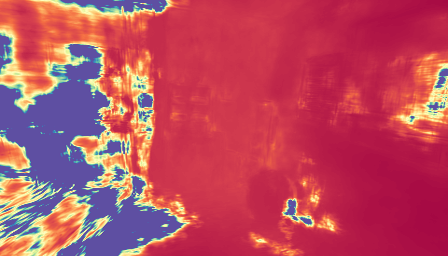}
\\[0.35em]

\includegraphics[width=0.22\textwidth]{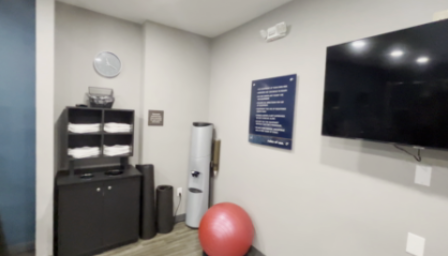}
&
\includegraphics[width=0.22\textwidth]{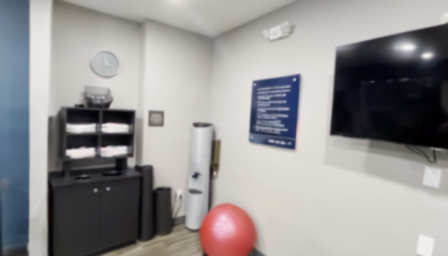}
&
\includegraphics[width=0.22\textwidth]{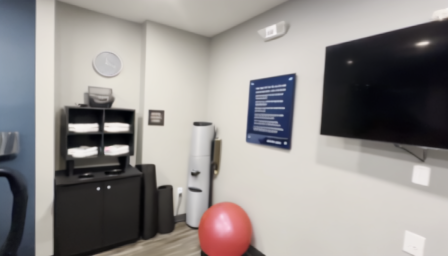}
&
\includegraphics[width=0.22\textwidth]{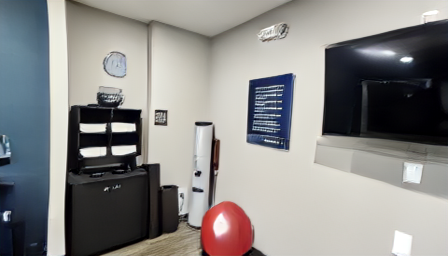}
\\

&
\includegraphics[width=0.22\textwidth]{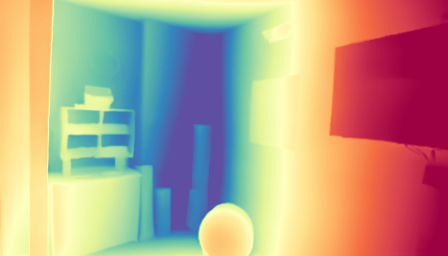}
&
\includegraphics[width=0.22\textwidth]{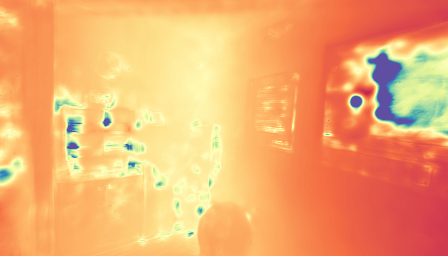}
&
\includegraphics[width=0.22\textwidth]{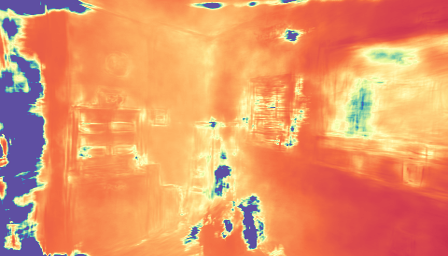}
\\[0.35em]

\includegraphics[width=0.22\textwidth]{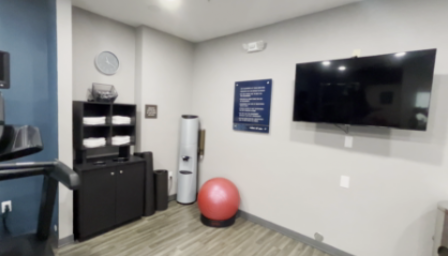}
&
\includegraphics[width=0.22\textwidth]{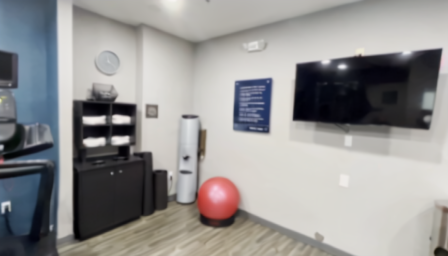}
&
\includegraphics[width=0.22\textwidth]{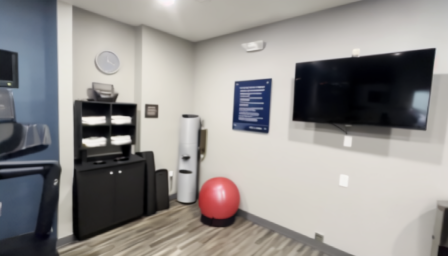}
&
\includegraphics[width=0.22\textwidth]{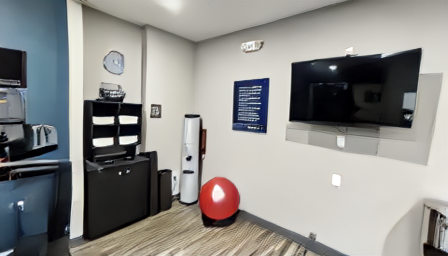}
\\

&
\includegraphics[width=0.22\textwidth]{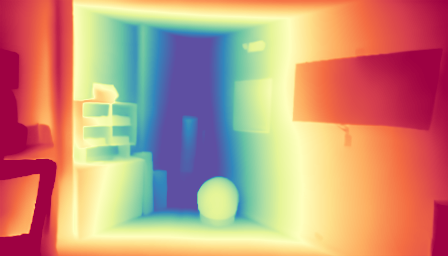}
&
\includegraphics[width=0.22\textwidth]{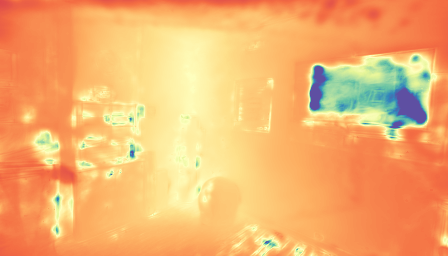}
&
\includegraphics[width=0.22\textwidth]{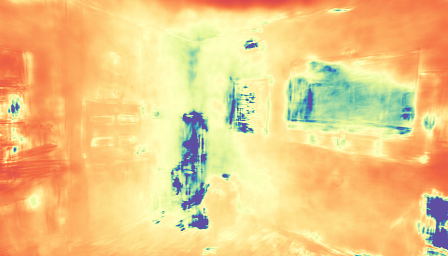}
\\[0.35em]

\includegraphics[width=0.22\textwidth]{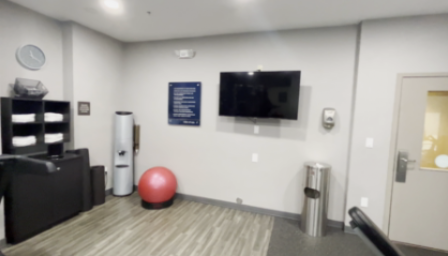}
&
\includegraphics[width=0.22\textwidth]{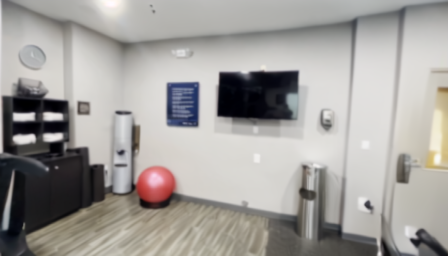}
&
\includegraphics[width=0.22\textwidth]{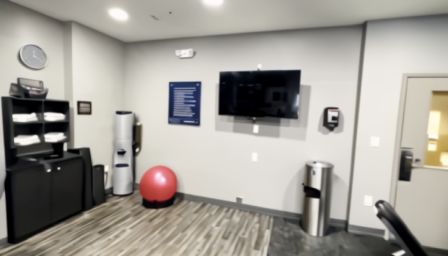}
&
\includegraphics[width=0.22\textwidth]{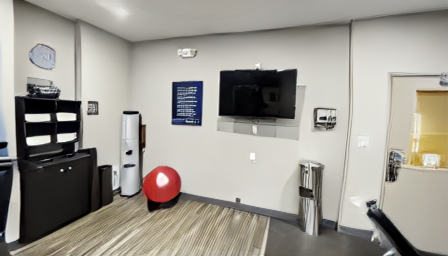}
\\

&
\includegraphics[width=0.22\textwidth]{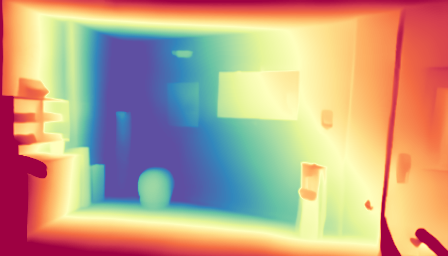}
&
\includegraphics[width=0.22\textwidth]{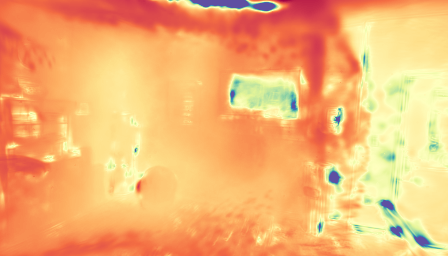}
&
\includegraphics[width=0.22\textwidth]{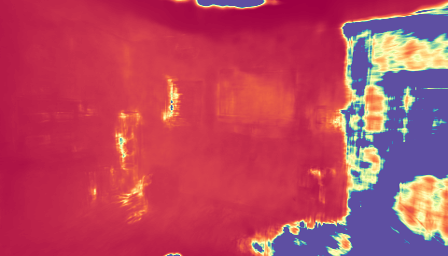}
\\

\end{tabular}

%% file: figures/app_qualitative/qual_dl3dv3.tex
\scriptsize
\sffamily
\setlength{\tabcolsep}{1pt}
\renewcommand{\arraystretch}{1.0}

\newcommand{\imgwidth}{0.21}

\begin{tabular}{cccc}

\multicolumn{4}{c}{$\mathcal{I}^c$}\\[0.35em]

\multicolumn{4}{c}{
    \includegraphics[width=0.22\textwidth]{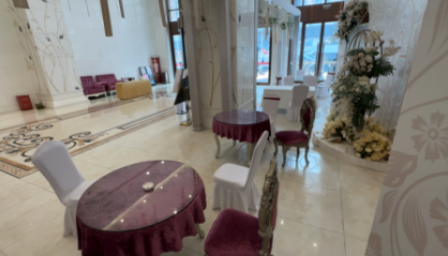}
    \includegraphics[width=0.22\textwidth]{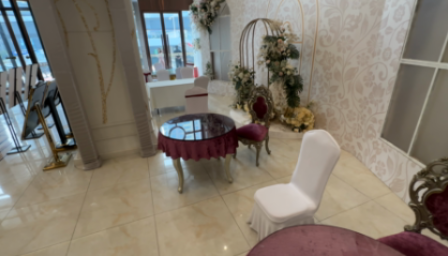}
    \includegraphics[width=0.22\textwidth]{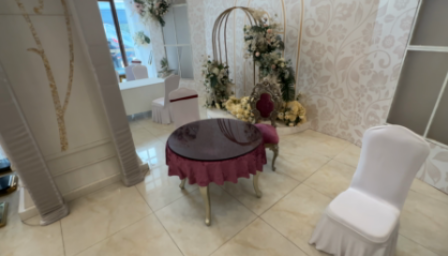}
    \includegraphics[width=0.22\textwidth]{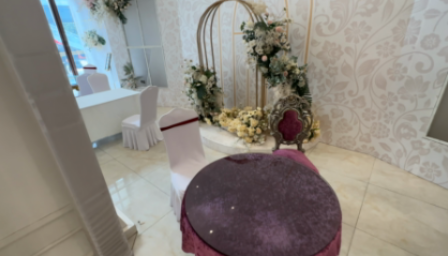}
}\\[0.35em]

\midrule

Target 
& \ours \textit{(Ours)}
& MVSplat360~\cite{chen2024mvsplat360}
& latentSplat~\cite{wewer2024latentsplat}
\\[0.35em]

\includegraphics[width=0.22\textwidth]{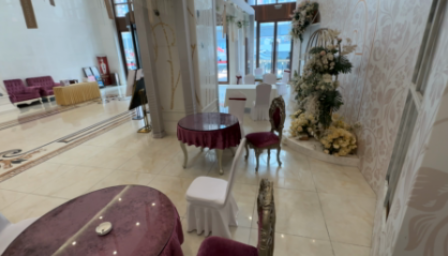}
&
\includegraphics[width=0.22\textwidth]{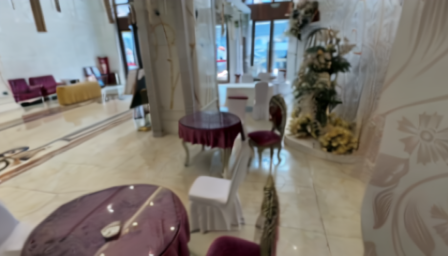}
&
\includegraphics[width=0.22\textwidth]{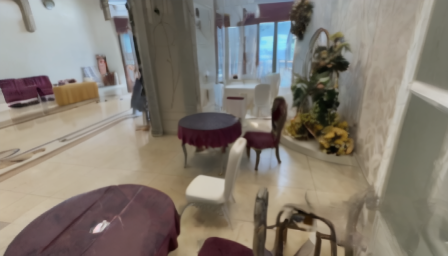}
&
\includegraphics[width=0.22\textwidth]{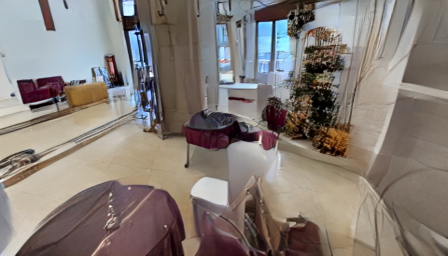}
\\

&
\includegraphics[width=0.22\textwidth]{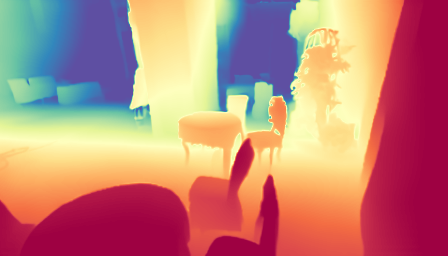}
&
\includegraphics[width=0.22\textwidth]{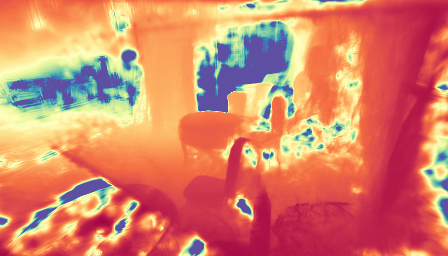}
&
\includegraphics[width=0.22\textwidth]{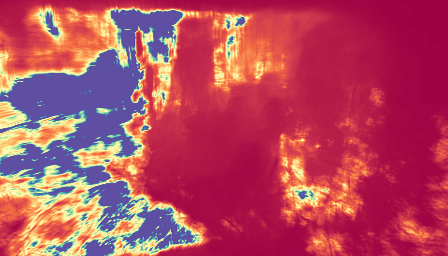}
\\[0.35em]

\includegraphics[width=0.22\textwidth]{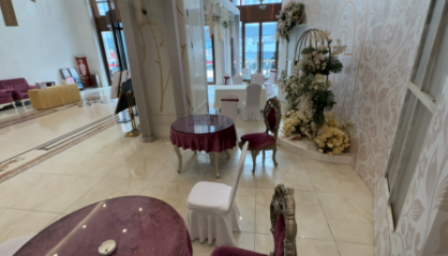}
&
\includegraphics[width=0.22\textwidth]{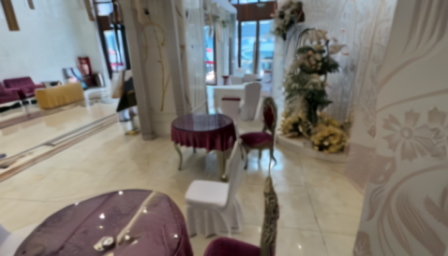}
&
\includegraphics[width=0.22\textwidth]{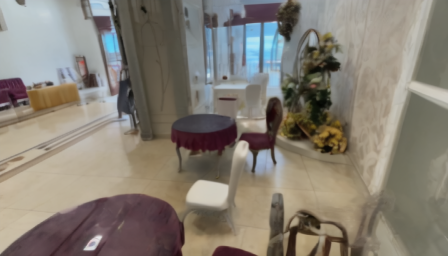}
&
\includegraphics[width=0.22\textwidth]{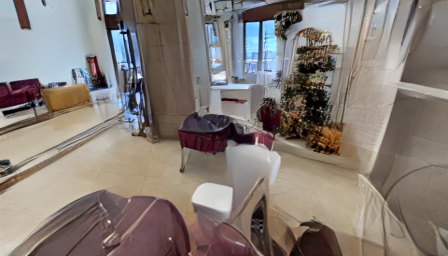}
\\

&
\includegraphics[width=0.22\textwidth]{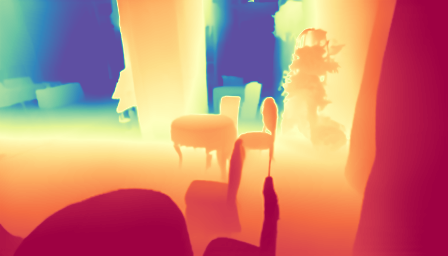}
&
\includegraphics[width=0.22\textwidth]{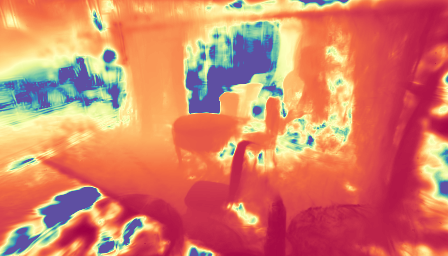}
&
\includegraphics[width=0.22\textwidth]{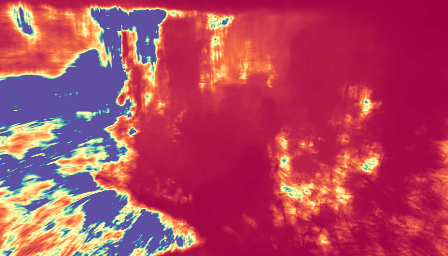}
\\[0.35em]

\includegraphics[width=0.22\textwidth]{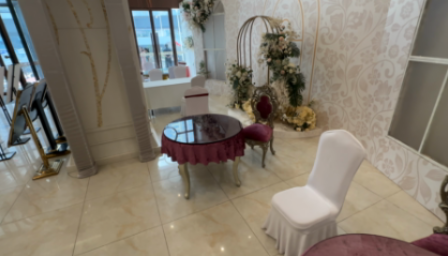}
&
\includegraphics[width=0.22\textwidth]{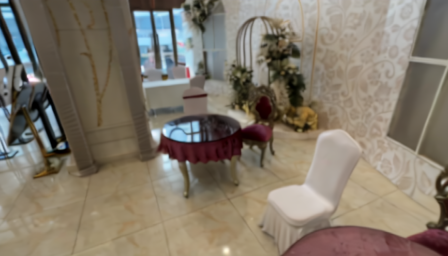}
&
\includegraphics[width=0.22\textwidth]{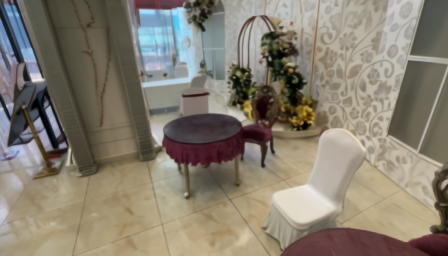}
&
\includegraphics[width=0.22\textwidth]{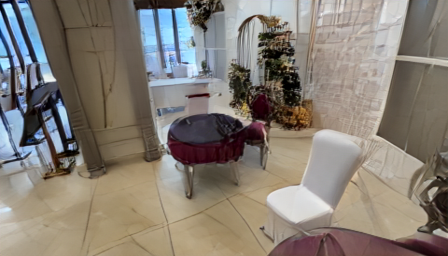}
\\

&
\includegraphics[width=0.22\textwidth]{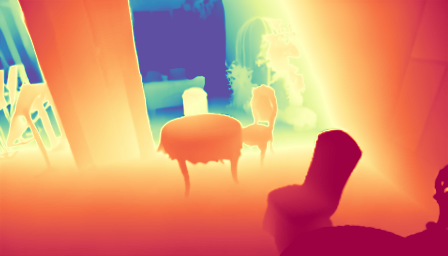}
&
\includegraphics[width=0.22\textwidth]{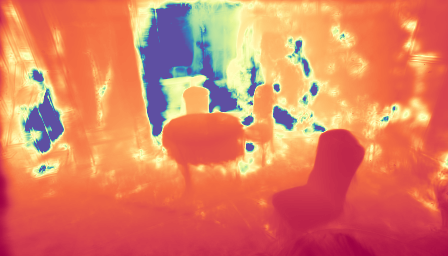}
&
\includegraphics[width=0.22\textwidth]{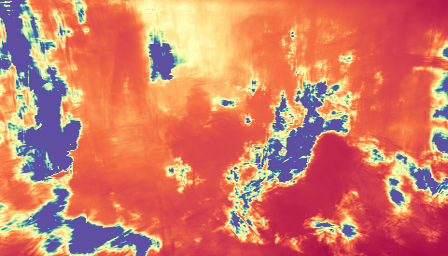}
\\[0.35em]

\includegraphics[width=0.22\textwidth]{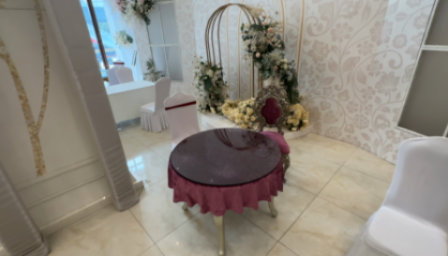}
&
\includegraphics[width=0.22\textwidth]{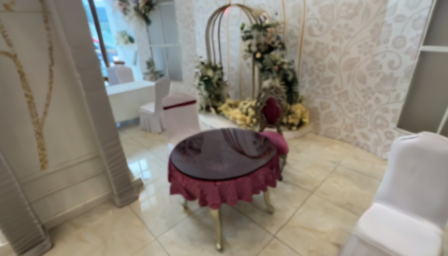}
&
\includegraphics[width=0.22\textwidth]{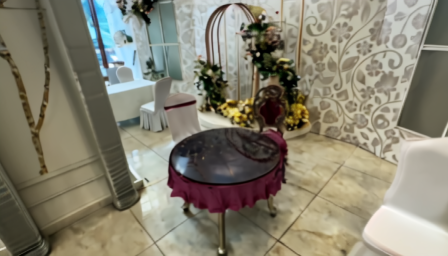}
&
\includegraphics[width=0.22\textwidth]{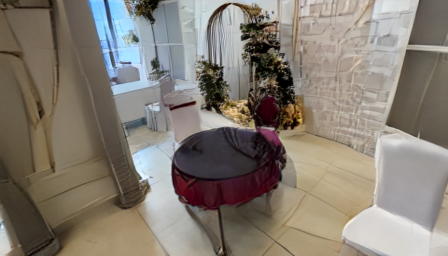}
\\

&
\includegraphics[width=0.22\textwidth]{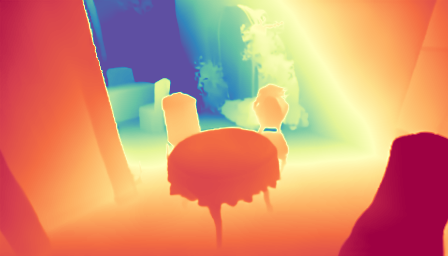}
&
\includegraphics[width=0.22\textwidth]{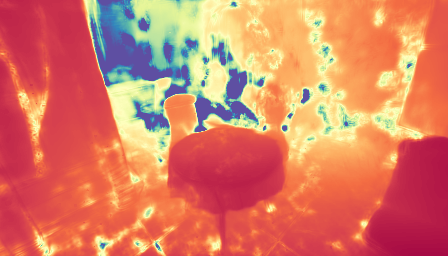}
&
\includegraphics[width=0.22\textwidth]{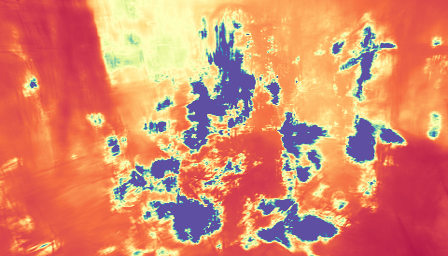}
\\

\end{tabular}

%% file: figures/app_qualitative/qual_dl3dv6.tex
\scriptsize
\sffamily
\setlength{\tabcolsep}{1pt}
\renewcommand{\arraystretch}{1.0}

\newcommand{\imgwidth}{0.21}

\begin{tabular}{cccc}

\multicolumn{4}{c}{$\mathcal{I}^c$}\\[0.35em]

\multicolumn{4}{c}{
    \includegraphics[width=0.22\textwidth]{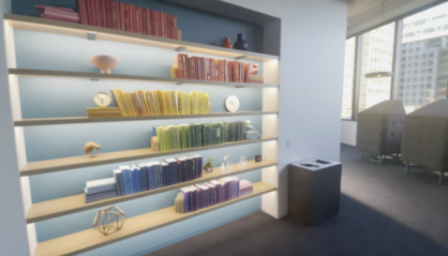}
    \includegraphics[width=0.22\textwidth]{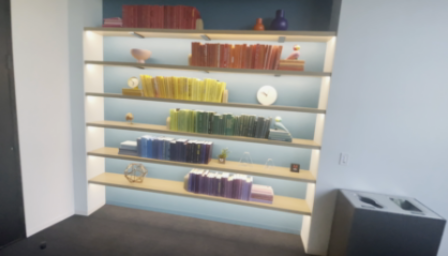}
    \includegraphics[width=0.22\textwidth]{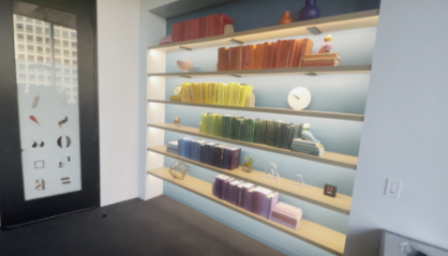}
    \includegraphics[width=0.22\textwidth]{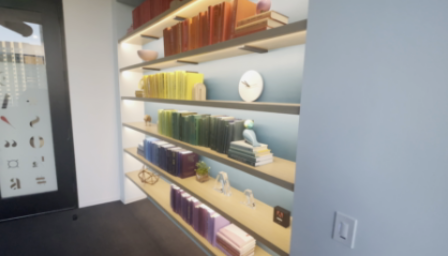}
}\\[0.35em]

\midrule

Target
& \ours \textit{(Ours)}
& MVSplat360~\cite{chen2024mvsplat360}
& latentSplat~\cite{wewer2024latentsplat}
\\[0.35em]

\includegraphics[width=0.22\textwidth]{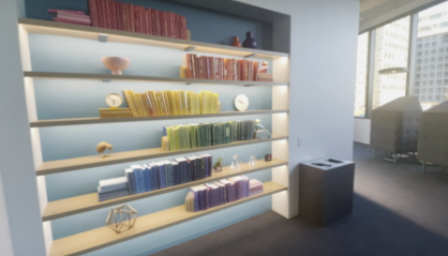}
&
\includegraphics[width=0.22\textwidth]{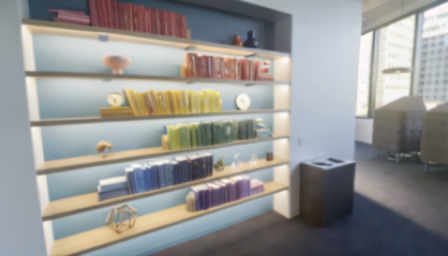}
&
\includegraphics[width=0.22\textwidth]{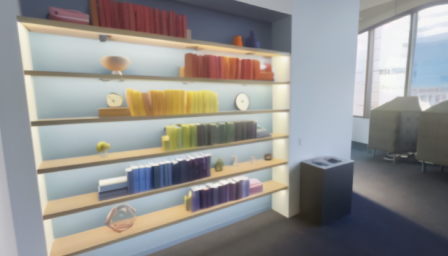}
&
\includegraphics[width=0.22\textwidth]{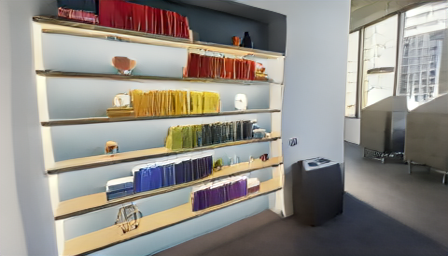}
\\

&
\includegraphics[width=0.22\textwidth]{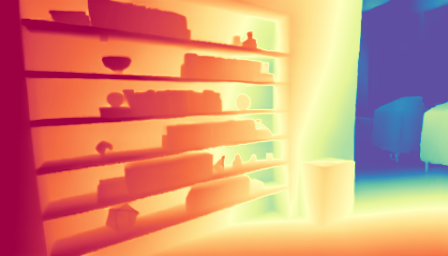}
&
\includegraphics[width=0.22\textwidth]{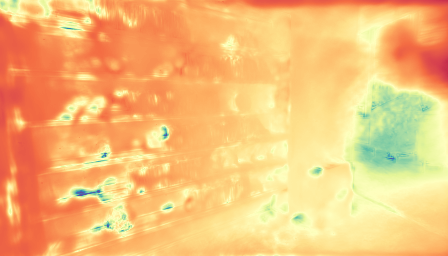}
&
\includegraphics[width=0.22\textwidth]{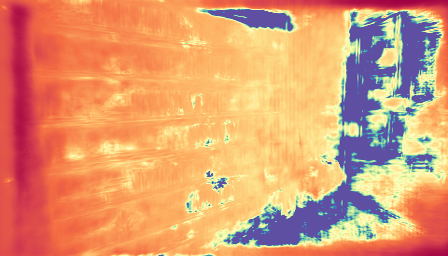}
\\[0.35em]

\includegraphics[width=0.22\textwidth]{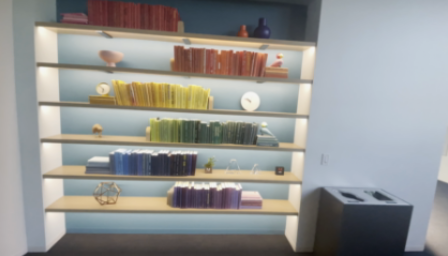}
&
\includegraphics[width=0.22\textwidth]{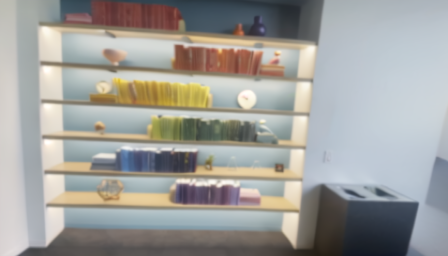}
&
\includegraphics[width=0.22\textwidth]{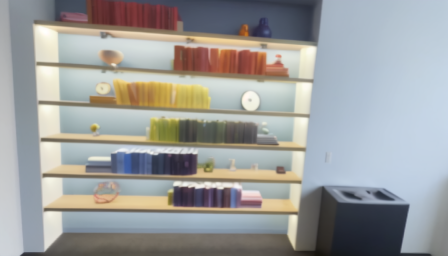}
&
\includegraphics[width=0.22\textwidth]{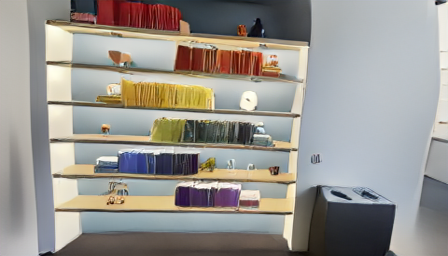}
\\

&
\includegraphics[width=0.22\textwidth]{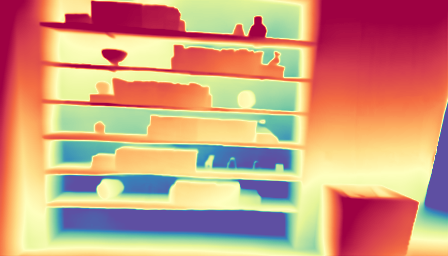}
&
\includegraphics[width=0.22\textwidth]{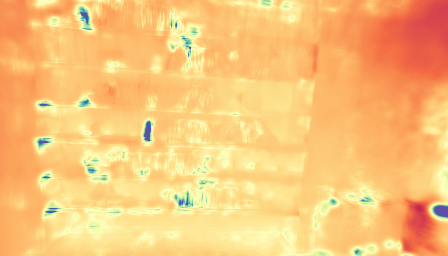}
&
\includegraphics[width=0.22\textwidth]{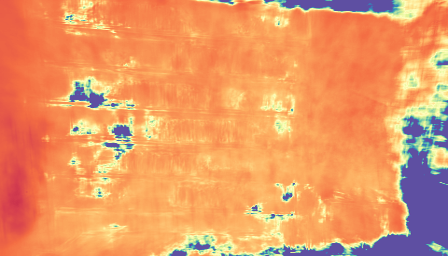}
\\[0.35em]

\includegraphics[width=0.22\textwidth]{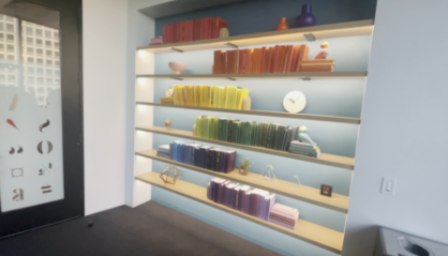}
&
\includegraphics[width=0.22\textwidth]{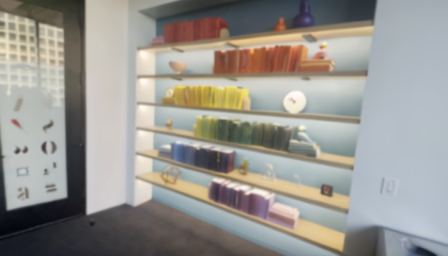}
&
\includegraphics[width=0.22\textwidth]{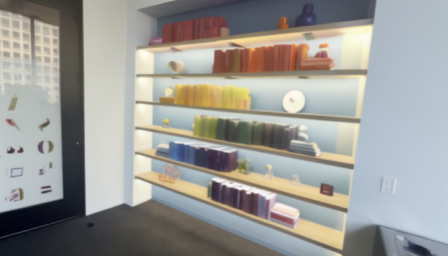}
&
\includegraphics[width=0.22\textwidth]{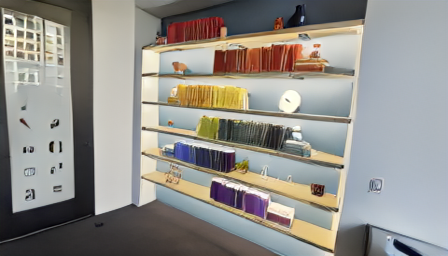}
\\

&
\includegraphics[width=0.22\textwidth]{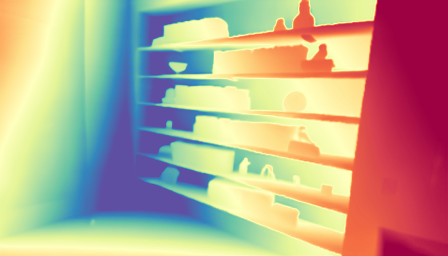}
&
\includegraphics[width=0.22\textwidth]{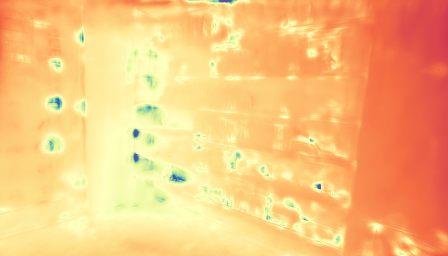}
&
\includegraphics[width=0.22\textwidth]{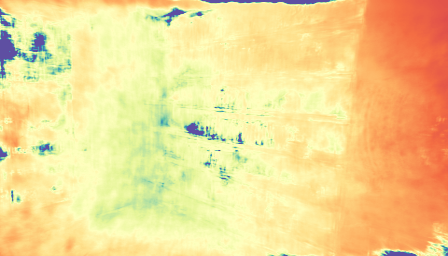}
\\[0.35em]

\includegraphics[width=0.22\textwidth]{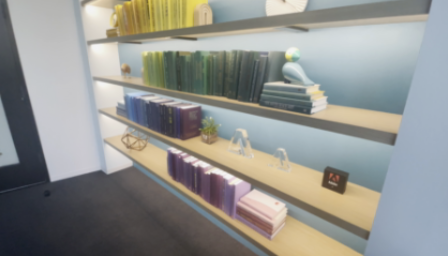}
&
\includegraphics[width=0.22\textwidth]{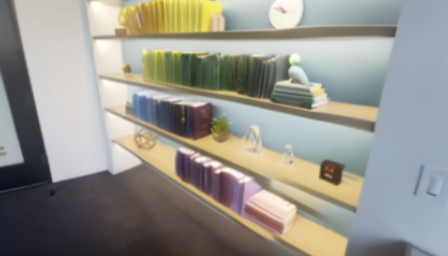}
&
\includegraphics[width=0.22\textwidth]{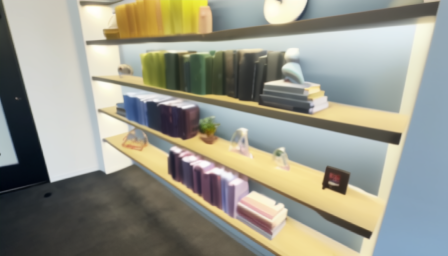}
&
\includegraphics[width=0.22\textwidth]{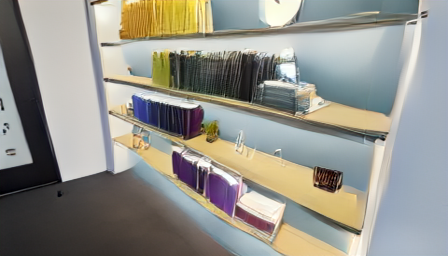}
\\

&
\includegraphics[width=0.22\textwidth]{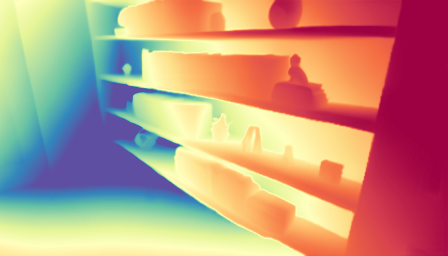}
&
\includegraphics[width=0.22\textwidth]{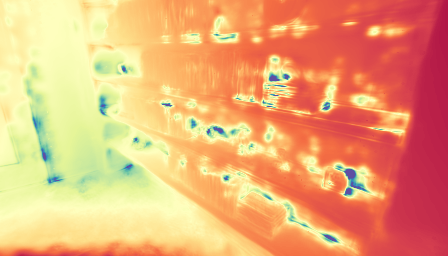}
&
\includegraphics[width=0.22\textwidth]{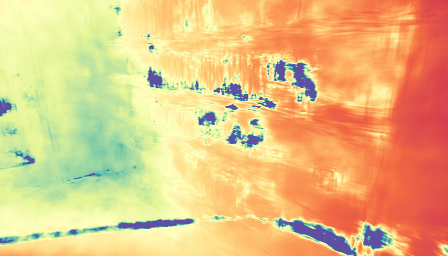}
\\

\end{tabular}

%% file: figures/app_qualitative/qual_dl3dv4.tex
\scriptsize
\sffamily
\setlength{\tabcolsep}{1pt}
\renewcommand{\arraystretch}{1.0}

\newcommand{\imgwidth}{0.21}

\begin{tabular}{cccc}

\multicolumn{4}{c}{$\mathcal{I}^c$}\\[0.35em]

\multicolumn{4}{c}{
    \includegraphics[width=0.22\textwidth]{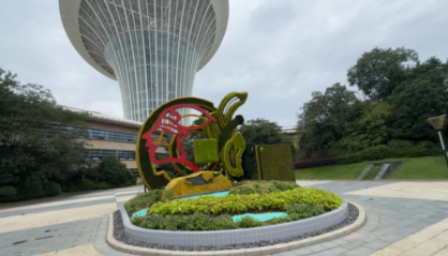}
    \includegraphics[width=0.22\textwidth]{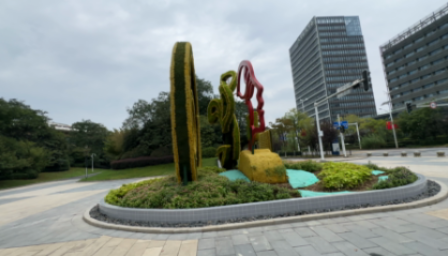}
    \includegraphics[width=0.22\textwidth]{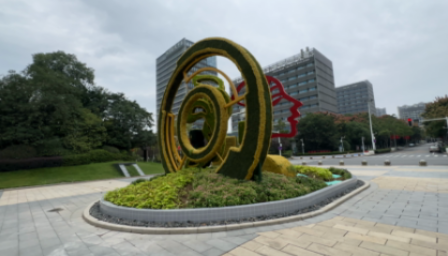}
    \includegraphics[width=0.22\textwidth]{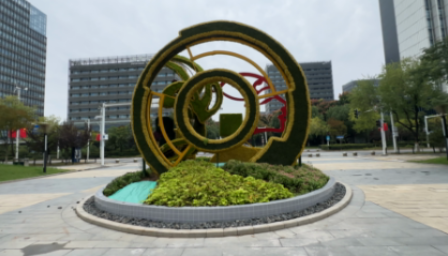}
}\\[0.35em]

\midrule

Target
& \ours \textit{(Ours)}
& MVSplat360~\cite{chen2024mvsplat360}
& latentSplat~\cite{wewer2024latentsplat}
\\[0.35em]

\includegraphics[width=0.22\textwidth]{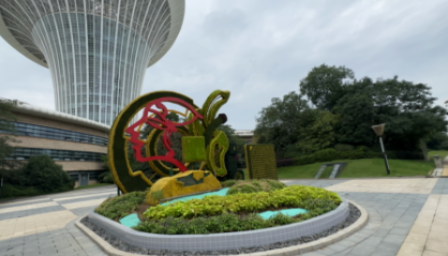}
&
\includegraphics[width=0.22\textwidth]{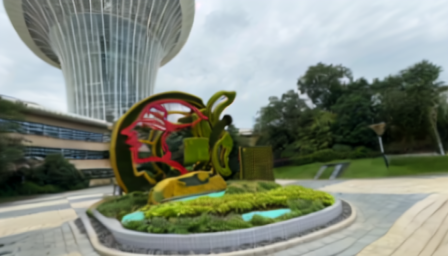}
&
\includegraphics[width=0.22\textwidth]{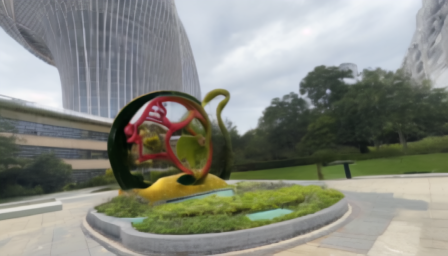}
&
\includegraphics[width=0.22\textwidth]{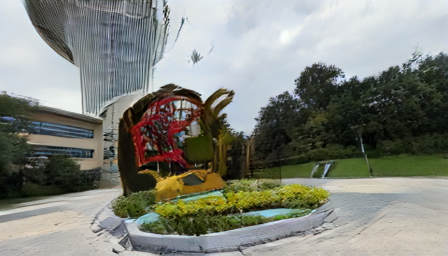}
\\

&
\includegraphics[width=0.22\textwidth]{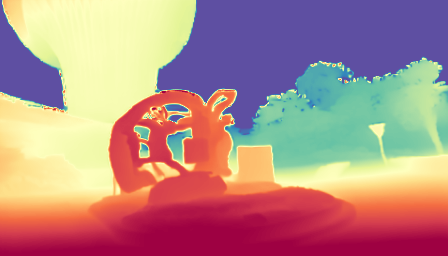}
&
\includegraphics[width=0.22\textwidth]{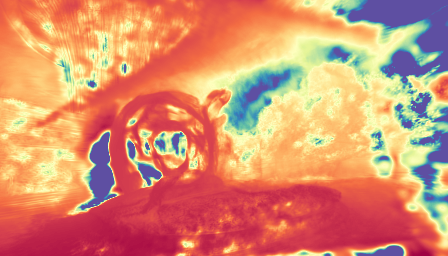}
&
\includegraphics[width=0.22\textwidth]{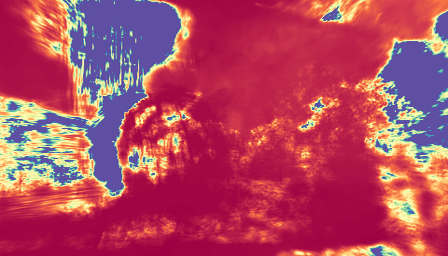}
\\[0.35em]

\includegraphics[width=0.22\textwidth]{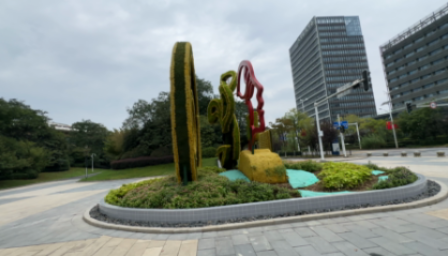}
&
\includegraphics[width=0.22\textwidth]{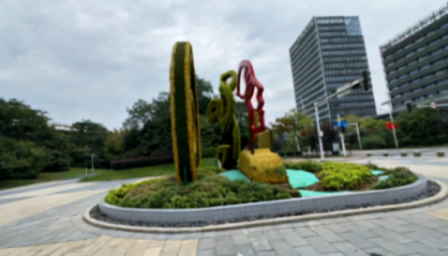}
&
\includegraphics[width=0.22\textwidth]{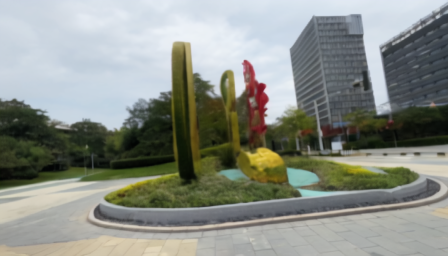}
&
\includegraphics[width=0.22\textwidth]{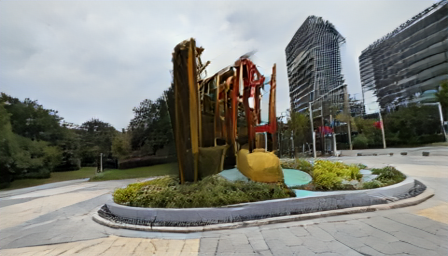}
\\

&
\includegraphics[width=0.22\textwidth]{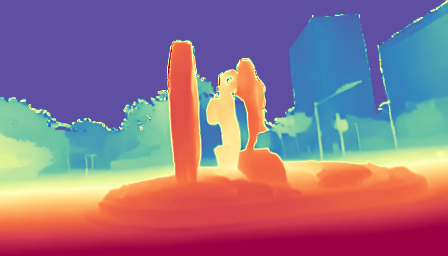}
&
\includegraphics[width=0.22\textwidth]{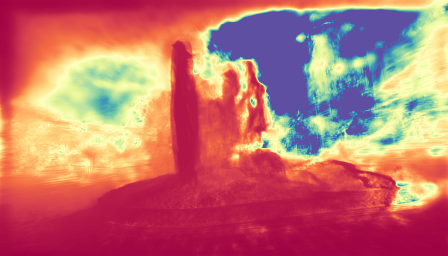}
&
\includegraphics[width=0.22\textwidth]{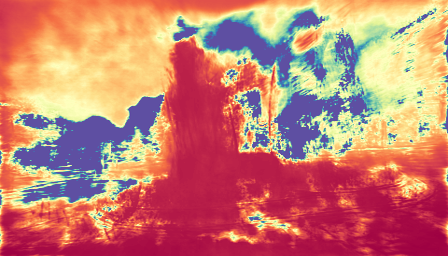}
\\[0.35em]

\includegraphics[width=0.22\textwidth]{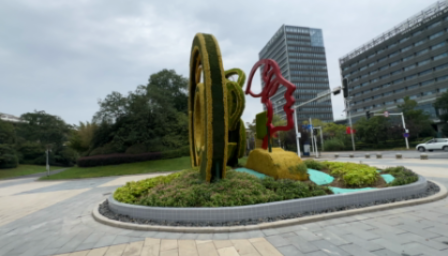}
&
\includegraphics[width=0.22\textwidth]{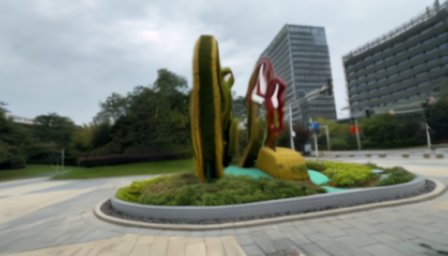}
&
\includegraphics[width=0.22\textwidth]{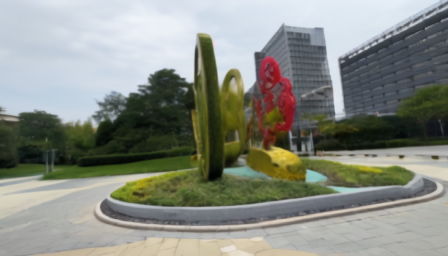}
&
\includegraphics[width=0.22\textwidth]{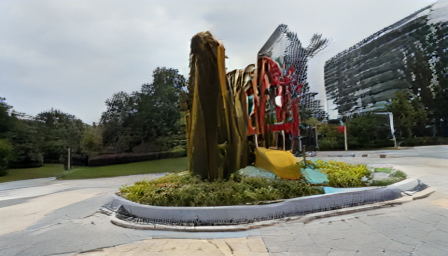}
\\

&
\includegraphics[width=0.22\textwidth]{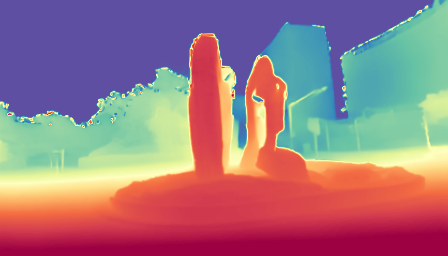}
&
\includegraphics[width=0.22\textwidth]{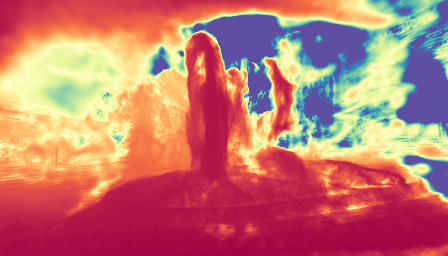}
&
\includegraphics[width=0.22\textwidth]{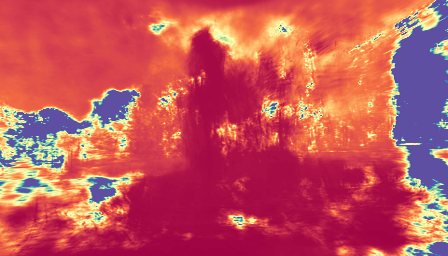}
\\[0.35em]

\includegraphics[width=0.22\textwidth]{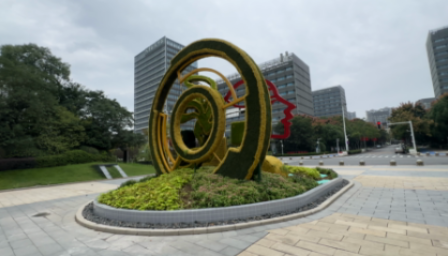}
&
\includegraphics[width=0.22\textwidth]{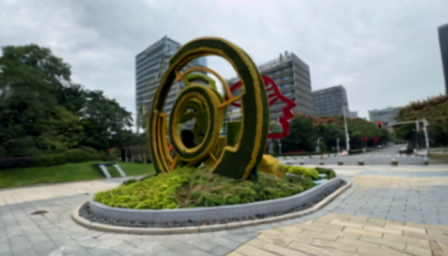}
&
\includegraphics[width=0.22\textwidth]{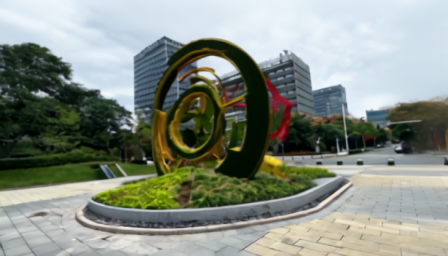}
&
\includegraphics[width=0.22\textwidth]{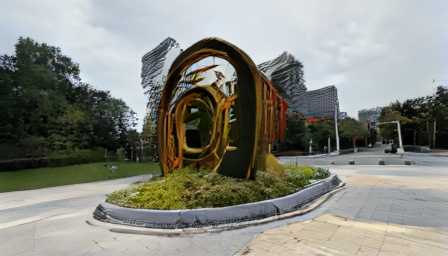}
\\

&
\includegraphics[width=0.22\textwidth]{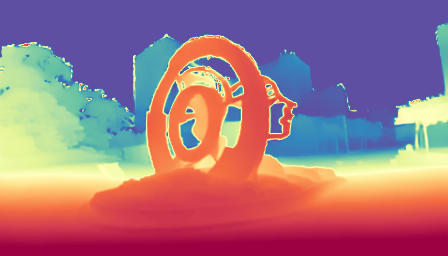}
&
\includegraphics[width=0.22\textwidth]{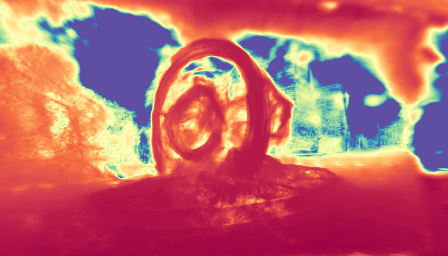}
&
\includegraphics[width=0.22\textwidth]{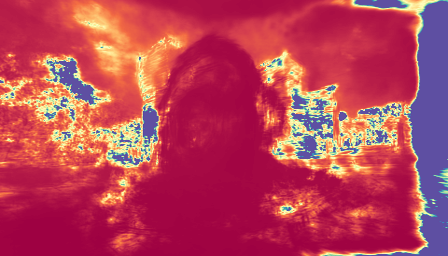}
\\

\end{tabular}